\documentclass[10pt,twocolumn,letterpaper]{article}
\usepackage{cvpr}              

\usepackage{times}
\usepackage{epsfig}
\usepackage{graphicx}
\usepackage{amsmath}
\usepackage{amssymb}

\usepackage{multirow} 
\usepackage{makecell} 
\usepackage[table,dvipsnames,xcdraw]{xcolor} 

\usepackage{colortbl} 
\usepackage{diagbox} 
\usepackage{siunitx}

\usepackage{enumitem} 
\usepackage{arydshln} 
\usepackage{tabularx}
\usepackage{booktabs} 
\usepackage{xspace} 
\usepackage[subrefformat=parens,labelformat=parens]{subcaption} 
\usepackage[Export]{adjustbox} 
\usepackage{pifont} 
\usepackage{balance} 
\usepackage{wrapfig}
\usepackage[linesnumbered,ruled]{algorithm2e}

\definecolor{cvprblue}{rgb}{0.21,0.49,0.74}
\usepackage[pagebackref,breaklinks,colorlinks,citecolor=cvprblue]{hyperref}

\usepackage[capitalize]{cleveref}

\def\eg{\emph{e.g.}\xspace} 
\def\ie{\emph{i.e.}\xspace}

\newcommand{\ours}{EN-SLAM\xspace}
\newcounter{boldpara}
\newcommand{\boldparagraph}[1]{
    \refstepcounter{boldpara}
    \vspace{0.1em}\noindent{\bf #1}
}

\newcommand{\greencheck}{{\color{green}\checkmark}}
\newcommand{\graycheck}{{\color{gray}\checkmark}}

\newcommand{\redx}{{\color{red}\ding{55}}}
\newcommand{\blackx}{{\color{black}\ding{55}}}

\newcommand\blfootnote[1]{%
  \begingroup
  \renewcommand\thefootnote{}\footnote{#1}%
  \addtocounter{footnote}{-1}%
  \endgroup
}

\colorlet{colorFst}{Green!25}       
\colorlet{colorSnd}{SpringGreen!45} 
\colorlet{colorTrd}{Yellow!30}      
\colorlet{colorLow}{darkgray!30}    
\definecolor{R1}{HTML}{E97451}
\definecolor{R2}{HTML}{008080}
\definecolor{R3}{HTML}{0047AB}
\colorlet{cmt}{darkgray!80}    
\colorlet{supp}{darkgray!50}    

\newcommand{\fs}{\cellcolor{colorFst}\bf}   
\newcommand{\nd}{\cellcolor{colorSnd}}      

\definecolor{gray}{rgb}{0.65,0.65,0.65}
\definecolor{mycol}{rgb}{0.90,0.95,1.0}

\def\paperID{357}
\def\confName{CVPR}
\def\confYear{2024}

\title{Implicit Event-RGBD Neural SLAM}

\author{
        Delin Qu$^{1,2}$\footnotemark[1]  \hspace{1em} 
        Chi Yan$^{2,5}$\footnotemark[1] \hspace{1em} 
        Dong Wang$^{2}$ \hspace{1em} 
        Jie Yin$^{3}$ \hspace{1em} 
        Qizhi Chen$^{2}$ \\
        Dan Xu$^{5}$ \hspace{1em}
        Yiting Zhang$^{2}$ \hspace{1em}
        Bin Zhao$^{2,4}$\footnotemark[2] \hspace{1em}
        Xuelong Li$^{2}$ \\
        $^{1}$Fudan University \hspace{2em}
        $^{2}$Shanghai AI Laboratory \hspace{2em}
        $^{3}$Shanghai Jiao Tong University \\
        $^{4}$Northwestern Polytechnical University \hspace{1em}
        $^{5}$Hong Kong University of Sciences and Technology
}

\begin{document}
\maketitle

\begin{abstract}
  Implicit neural SLAM has achieved remarkable progress recently. Nevertheless, existing methods face significant challenges in non-ideal scenarios, such as motion blur or lighting variation, which often leads to issues like convergence failures, localization drifts, and distorted mapping. To address these challenges, we propose \textbf{\ours}, the first event-RGBD implicit neural SLAM framework, which effectively leverages the high rate and high dynamic range advantages of event data for tracking and mapping. Specifically, EN-SLAM proposes a differentiable CRF (Camera Response Function) rendering technique to generate distinct RGB and event camera data via a shared radiance field, which is optimized by learning a unified implicit representation with the captured event and RGBD supervision. Moreover, based on the temporal difference property of events, we propose a temporal aggregating optimization strategy for the event joint tracking and global bundle adjustment, capitalizing on the consecutive difference constraints of events, significantly enhancing tracking accuracy and robustness. Finally, we construct the simulated dataset \textbf{DEV-Indoors} and real captured dataset \textbf{DEV-Reals} containing 6 scenes, 17 sequences with practical motion blur and lighting changes for evaluations. Experimental results show that our method outperforms the SOTA methods in both tracking ATE and mapping ACC with a real-time $17$ FPS in various challenging environments. Project page: \href{https://delinqu.github.io/EN-SLAM}{https://delinqu.github.io/EN-SLAM}.
  \blfootnote{$\ast$ Authors contributed equally: \href{mailto:dlqu22@m.fudan.edu.cn}{dlqu22@m.fudan.edu.cn}}
  \blfootnote{$\dagger$ Corresponding author}
\end{abstract}

\section{Introduction}
\label{sec:intro}
\begin{figure}[t]
    \vspace{-2ex}
    \begin{center}
        \includegraphics[width=1\linewidth]{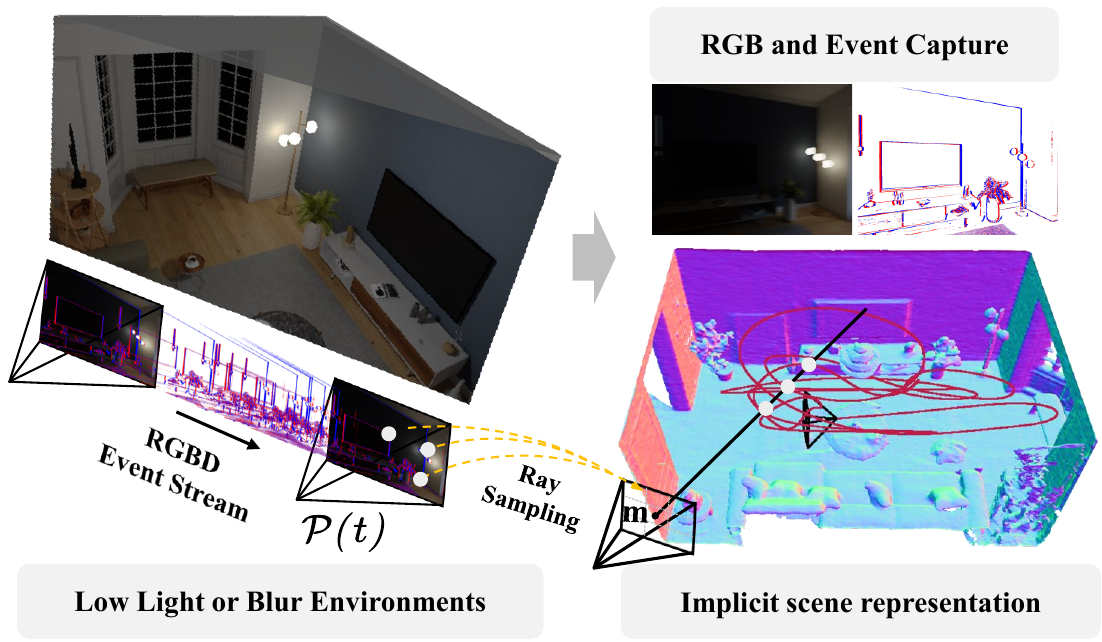}
    \end{center}
    \vspace{-1ex}
    \caption{Illustration of the proposed implicit event-RGBD neural SLAM system \textbf{EN-SLAM} under non-ideal environments. The dynamic range of RGB sensors is relatively low and suffers from motion blur. Instead, event cameras show great potential in non-ideal environments due to their high dynamic range and low latency advantages. Our method samples rays from two independent RGBD and event cameras to jointly train a single implicit neural field with both modalities. This hybrid shared mechanism provides a natural fusion approach, avoiding alignment issues. It also leverages the advantages of both modalities, resulting in dense, more robust, and higher-quality reconstruction results.}
    \label{fig:head}
    \vspace{-2ex}
\end{figure}
Simultaneous Localization and Mapping (SLAM) is an essential problem in computer vision and robotics, widely applied in tasks such as virtual and augmented reality~\cite{Desai2014ARP}, robot navigation~\cite{Hne2011StereoDM} and autonomous driving~\cite{Bresson2017SimultaneousLA} over last decades. Exploration in extreme environments~\cite{ebadi2022present,liao2023revisiting,Qu_2023_ICCV} remains challenging for visual SLAM (vSLAM) systems due to the lack of visual features caused by motion blur and lighting variation in diverse environments~\cite{wang2020tartanair,zhao2023subtmrs,drsc}.

\begin{figure}[t]
    \centering
    {\footnotesize
        \setlength{\tabcolsep}{0.2pt}
        \renewcommand{\arraystretch}{0}
        \newcommand{\sz}{0.33}
        \begin{tabular}{ccc}
            \includegraphics[valign=c,width=\sz\linewidth]{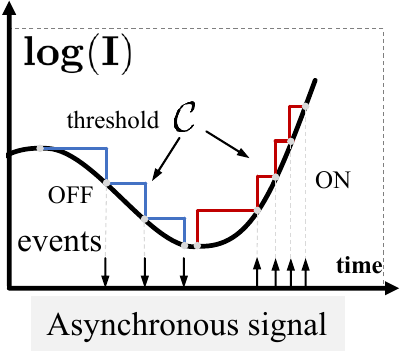} & \includegraphics[valign=c,width=\sz\linewidth]{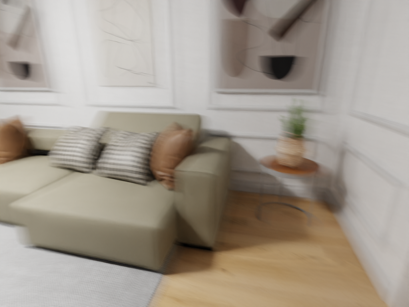} & \includegraphics[valign=c,width=\sz\linewidth]{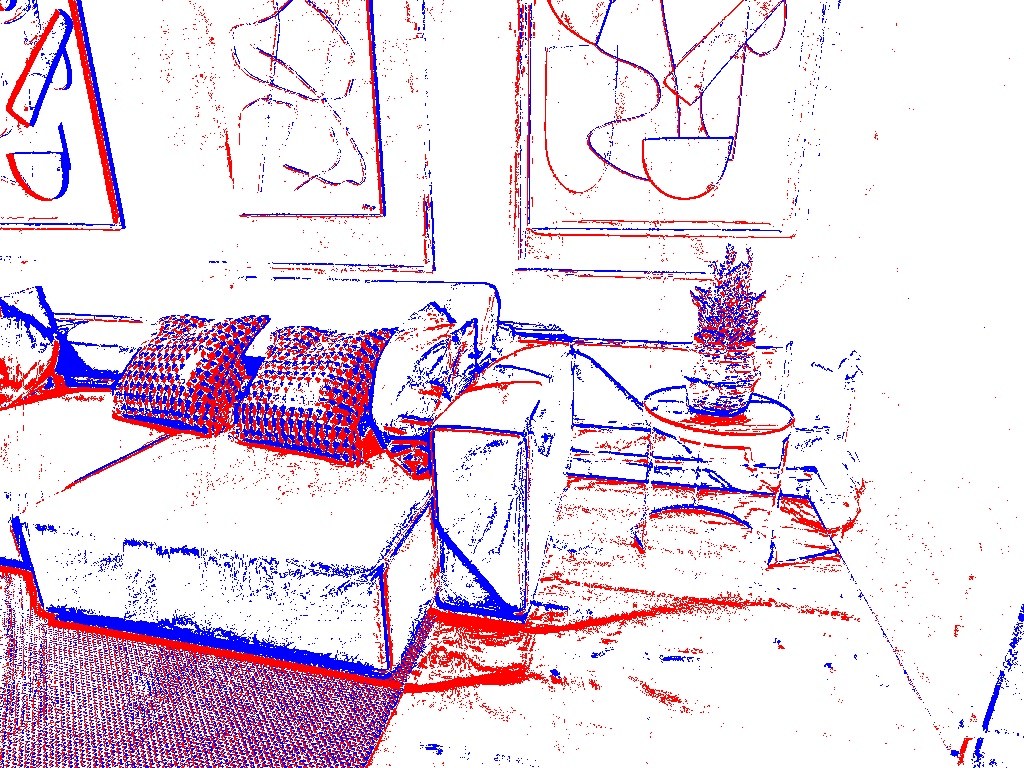} \\
            event trigger mechanism                                                           & RGB capture (blur)                                                       & event capture
        \end{tabular}
    }
    \caption{\textbf{Illustration of the Event Generation Model (EGM).} An event is triggered at a single pixel if the corresponding logarithmic change in luminance exceeds a threshold $C$.}
    \label{fig:EGM}
    \vspace{-2.ex}
\end{figure}

As a novel representation for myriad of signals, Neural Radiance Fields (NeRF)~\cite{mildenhall2020nerf} has innovated great progress in SLAM recently, demonstrating significant improvements in map memory consumption, hole filling, and mapping quality~\cite{Sucar2021iMAPIM,Zhu2021NICESLAMNI,Wang2023CoSLAMJC,Johari2022ESLAMED,pointslam,zhang2023goslam}. While the existing NeRF-based neural vSLAM methods address the limitations of traditional SLAM frameworks~\cite{Newcombe2011DTAMDT,Whelan2012KintinuousSESE,Engel2014LSDSLAMLD,MurArtal2016ORBSLAM2AO,Schps2019BADSB} in accurate dense 3D map reconstruction, they are primarily designed for well-lit scenes and always fail in practical SLAM scenarios with motion blur~\cite{Qi_2023_ICCV,Zhang_2023_ICCV} and lighting variation~\cite{Ma_2023_ICCV,Rudnev2022EventNeRFNR}. These methods produce unsatisfying results under non-ideal environments~\cite{Wang2023CoSLAMJC} because of the following limitations: \noindent\textbf{1) View-inconsistency:} When the camera encounters rapid velocity variation in~\cref{fig:EGM} (2nd), the scene may exhibit discontinuous blur, leading to view-inconsistency among frames, further causing heavy artifacts in the reconstructed map. \noindent\textbf{2) Low dynamic range:} In lighting variation scenes illustrated in~\cref{fig:head}, the dynamic range of the RGB sensor is relatively low, and the information on the dark and overexposure areas is lost, leading to tracking drifts and mapping distortions.

To address the issues in non-ideal scenarios of existing neural vSLAM, we introduce utilizing the advantages of high dynamic range (HDR) and temporal resolution of event data to compensate for the lost information, thereby improving the robustness, efficiency, and accuracy of current neural vSLAM in extreme environments. Fig.~\ref{fig:EGM} shows the event generation model that an event is triggered at a single pixel if the corresponding logarithmic change in luminance exceeds a threshold $C$. This asynchronous mechanism shows excellent potential in non-ideal environments due to its advantages in low latency~\cite{zhou2021event,rebecq2016evo,kim2016real}, high dynamic range~\cite{rebecq2019high}, and high temporal resolution~\cite{tulyakov2021time,tulyakov2022time}. \cref{fig:head} and \cref{fig:EGM} illustrate its superiority in dark and fast motion, and event sensors capture higher-quality signals than RGB sensors. However, applying events into NeRF-based vSLAM is challenging due to the significant distinction in imaging mechanisms between event and RGB cameras. Moreover, the requirement of highly accurate camera poses and careful optimization in traditional surface density estimation~\cite{Ma_2023_ICCV} further complicates the integration.

\begin{figure*}[t]
    \vspace{-5.5ex}
    \begin{center}
        \includegraphics[width=1\linewidth]{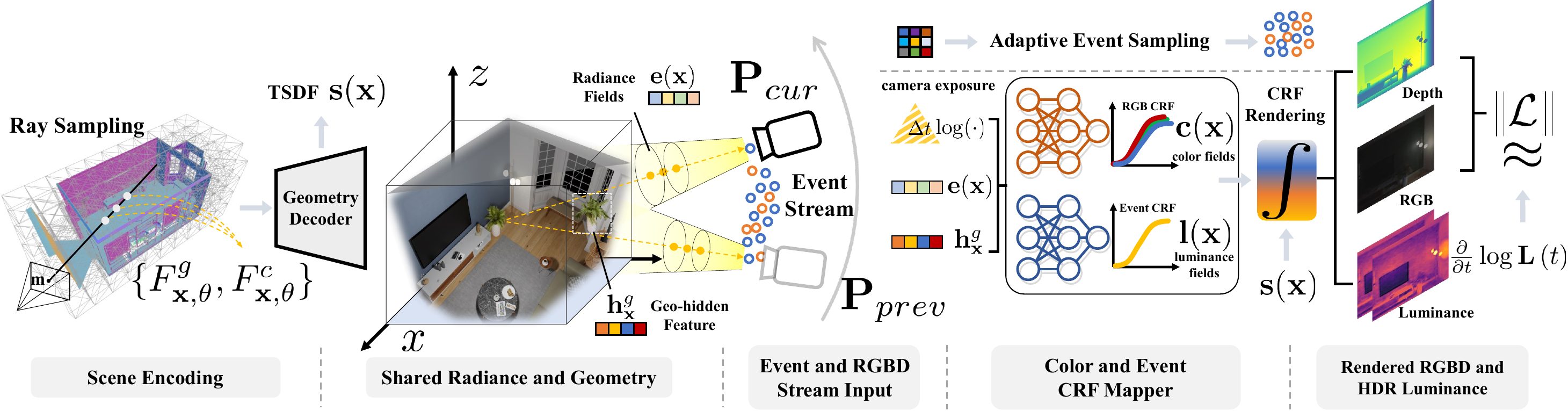}
    \end{center}
    \vspace{-1ex}
    \caption{\textbf{Overview of \ours}. EN-SLAM decodes the scene encoding to a shared geometry and radiance representation, and decomposes the radiance into RGB color $\mathbf{c}(\mathbf{x})$ and event luminance $\mathbf{l}(\mathbf{x})$ via differentiable CRF Mappers. We iteratively optimize the pose and scene representation by minimizing losses, in tracking and global BA with the event temporal aggregating techniques in~\cref{algorithm:ETA}.}
    \vspace{-2ex}
    \label{fig:pipeline}
\end{figure*}

In order to overcome these obstacles, we present \textit{\textbf{\ours}}, the first event-RGBD implicit neural SLAM framework that effectively harnesses the advantages of event and RGBD streams. The overview of \ours is shown in Fig.~\ref{fig:pipeline}. Our method models the differentiable imaging process of two distinct cameras and utilizes shared radiance fields to jointly learn a hybrid unified representation from events and RGBD data. By integrating the event generation model into the optimization process, we introduce the event temporal aggregating (ETA) optimization strategy for event joint tracking and global bundle adjustment (BA). This strategy effectively leverages the temporal difference property of events, providing efficient consecutive difference constraints and significantly improving the performance. Additionally, we construct two datasets: the simulated dataset \textbf{DEV-Indoors} and the real captured dataset \textbf{DEV-Reals}, which consist of 6 scenes and 17 sequences with practical motion blur and lighting changes. Contributions can be summarized as follows:

\begin{itemize}[leftmargin=10pt]
    \item We present \textbf{\ours}, the first event-RGBD implicit neural SLAM framework that efficiently leverages event stream and RGBD to overcome challenges in extreme motion blur and lighting variation scenes.
    \item A differentiable CRF rendering technique is proposed to map a unified representation in the shared radiance field to RGB and event camera data for addressing the significant distinction between event and RGB. A temporal aggregating optimization strategy that capitalizes the consecutive difference constraints of the event stream is present and significantly improves the camera tracking accuracy and robustness.
    \item We construct a simulated \textbf{DEV-Indoors} and real captured \textbf{DEV-Reals} dataset containing 17 sequences with practical motion blur and lighting changes. A wide range of evaluations demonstrate competitive real-time performance under various challenging environments.
\end{itemize}

\section{Related Work}
\label{sec:relate}

\boldparagraph{Neural Implicit vSLAM.}
Existing NeRF-based visual SLAM methods have made significant improvements in dense map reconstruction. iMAP~\cite{Sucar2021iMAPIM} first introduces NeRF into SLAM, and NICE-SLAM~\cite{Zhu2021NICESLAMNI} expands the reconstructable environment size by introducing multi-scale feature grids. Vox-Fusion~\cite{Yang2022VoxFusionDT} utilizes an octree-based structure to expand the scene dynamically. Besides, CoSLAM~\cite{Wang2023CoSLAMJC} combines coordinate and sparse parametric encoding to achieve fast convergence and surface hole filling in reconstruction. Parallel works ESLAM~\cite{Johari2022ESLAMED} and Point-SLAM~\cite{pointslam} represent scenes as multi-scale feature planes and neural point clouds, respectively, to improve the efficiency and accuracy. Beyond NeRF, GS-SLAM~\cite{yan2024gsslam} utilizes 3D Gaussian~\cite{kerbl3Dgaussians} for scene representation and achieves photo-realistic reconstruction performance. However, these methods are designed for well-lit indoor scenes and commonly encounter challenges in non-ideal SLAM processes, such as motion blur and lighting variation. In contrast, we introduce utilizing the advantages of high dynamic range and temporal resolution of events to compensate for the lost information, thereby improving the robustness and accuracy of current neural implicit methods.

\boldparagraph{Event-based SLAM.} Events have been incorporated into traditional visual SLAM systems to address the motion blur and lighting variation. These methods can be divided into three main types: \textit{feature-based methods}, \textit{direct methods}, and \textit{motion-compensation methods}. Feature-based methods, such as USLAM~\cite{vidal2018ultimate}, EIO~\cite{guan2022monocular} and PL-EVIO~\cite{guan2022pl}, track point or line features from event data~\cite{li2019fa,rebecq2017real}, and perform the camera tracking and mapping in parallel threads. However, the feature extraction algorithms~\cite{rosten2006machine,vasco2016fast,Brandli2016ELiSeDA,Manderscheid2019SpeedIT} rely heavily on frame-based feature detection, facing challenges for motion-dependent event data appearance. Direct methods employ events without explicit data association by aligning the photometric event image~\cite{Kim2014SimultaneousMA,EDS,kim2016real} or utilizing the spatiotemporal information for event representations alignment~\cite{rebecq2016evo,zhou2021event,Rebecq2018EMVSEM}. Direct methods are well suited for events, but mainly focus on event-based visual odometry, leaving the visual dense mapping unexplored. Motion-compensation methods optimize event alignment in motion-compensated event frames by maximizing the contrast~\cite{wang2022visual,peng2021globally}, minimizing the dispersion~\cite{nunes2021robust} and align probabilistic~\cite{gu2021spatio}. However, they suffer from the collapse in a broad range of camera motions. Thus, currently, event-based SLAM demonstrates significant potential but lacks sufficient exploration in dense map reconstructions~\cite{huang2023eventbased}.

\boldparagraph{Neural Radiance Fields using Events.}
Event-based NeRF is in the nascent stages, and several studies have demonstrated the possibility of view synthesis from events via implicit neural fields. Event-NeRF~\cite{Rudnev2022EventNeRFNR} proposes an approach for inferring NeRF from a monocular color event stream that enables novel view synthesis. E-NeRF~\cite{Klenk2022ENeRFNR} and E$^2$NeRF~\cite{Qi_2023_ICCV} tackle the NeRF estimation from event cameras under strong motion blur. They develop normalized and rendering loss to address varying contrast thresholds and enhance neural volumetric representation. The parallel work Ev-NeRF~\cite{Hwang2022EvNeRFEB} conducts a threshold-bound loss with the ReLU function to address the lack of RGB images. In addition to reconstruction, $\Delta_t$NeRF~\cite{Masuda2023EventbasedCT} proposes an event camera tracker by minimizing the error between sparse events and the temporal gradient of the scene representation on the simplified intensity-change events. However, the traditional surface density estimation in NeRF requires highly accurate camera poses and careful optimization~\cite{Ma_2023_ICCV}, thus making it exceptionally challenging to apply NeRF to the event-based SLAM.
\section{Methodology}
\label{sec:method}
The overview of our method is shown in~\cref{fig:pipeline}. Given an input RGBD stream $\left\{\mathbf{I}_{i}, \mathbf{D}_{i}\right\}_{i=1}^{J}$ and event stream $\left\{\mathbf{E}_{k}\right\}_{k=1}^{N}$ with known camera intrinsics $\mathbf{K} \in \mathbb{R}^{3 \times 3}$ and $\mathbf{K}^{\prime} \in \mathbb{R}^{3 \times 3}$, we aim to leverage event and RGBD to reconstruct the camera poses $\left\{\mathbf{P}_{i}\right\}_{i=1}^{J}$ and the implicit scene representation. In~\cref{sec:share_decouple}, the scene encoding is decoded to a unified geometry and radiance representation. Then, the shared radiance is decomposed into RGB color $\mathbf{c}(\mathbf{x})$ and event luminance $\mathbf{l}(\mathbf{x})$ via differentiable CRF Mappers in~\cref{sec:decomposition,sec:crf_rendering} to address the imaging distinction of event and RGB cameras. Finally, \ours iteratively optimizes the pose and scene representation by minimizing the re-rendering loss between the observed RGBD-E (RGBD and events) and rendering results in tracking and global BA of~\cref{sec:tracking_ba}.
\subsection{Unified Implicit Scene Representation}
\label{sec:share_decouple}
As shown in \cref{fig:pipeline}, we represent the scene $\mathbf{S}$ with multi-resolution geometric features and color grid features:
\begin{equation}
    \mathbf{S} = \{(F^g_{\mathbf{x},\theta}, F^c_{\mathbf{x},\theta}) \, | \, \theta=1,...,\Theta \} \enspace,
\end{equation}
where $\mathbf{x}$ and $\theta$ denote the coordinate and resolution level. There are two challenges that hinder us from learning a scene representation from different RGB and event modalities. Firstly, the event data is sparse and records logarithmic changes in luminance. Secondly, different cameras hold distinct physical imaging process mechanisms. Despite this, the geometry and radiance fields remain consistent during the camera imaging. In this case, we propose to learn a shared unified geometry hidden feature $\mathbf{h}_\mathbf{x}^g$ and radiance representation $\mathbf{e}(\mathbf{x})$ across distinct cameras. The geometric grid feature $F^g_{\mathbf{x},\theta}$ and color feature $F^c_{\mathbf{x},\theta}$ are simultaneously mapped to geometry hidden vector $\mathbf{h}_\mathbf{x}^g$, radiance fields $\mathbf{e}(\mathbf{x})$ and TSDF (truncated signed distance function) $\mathbf{s}(\mathbf{x})$,  by a geometry decoder $\boldsymbol{f_g}$:
\begin{equation}\label{eq:geo_decoder}
    \begin{aligned}
        \boldsymbol{f_g}\left(F^g_{\mathbf{x},\theta}, F^c_{\mathbf{x},\theta}\right) \mapsto(\mathbf{h}_\mathbf{x}^g, \mathbf{e}(\mathbf{x}), \mathbf{s}(\mathbf{x})).
    \end{aligned}
\end{equation}
The geometry hidden vector $\mathbf{h}_\mathbf{x}^g$ and radiance fields $\mathbf{e}(\mathbf{x})$ are shared by the color and event CRF decoders.
\subsection{Decomposition of the Radiance Fields}
\label{sec:decomposition}
In standard imaging devices, the incoming radiance undergoes linear and nonlinear image processing before being mapped into pixel values and stored in images. This entire image processing can be represented by a single function $\boldsymbol{f_c}$ called the camera response function (CRF)~\cite{Dufaux2016HighDR}. However, the traditional NeRF method~\cite{mildenhall2020nerf} simplifies the imaging process, leading to discrepancies between the rendering and actual images. This deviation is further amplified in multi-modal implicit representations. As \cref{fig:ablation_crf} shows, the captures of RGB and event cameras are significantly distinct, which can lead to joint optimization fluctuation. To address this issue, we model the radiance $\mathbf{e}(\mathbf{x})$ and exposure $\Delta \mathbf{t}_c$ of a ray $\mathbf{r}$ but take the aperture and others as implicit factors to obtain the color field $\mathbf{c}(\mathbf{x})$~\cite{Szeliski2010ComputerVA,Huang2021HDRNeRFHD} by differentiable tone-mapping:
\begin{equation}\label{eq:tone_mapping}
    \begin{aligned}
        \mathbf{c}(\mathbf{x}) = \boldsymbol{f_c}\left(\mathbf{e}(\mathbf{x}) \Delta \mathbf{t}_c \right).
    \end{aligned}
\end{equation}
To facilitate optimization, we convert all numerical values into the logarithmic domain and present the inverse function of $\left(\ln \boldsymbol{f_c}^{-1}\right)^{-1}$ as $\mathbf{\Psi_c}$:
\begin{equation}\label{eq:color_crf_mapper}
    \begin{aligned}
        \mathbf{c}(\mathbf{x}) & = \left(\ln \boldsymbol{f_c}^{-1}\right)^{-1}\left(\ln\mathbf{e}(\mathbf{r}) + \ln \Delta \mathbf{t}_c \right) \\
                               & = \mathbf{\Psi_c} \left(\ln\mathbf{\mathbf{e}(\mathbf{x})} + \ln \Delta \mathbf{t}_c \right),
    \end{aligned}
\end{equation}
As for the event camera, directly obtaining the event data is not feasible. However, we can predict high dynamic range luminance $l(\mathbf{x})$ and derive events using the event generation model: As shown in \cref{fig:EGM}, an event $E_{k} = (u_k, v_k, t_k, p_k)$ at image coordinate $\mathbf{m}_k = [u_k, v_k, 1]^T$ is triggered if the corresponding logarithmic brightness change $\mathbf{L}(\mathbf{m},t)$ exceeds a threshold $C$:
\begin{equation}\label{eq:EGM}
    \begin{aligned}
        \mathbf{L}(\mathbf{m}_k,t_k) - \mathbf{L}(\mathbf{m}_k,t_{k-1}) = p_k C, \enspace p_k \in \left \{ +1, -1 \right \}.
    \end{aligned}
\end{equation}
The logarithmic brightness $\mathbf{L}(\mathbf{m}, t)$ can be obtained by:
\begin{equation}\label{eq:linlog}
    \resizebox{0.85\linewidth}{!}{
        \begin{math}
            \begin{aligned}
                \mathbf{L}(\mathbf{m}, t)=\operatorname{lin} \log (\mathbf{I}_e(\mathbf{m}))=\left\{\begin{array}{ll}
                                                                                                        \mathbf{I}_e(\mathbf{m}) \cdot \ln (B) / B, & \text { if } \mathbf{I}_e(\mathbf{m})<B \\
                                                                                                        \ln (\mathbf{I}_e(\mathbf{m})),             & \text {else}
                                                                                                    \end{array}\right.
            \end{aligned}
        \end{math}
    },
\end{equation}
where $B$ denotes the linear region threshold~\cite{Hu2020V2EFV} and the imaging brightness $\textbf{I}_e(\mathbf{m})$ of event camera equals to the corresponding luminance of ray $\mathbf{r}$. By applying the modeling approach in~\cref{eq:tone_mapping,eq:color_crf_mapper} to the CRF of an event camera, we establish the relation among the luminance field $l(\mathbf{x})$, the radiance $\mathbf{e}(\mathbf{x})$ and exposure:
\begin{equation}\label{eq:lum_crf_mapper}
    \begin{aligned}
        l(\mathbf{x}) = \mathbf{\Psi}_l \left(\ln\mathbf{e}(\mathbf{x}) + \ln \Delta \mathbf{t}_l \right),
    \end{aligned}
\end{equation}
where $\mathbf{\Psi}_l$ and $\Delta \mathbf{t}_l$ denote the luminance tone-mapping and pseudo exposure of the event camera. In this way, we decompose the shared radiance field $\mathbf{e}(\mathbf{x})$ into the RGB and event camera imaging processes through two differentiable tone-mapping processes.
\subsection{Differentiable CRF Rendering}
\label{sec:crf_rendering}
Upon obtaining the color and luminance fields in~\cref{sec:decomposition}, we render the final imaging RGB, luminance, and depth by integrating predicted values along the samples in a ray $\mathbf{r}$:
\begin{align}
    \mathbf{x}_i = \mathbf{O} + z_i \mathbf{d}, \quad i \in \{1, ..., M\} \enspace,
    \label{eq:point-sample}
\end{align}
where $\mathbf{O} \in \mathbb{R}^{3}$ is the camera origin, $\mathbf{d}\in\mathbb{R}^{3}$, $\left \| \mathbf{d} \right \| = 1$ is the ray direction, and $z_i \in \mathbb{R}$ denotes the depth. Hence, we obtain the final imaging color $\hat{\mathbf{c}}$, luminance $\hat{l}$ and depth $\hat{d}$:

\vspace{-2ex}
\begin{equation}\label{eq:render}
    \resizebox{0.7\linewidth}{!}{
        \begin{math}
            \begin{aligned}
                \hat{\mathbf{c}}(\mathbf{r}, \Delta t_c) & =\sum_{i=1}^{i=M} w_i \mathbf{\Psi_c}(\ln \mathbf{e}(\mathbf{x}_i)+\ln \Delta t_c), \\
                \hat{l}(\mathbf{r}, \Delta t_l)          & =\sum_{i=1}^{i=M} w_i \mathbf{\Psi}_l(\ln \mathbf{e}(\mathbf{x}_i)+\ln \Delta t_l), \\
                \hat{d}(\mathbf{r})                      & =\sum_{i=1}^{i=M} w_i z_i.                                                          \\
            \end{aligned}
        \end{math}
    }
\end{equation}
\vspace{-2.5ex}

\noindent We utilize the simple bell-shaped model~\cite{azinovic2022neural} and compute weights $w_i$ by two sigmoid functions $\sigma(\cdot)$ to convert predicted TSDF $\mathbf{s}(\mathbf{x}_i)$ into weight $w_i$:
\begin{equation}\label{eq:sdf_weight}
    \begin{aligned}
        w_i =\sigma\left(\frac{\mathbf{s}(\mathbf{x}_i)}{tr}\right) \sigma\left(-\frac{\mathbf{s}(\mathbf{x}_i)}{tr}\right),
    \end{aligned}
\end{equation}
where $tr$ is the truncation distance of a ray $\mathbf{r}$.
\begin{figure}[t]
    \vspace{-1ex}
    \begin{center}
        \includegraphics[width=1\linewidth]{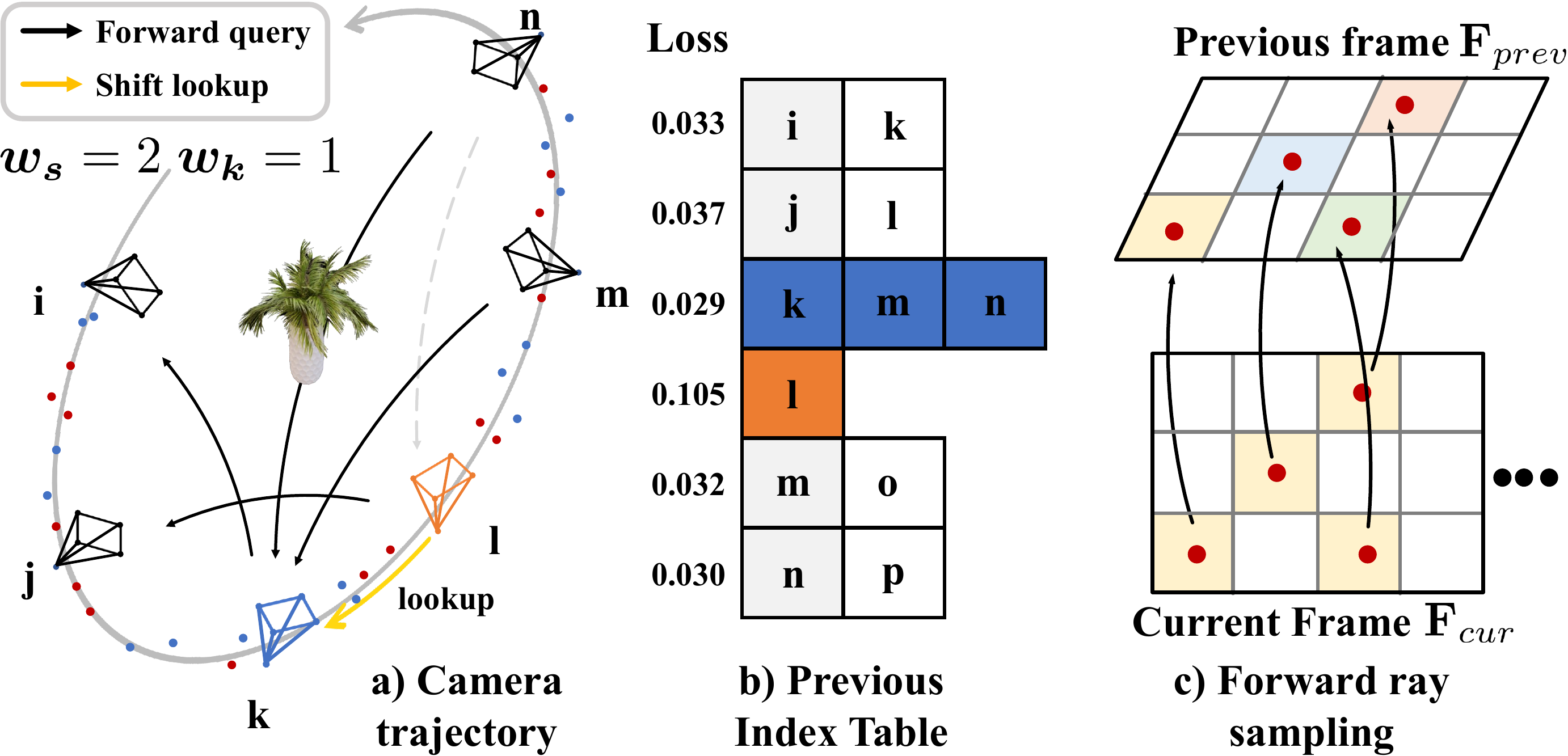}
    \end{center}
    \vspace{-2ex}
    \caption{The illustration of event temporal aggregating optimization strategy. In the tracking and global BA stages, EN-SLAM adaptively forwards query the previous frame according to the previous index table, and sample rays from different views perform joint optimization in \cref{eq:event_loss}.}
    \label{fig:event_temporal_aggregating}
    \vspace{-2ex}
\end{figure}

\subsection{Tracking and Bundle Adjustment}
\label{sec:tracking_ba}
In this section, to leverage the HDR and temporal difference properties of events, we propose an event joint tracking and global BA strategy in \cref{sec:event_temporal_aggregating} that incorporates events into optimization, thus improving the accuracy and robustness. Besides, we introduce adaptive forward-query and sampling strategies in \cref{sec:adaptive_forward_query} and \cref{sec:sampler}, which select event data and ray samples with more elevated confidence for optimization, thereby boosting the convergence.
\subsubsection{Event Temporal Aggregating Optimization}
\label{sec:event_temporal_aggregating}
The overview of ETA is shown in~\cref{fig:event_temporal_aggregating,algorithm:ETA}. \textbf{For tracking}, we representate the camera pose $\mathbf{P}_{cur} = \exp{(\xi^{\wedge}_{t})} \in \mathbb{SE}(3)$ of current frame $\mathbf{F}_{cur}$ and initialize with constant assumption. By selecting $N_t$ rays within $\mathbf{F}_{cur}$ and performing an adaptive event forward query in \cref{sec:adaptive_forward_query} with a probability-weighted sampling in \cref{sec:sampler}, we get the event stream and previous rays. Then, we iteratively optimize the pose by minimizing objective functions. \textbf{For global BA}, $N_{ba}$ rays from the global keyframe list are sampled to be the subset of pixels $\left\{PX_{cur}\right\}$. And then, the forward query and probability-weighted sampling are performed for each sample to get the previous subset $\left\{PX_{prev}\right\}$ and events $\left\{E_k\right\}_{k=prev}^{cur}$. Finally, a joint optimization is performed to optimize the geometry decoder $\boldsymbol{f_g}$, differentiable CRF $\left\{\mathbf{\Psi_c}, \mathbf{\Psi}_l\right\}$, and poses $\left\{\mathbf{P_i}\right\}_{i=0}^{cur}$.

\subsubsection{Adaptive Forward Event Query}
\label{sec:adaptive_forward_query}
As shown in \cref{fig:event_temporal_aggregating}, ETA performs an adaptive event forward window selection in tracking and BA by constructing a previous index table to prioritize reliable prior frames for optimization. Specifically, ETA uses a default window $w_d$ and performs a forward query. If the loss of a queried frame exceeds a threshold $\mathcal{L}_s$, we conduct a shift lookup within a neighborhood of length $2\times w_k$, selecting the frame with the minimum loss as the forward frame for event loss calculation \cref{eq:event_loss}. This event temporal constraint provides a stable local constraint between the participating frames and effectively leverages the high HDR property of events.

\vspace{-2ex}
\begin{algorithm}[htbp]
    \small
    \SetKwInOut{Input}{Input}
    \SetKwInOut{Output}{Output}
    \Input{RGBD-E stream $\left\{\mathbf{I}_{i}, d_{i}\right\}_{i=1}^{J}$ and $\left\{\mathbf{E}_{k}\right\}_{k=1}^{N}$.}
    \Output{Loss $\{\mathcal{L}_{ev}, \mathcal{L}_{rgb}, \mathcal{L}_{d}, \mathcal{L}_{sdf}, \mathcal{L}_{fs}\}$.}
    \While{$j < J$}{
    \For{$i \in \{j\}~\mathbf{if}~\mathbf{not}~\mathbf{BA}~\mathbf{else}~\{0,1,...,j\}$}{
    \tcc{Forward Query~\cref{sec:adaptive_forward_query}}
    $\mathbf{F}_{cur}, \mathbf{F}_{prev} \gets$ Tab($i$), Tab($i - w_d$)\;
    \If{$\mathcal{L}_{total} > \mathcal{L}_s$ }{
        $\mathbf{F}_{prev} \stackrel{\text {min}}{\gets} \text{Tab}(i-w_s-w_d, i+w_s)$.Loss
    }
    Probability-weighted ray sampling: $\text{Rays}_{cur}$, $\text{Rays}_{prev}$ $\gets$ $\mathbf{F}_{cur}, \mathbf{F}_{prev}$ \tcp*{\cref{sec:sampler}}
    Ray rendering: $\hat{\mathbf{L}}(\mathbf{m}, t_\beta) - \hat{\mathbf{L}}(\mathbf{m}, t_\alpha)$\tcp*{\cref{sec:crf_rendering}}
    Event accumulation: $\sum_{t_k=t_{cur}}^{t_k=t_{prev}} p_k C$\;
    Calculate loss: $\mathcal{L}_{ev}, \mathcal{L}_{rgb}, \mathcal{L}_{d}, \mathcal{L}_{sdf}, \mathcal{L}_{fs}$ \tcp*{\cref{eq:event_loss,eq:rgb_depth_loss,eq:loss_fn}}
    Tab(i) $\gets \{\mathcal{L}_{total}, \text{cur}, \text{prev}\}$
    }
    }
    \caption{{\small Event temporal aggregating optimization}}
    \label{algorithm:ETA}
\end{algorithm}
\vspace{-3ex}

\subsubsection{Probability-weighted Sampling Strategy}
\label{sec:sampler}
\begin{figure}[t]
    \vspace{-5.0ex}
    \begin{center}
        \includegraphics[width=0.8\linewidth]{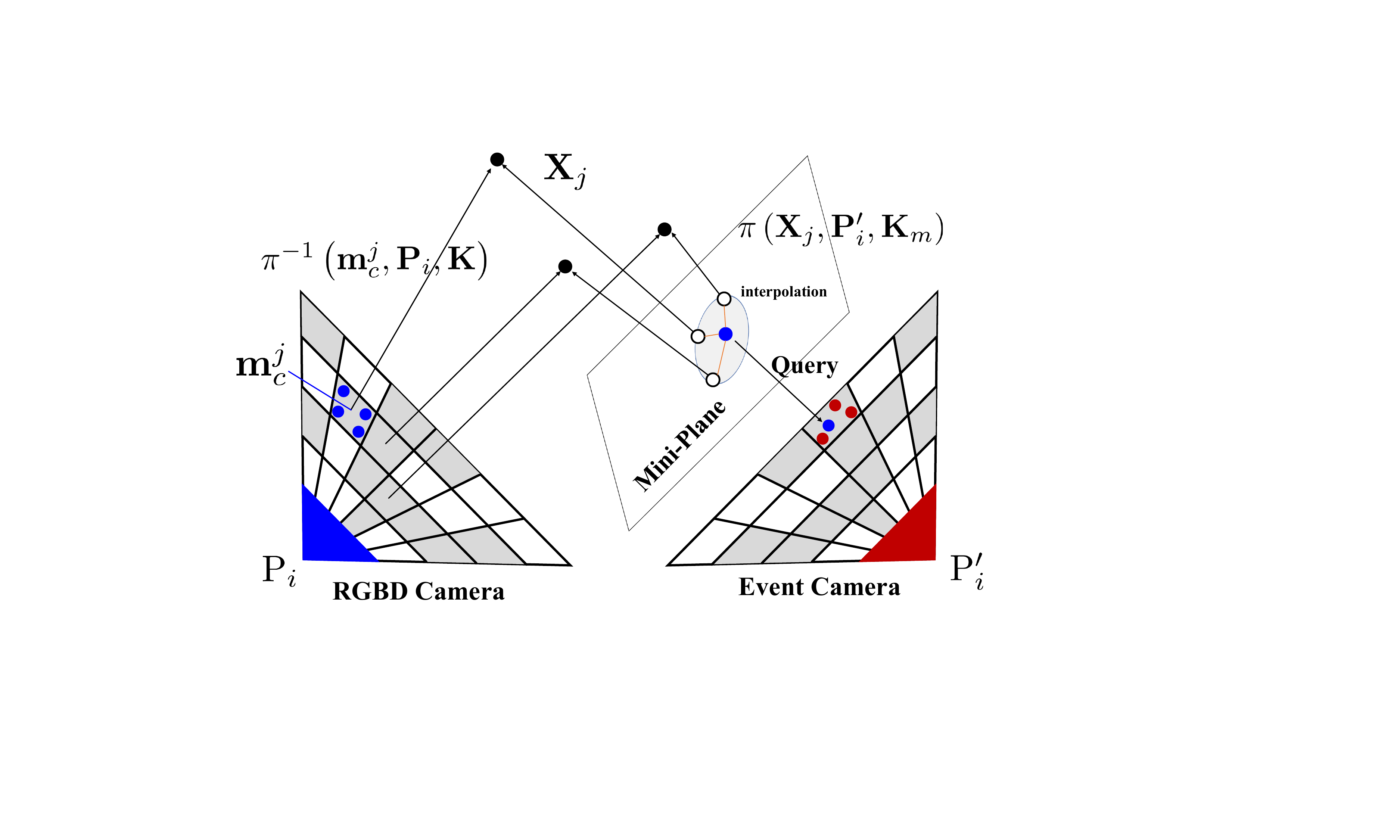}
    \end{center}
    \vspace{-2.ex}
    \caption{Illustration of the proposed probability-weighted sampling strategy. We utilize the loss of the RGBD plane (left) to guide ray sampling in the event plane (right).}
    \label{fig:prob_weight_sampling}
    \vspace{-2ex}
\end{figure}

To take advantage of hybrid multimodality and reduce computational costs, we propose to utilize the RGB loss to guide ray sampling in the event plane. As shown in Fig.~\ref{fig:prob_weight_sampling}, the algorithm starts by dividing the RGB image into $h \times w$ patches and randomly sampling N$_{c}$ rays from each patch to obtain the loss for each sample. Then, we calculate the average loss of each patch and project the center $\mathbf{m}_c$ to a downsampled mini-plan plane of the event camera:

\vspace{-2ex}
\begin{equation}\label{eq:prj}
    \resizebox{0.6\linewidth}{!}{
        \begin{math}
            \mathbf{m}=
            \begin{aligned}
                \frac{1}{Z_{e}} \mathbf{K}_{m}
                \begin{bmatrix}
                    \mathbf{I}_{3\times3}| \mathbf{0}_{3\times1}
                \end{bmatrix}
                \begin{bmatrix}
                    \mathbf{T}_{ec}\mathbf{K}^{-1}\mathbf{m}Z_c
                    \\
                    1
                \end{bmatrix},
            \end{aligned}
        \end{math}
    }
\end{equation}
\vspace{-2ex}

\noindent where $Z_c$ and $Z_e$ are the depths of two planes, $\mathbf{K}_m$ is the intrinsic of event mini-plane, and $\mathbf{T}_{ec}$ denotes the transformation between cameras. We apply the bilinear interpolation to compute the loss for each pixel in the mini-plan. Finally, the divided patches of the event plane query the loss $\{\mathcal{L}_{e}^{q}\}_{q=0}^{Q}$ from the mini-plane and sample rays with probability distribution $\boldsymbol f(j)=\frac{\mathcal{L}_e^{q}}{\sum_{q=1}^{Q} \mathcal{L}_e^{q}}$.

\subsubsection{Objective Functions} According to the EGM in~\cref{eq:EGM}, although it is not possible to directly model the luminance signals supervision, the logarithmic brightness differences $\hat{\mathbf{L}}(\mathbf{m}, t_\beta) - \hat{\mathbf{L}}(\mathbf{m}, t_\alpha)$ can be rendered from two camera poses $\mathbf{P}_\alpha$ and $\mathbf{P}_\beta$ with \cref{eq:render}. By integrating it with \cref{eq:EGM,eq:linlog}, we obtain:
\begin{equation}\label{eq:joint_event}
    \resizebox{0.88\linewidth}{!}{
        \begin{math}
            \begin{aligned}
                \mathbf{L}(\mathbf{m},t_\beta) - \mathbf{L}(\mathbf{m},t_\alpha) = \sum_{t_k=t_\alpha}^{t_k=t_\beta} p_k C \approx \hat{\mathbf{L}}(\mathbf{m}, t_\beta) - \hat{\mathbf{L}}(\mathbf{m}, t_\alpha).
            \end{aligned}
        \end{math}
    }
\end{equation}
Thus, we establish the relation between events and rendering, and define event reconstruction loss as:
\begin{equation}\label{eq:event_loss}
    \resizebox{0.85\linewidth}{!}{
        \begin{math}
            \begin{aligned}
                \mathcal{L}_{\text {ev}}\left(t_\beta, t_\alpha\right)=\operatorname{MSE}\left(
                \sum_{t_k=t_\alpha}^{t_k=t_\beta} p_k C - \hat{L}(\mathbf{m}, t_\beta) + \hat{L}(\mathbf{m}, t_\alpha)
                \right).
            \end{aligned}
        \end{math}
    }
\end{equation}
In our implementation, we perform a normalization on~\cref{eq:event_loss} to eliminate $C$ when it is unavailable.
The color and depth rendering losses~\cite{Wang2023CoSLAMJC} in a valid ray batch $R$ between the rendering and observations are also utilized:
\begin{equation}\label{eq:rgb_depth_loss}
    \resizebox{0.88\linewidth}{!}{
        \begin{math}
            \begin{aligned}
                \mathcal{L}_{\text{rgb}} =\frac{1}{\left | R \right | }\sum_{\mathbf{r} \in \mathbf{R}}(\hat{\mathbf{c}}(\mathbf{r}, \Delta t_c)) - c(\mathbf{r}))^{2},
                \mathcal{L}_{\text {d}} =\frac{1}{\left | R \right | }\sum_{\mathbf{r} \in \mathbf{R}}(\hat{d}(\mathbf{r}) - d(\mathbf{r}))^{2},
            \end{aligned}
        \end{math}
    }
\end{equation}
\noindent where $c(\mathbf{r})$ and $d(\mathbf{r})$ are the ground truth color and depth. To achieve an accurate geometry reconstruction, we apply the approximated SDF loss and free-space loss~\cite{Wang2023CoSLAMJC} to sampled point $\mathbf{x}$ near the surface ($S_{r}^{tr} = \{\mathbf{x}~|~|d(\mathbf{r}) - d(\mathbf{x})| \le tr\}$) and far from the surface ($S_{r}^{fs} = \{\mathbf{x}~|~|d(\mathbf{r}) - d(\mathbf{x})| > tr\}$):

\vspace{-3ex}
\begin{equation}\label{eq:loss_fn}
    \resizebox{0.8\linewidth}{!}{
        \begin{math}
            \begin{aligned}
                \mathcal{L}_{s d f} & =\frac{1}{\left|R\right|} \sum_{r \in R} \frac{1}{\left|S_{r}^{tr}\right|} \sum_{\mathbf{x} \in S_{r}^{tr}}\left(\mathbf{x}-(\hat{d}(\mathbf{r}) - d(\mathbf{r}))\right)^{2}, \\
                \mathcal{L}_{fs}    & =\frac{1}{\left|R\right|} \sum_{r \in R} \frac{1}{\left|S_{r}^{f s}\right|} \sum_{\mathbf{x} \in S_{r}^{fs}}\left(\mathbf{x}-tr\right)^{2} .
            \end{aligned}
        \end{math}
    }
\end{equation}
\vspace{-5ex}
\section{Dataset}
\label{sec:dataset}
\begin{figure*}[t]
    \vspace{-5ex}
    \centering
    {\footnotesize
        \setlength{\tabcolsep}{0pt}
        \renewcommand{\arraystretch}{0.5}
        \newcommand{\sz}{0.058}
        \begin{tabular}{cccccccc}
                                                           & Scene Model                                                                    & Norm Sequence                                                                       & Motion Blur Sequence                                                                & Dark Sequence                                                                       & \multicolumn{2}{c}{Scene with Motion Blur and Lighting Variation}                  &                                                                                                                                      \\
            \rotatebox[origin=c]{90}{\texttt{\#Room}}      & \includegraphics[valign=c,height=\sz\linewidth]{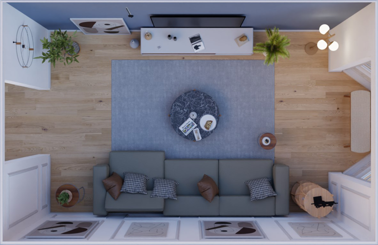}      & \includegraphics[valign=c,height=\sz\linewidth]{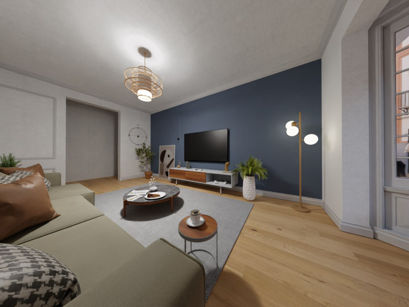}      & \includegraphics[valign=c,height=\sz\linewidth]{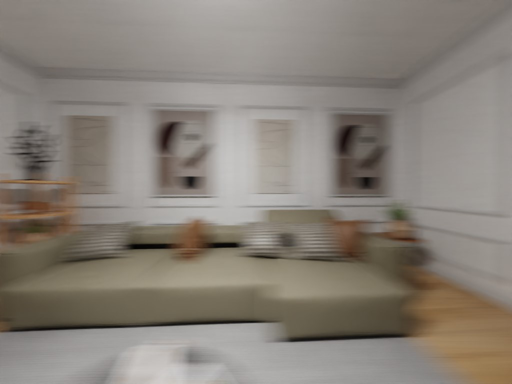}      & \includegraphics[valign=c,height=\sz\linewidth]{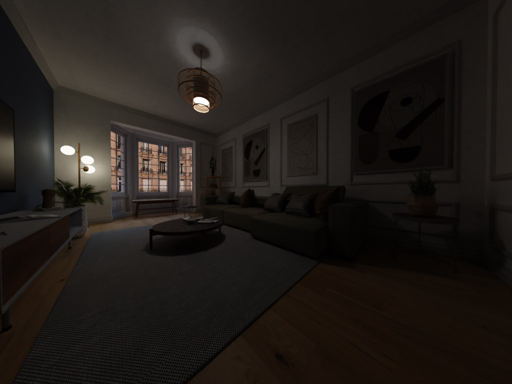}      & \includegraphics[valign=c,height=\sz\linewidth]{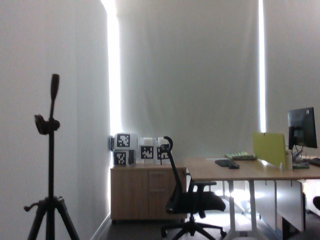} & \includegraphics[valign=c,height=\sz\linewidth]{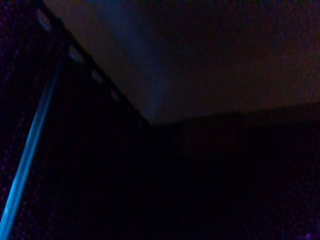} & \rotatebox[origin=c]{-90}{\texttt{\#Piofice}}   \\
            \rotatebox[origin=c]{90}{\texttt{\#Apartment}} & \includegraphics[valign=c,height=\sz\linewidth]{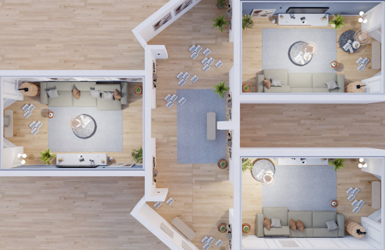} & \includegraphics[valign=c,height=\sz\linewidth]{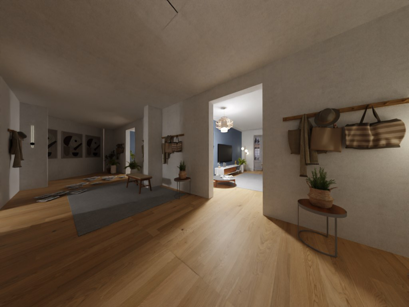} & \includegraphics[valign=c,height=\sz\linewidth]{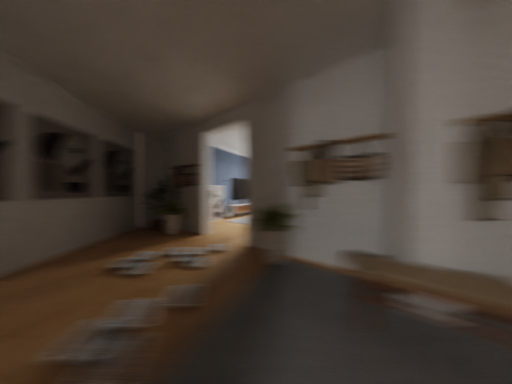} & \includegraphics[valign=c,height=\sz\linewidth]{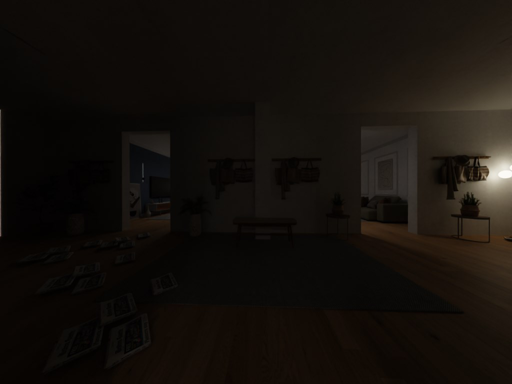} & \includegraphics[valign=c,height=\sz\linewidth]{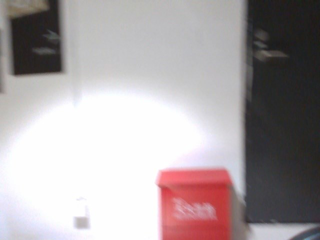}   & \includegraphics[valign=c,height=\sz\linewidth]{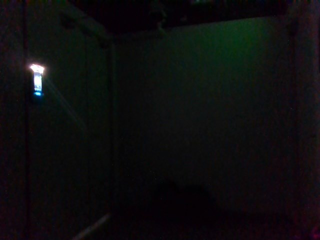}   & \rotatebox[origin=c]{-90}{\texttt{\#Garage}}    \\
            \rotatebox[origin=c]{90}{\texttt{\#Workshop}}  & \includegraphics[valign=c,height=\sz\linewidth]{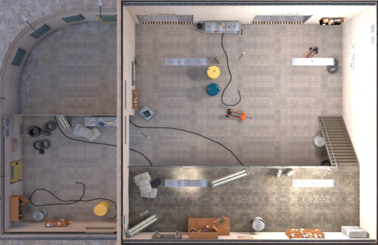} & \includegraphics[valign=c,height=\sz\linewidth]{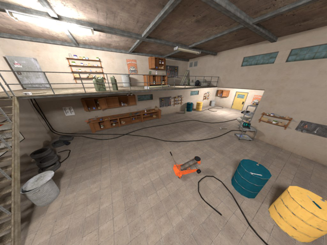} & \includegraphics[valign=c,height=\sz\linewidth]{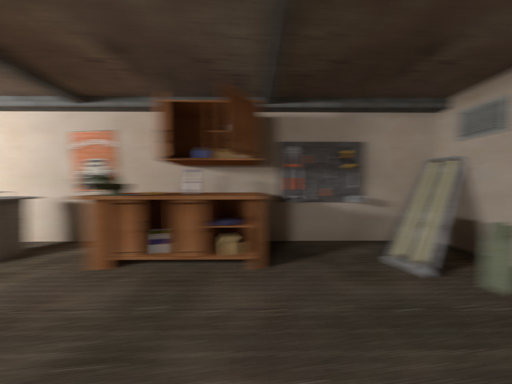} & \includegraphics[valign=c,height=\sz\linewidth]{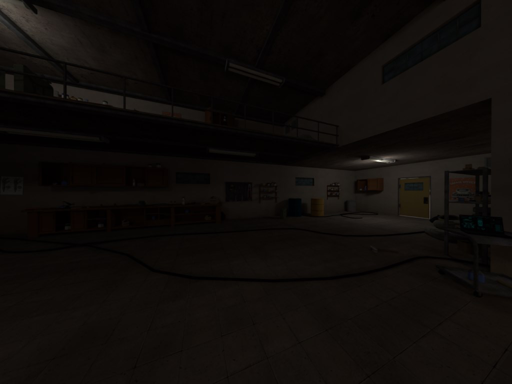} & \includegraphics[valign=c,height=\sz\linewidth]{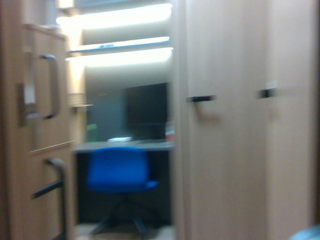}     & \includegraphics[valign=c,height=\sz\linewidth]{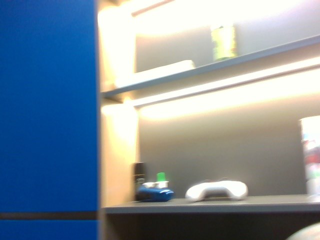}   & \rotatebox[origin=c]{-90}{\texttt{\#Dormitory}} \\
                                                           & \multicolumn{4}{c}{\cellcolor{green!10} DEV-Indoors}                            & \multicolumn{2}{c}{\cellcolor{red!10} DEV-Reals}                                                 &
        \end{tabular}
    }
    \caption{\textbf{Overview of the DEV-Indoors and DEV-Reals datasets}. \textbf{DEV-Indoors} is obtained through Blender~\cite{blender} and simulator~\cite{Gehrig_2020_CVPR}, covering normal, motion blur, and dark scenes, providing 9 subsets with RGB images, depth maps, event streams, meshes, and trajectories. \textbf{DEV-Reals} is captured from real scenes, providing 8 challenging subsets under motion blur and lighting variation.}
    \label{fig:dataset}
    \vspace{-2ex}
\end{figure*}
To our knowledge, there is currently no SLAM dataset that satisfactorily tackles challenges posed by strong motion blur and lighting variations while encompassing RGBD and event streams for NeRF-based SLAM. Common datasets lack depth~\cite{EDS} or ground truth meshes~\cite{Weikersdorfer2014Eventbased3S}. Additionally, they are primarily focused on outdoor scenes~\cite{M2DGR}, without significant motion blur~\cite{Bryner19icra} or lighting variation~\cite{InteriorNet18}. Therefore, in this paper, we construct simulated Dynamic Event RGBD Indoor (DEV-Indoors) and Dynamic Event RGBD Real captured (DEV-Reals) datasets, as shown in~\cref{fig:dataset}. Besides, we use Vector~\cite{Gao2022VECtorAV} dataset for evaluation as well. 

\noindent \textbf{1) DEV-Indoors} is rendered from 3 Blender~\cite{blender} models: \texttt{\#room}, \texttt{\#apartment}, and \texttt{\#workshop}. We generated 9 subsets containing high-quality color images, depth, meshes, and ground truth trajectories by varying the scene lighting and camera exposure time. The events are further generated via the events simulator~\cite{Gehrig_2020_CVPR}.

\noindent \textbf{2) Dev-Reals} is captured from 3 real scenes: \texttt{\#Pioffice}, \texttt{\#Garage} and \texttt{\#Dormitory}. Our capture system comprises a LiDAR (for ground truth pose), a Realsense D435I RGBD camera, and a DAVIS346 event camera. Eight sub-sequences are captured by modifying the lighting conditions and camera movement speed in the environment.
\section{Experiment}
\label{sec:experiment}
\noindent\textbf{Baselines.}
To the best of our knowledge, there is currently no event-based RGBD dense vSLAM with available public code that can be directly compared with our method. We opt EVO~\cite{rebecq2016evo}, ESVO~\cite{ESVO}, USLAM~\cite{vidal2018ultimate} as a reference from the most relevant event-based methods~\cite{gallego2017event,censi2014low,kueng2016low,Weikersdorfer2014Eventbased3S,EDS,zuo2022devo}. We also compare our method with the existing SOTA NeRF-based methods: iMAP~\cite{Sucar2021iMAPIM}, NICE-SLAM~\cite{Zhu2021NICESLAMNI}, CoSLAM~\cite{Wang2023CoSLAMJC}, and ESLAM~\cite{Johari2022ESLAMED}.

\noindent\textbf{Metric.} We use the absolute trajectory error~\cite{Sturm2012ASystems} (ATE) (cm) to measure the localization accuracy. For map reconstruction, we use the 2D Depth L1 (cm)~\cite{Zhu2021NICESLAMNI}, 3D accuracy (cm), completion (cm), and completion ratio (\%) to measure the scene geometry with mesh culling~\cite{azinovic2022neural,Johari2022ESLAMED}. The evaluation datasets are generated by randomly conducting 2000 poses and depths in Blender~\cite{blender}. We run all the methods 5 times and report the average results or \blackx + tracking success ratio if a method crashes.

\noindent \textbf{Implementation Details.} \ours is implemented in Python and trained on a desktop PC with a 5.50GHz Intel Core i9-13900K CPU and NVIDIA RTX 4090 GPU. We run \ours at 17 FPS and sample 1024 and 2048 rays in tracking and BA stages with 10 iterations by default. The event joint global BA is performed every 5 frames with 5\% of pixels from all keyframes. The model is trained using Adam optimizer with learning rate $lr_{rot} = 1e^{-3}, lr_{trans} = 1e^{-3}$, and loss weights $\lambda_{ev} = 0.05, \lambda_{rgb} = 5.0, \lambda_{d} = 0.1$. Default window $w_d$ and neighborhood window are set as 5 and 2, respectively. The exposures of RGB and event cameras are $5.21e^{-5}$ in DEV-Indoors. We use two sigmoid functions to fit the exposures if they are unavailable. Detailed settings can be found in the supplemental materials.
\subsection{Evaluation of Tracking and Mapping}
\label{sec:evaluation}
\begin{table}[t]
    \vspace{1.75ex}
    \centering
    \footnotesize
    \caption{\textbf{Tracking (ATE RSME [cm]) comparison on DEV-Indoors}. Our method outperforms previous works, demonstrating its robustness under motion blur and luminance variation.}
    \label{tab:tracking_devindoors}
    \setlength{\tabcolsep}{2pt}
    \renewcommand{\arraystretch}{0.8}
    \resizebox{\columnwidth}{!}{
        \begin{tabular}{rllllllllll}
            \toprule
            Method                             & \begin{tabular}[c]{@{}c@{}}\texttt{\#Rm}\\ \textbf{\texttt{norm}}\end{tabular} & \begin{tabular}[c]{@{}c@{}}\texttt{\#Rm}\\ \textbf{\texttt{blur}}\end{tabular} & \begin{tabular}[c]{@{}c@{}}\texttt{\#Rm}\\ \textbf{\texttt{dark}}\end{tabular} & \begin{tabular}[c]{@{}c@{}}\texttt{\#Apt}\\ \textbf{\texttt{norm}}\end{tabular} & \begin{tabular}[c]{@{}c@{}}\texttt{\#Apt}\\ \textbf{\texttt{blur}}\end{tabular} & \begin{tabular}[c]{@{}c@{}}\texttt{\#Apt}\\ \textbf{\texttt{dark}}\end{tabular} & \begin{tabular}[c]{@{}c@{}}\texttt{\#Wkp}\\ \textbf{\texttt{norm}}\end{tabular} & \begin{tabular}[c]{@{}c@{}}\texttt{\#Wkp}\\ \textbf{\texttt{blur}}\end{tabular} & \begin{tabular}[c]{@{}c@{}}\texttt{\#Wkp}\\ \textbf{\texttt{dark}}\end{tabular} & \begin{tabular}[c]{@{}c@{}}\texttt{\#all}\\ \textbf{\texttt{avg}}\end{tabular} \\
            \midrule
            iMAP~\cite{Sucar2021iMAPIM}        & 41.08                                                                                  & 50.58                                                                                  & 70.77                                                                                  & 25.75                                                                                    & 14.41                                                                                    & 1.06$e^5$                                                                                & 276.91                                                                                   & 891.86                                                                                   & 345.21                                                                                   & 214.57                                                                                   \\
            \midrule
            NICE-SLAM~\cite{Zhu2021NICESLAMNI} & 17.06                                                                                  & 29.54                                                                                  & 30.53                                                                                  & 25.17                                                                                    & 44.22                                                                                    & 48.28                                                                                    & \blackx 94 \%                                                                            & \blackx 33 \%                                                                            & \blackx 33\%                                                                             & 32.47                                                                                    \\
            \midrule
            CoSLAM\cite{Wang2023CoSLAMJC}      & \nd 10.71                                                                              & \nd 10.88                                                                              & \nd 26.64                                                                              & 10.02                                                                                    & 13.03                                                                                    & 30.75                                                                                    & 7.96                                                                                     & \nd 14.37                                                                                & 17.88                                                                                    & 15.80                                                                                    \\
            \midrule
            ESLAM~\cite{Johari2022ESLAMED}     & 10.72                                                                                  & 15.55                                                                                  & 40.42                                                                                  & \nd 9.99                                                                                 & \nd 12.79                                                                                & \nd 12.39                                                                                & \nd 7.01                                                                                 & 15.07                                                                                    & \nd 7.97                                                                                 & \nd 14.66                                                                                \\
            \midrule
            Ours                               & \fs \textbf{9.62}                                                                      & \fs \textbf{9.72}                                                                      & \fs \textbf{9.94}                                                                      & \fs \textbf{8.62}                                                                        & \fs \textbf{8.77}                                                                        & \fs \textbf{9.21}                                                                        & \fs \textbf{6.74}                                                                        & \fs \textbf{7.51}                                                                        & \fs \textbf{6.94}                                                                        & \fs \textbf{8.56}                                                                        \\
            \bottomrule
        \end{tabular}
    }
    \vspace{-4ex}
\end{table}
\begin{figure*}[t]
    \vspace{-5.5ex}
    \centering
    {\footnotesize
        \setlength{\tabcolsep}{0.2pt}
        \renewcommand{\arraystretch}{0}
        \newcommand{\sz}{0.141}
        \begin{tabular}{cccccccc}
                                                                  & RGB Input                                                                                 & Event Input                                                                            & NICE-SLAM~\cite{Zhu2021NICESLAMNI}                                                           & COSLAM~\cite{Wang2023CoSLAMJC}                                                             & ESLAM~\cite{Johari2022ESLAMED}                                                            & Ours                                                                                     & Ground Truth                                                                           \\
            \rotatebox[origin=c]{90}{{\tiny \texttt{\#Rm~blur}}}  & \includegraphics[valign=c,width=\sz\linewidth]{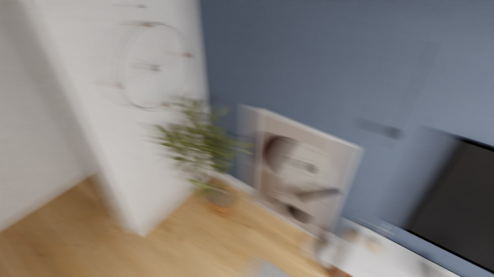} & \includegraphics[valign=c,width=\sz\linewidth]{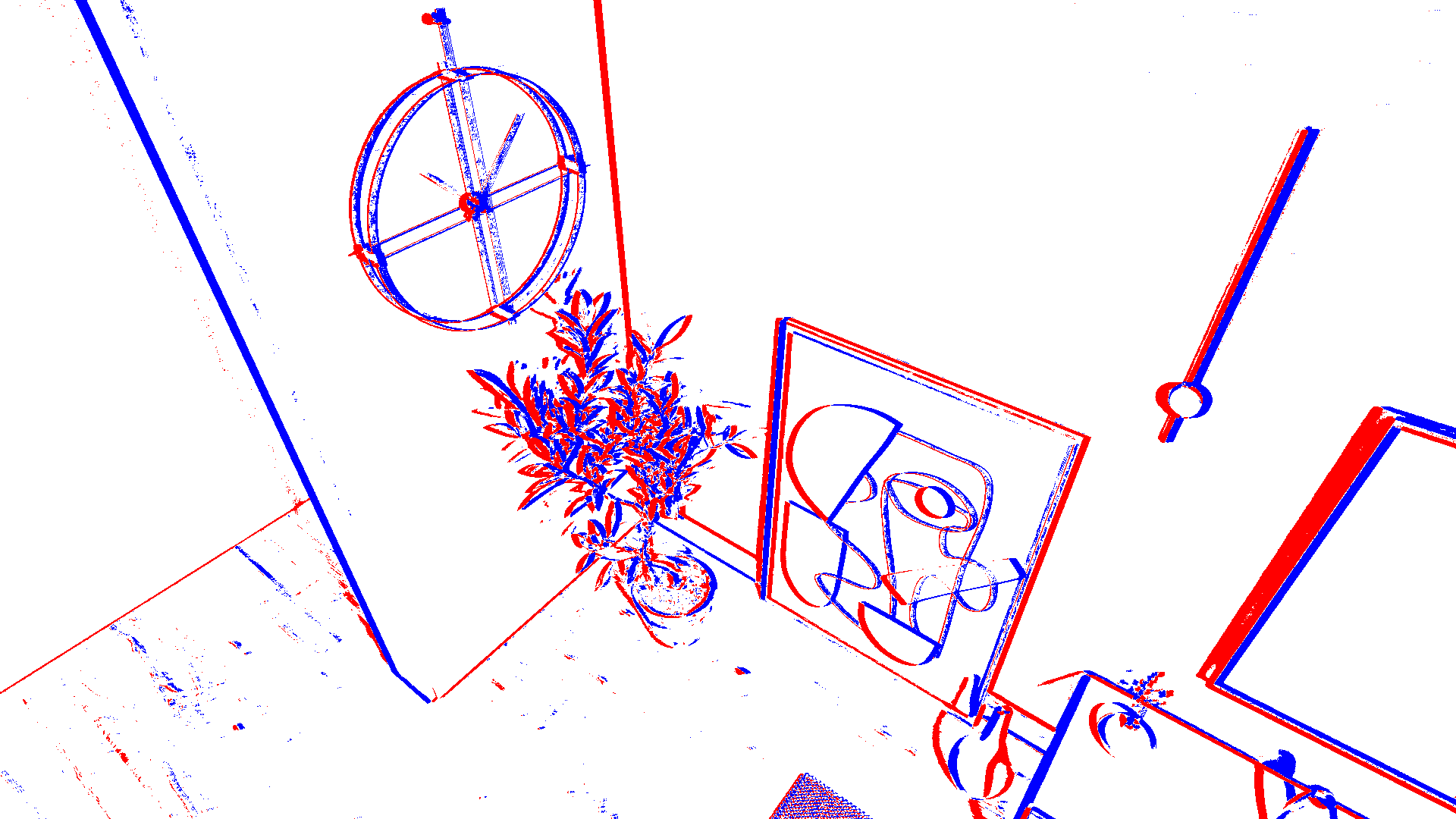} & \includegraphics[valign=c,width=\sz\linewidth]{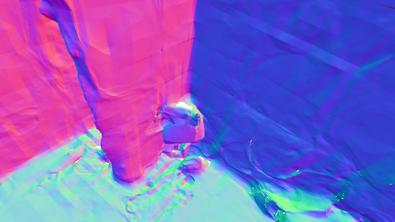} & \includegraphics[valign=c,width=\sz\linewidth]{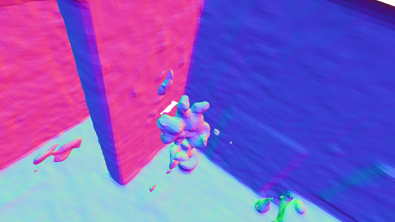} & \includegraphics[valign=c,width=\sz\linewidth]{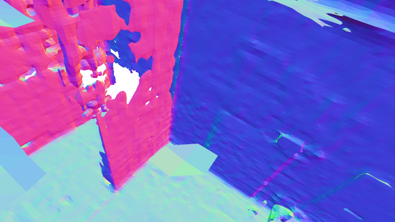} & \includegraphics[valign=c,width=\sz\linewidth]{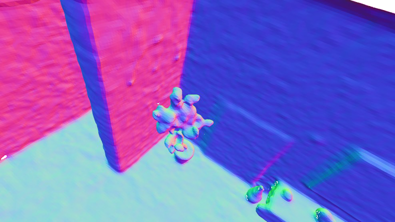} & \includegraphics[valign=c,width=\sz\linewidth]{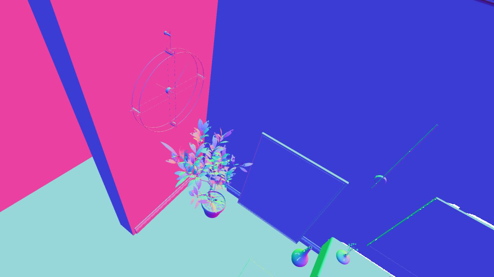} \\
            \rotatebox[origin=c]{90}{{\tiny \texttt{\#Wkp Blur}}} & \includegraphics[valign=c,width=\sz\linewidth]{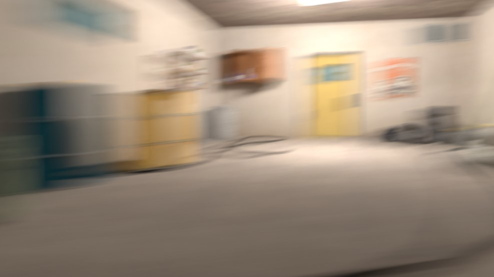}  & \includegraphics[valign=c,width=\sz\linewidth]{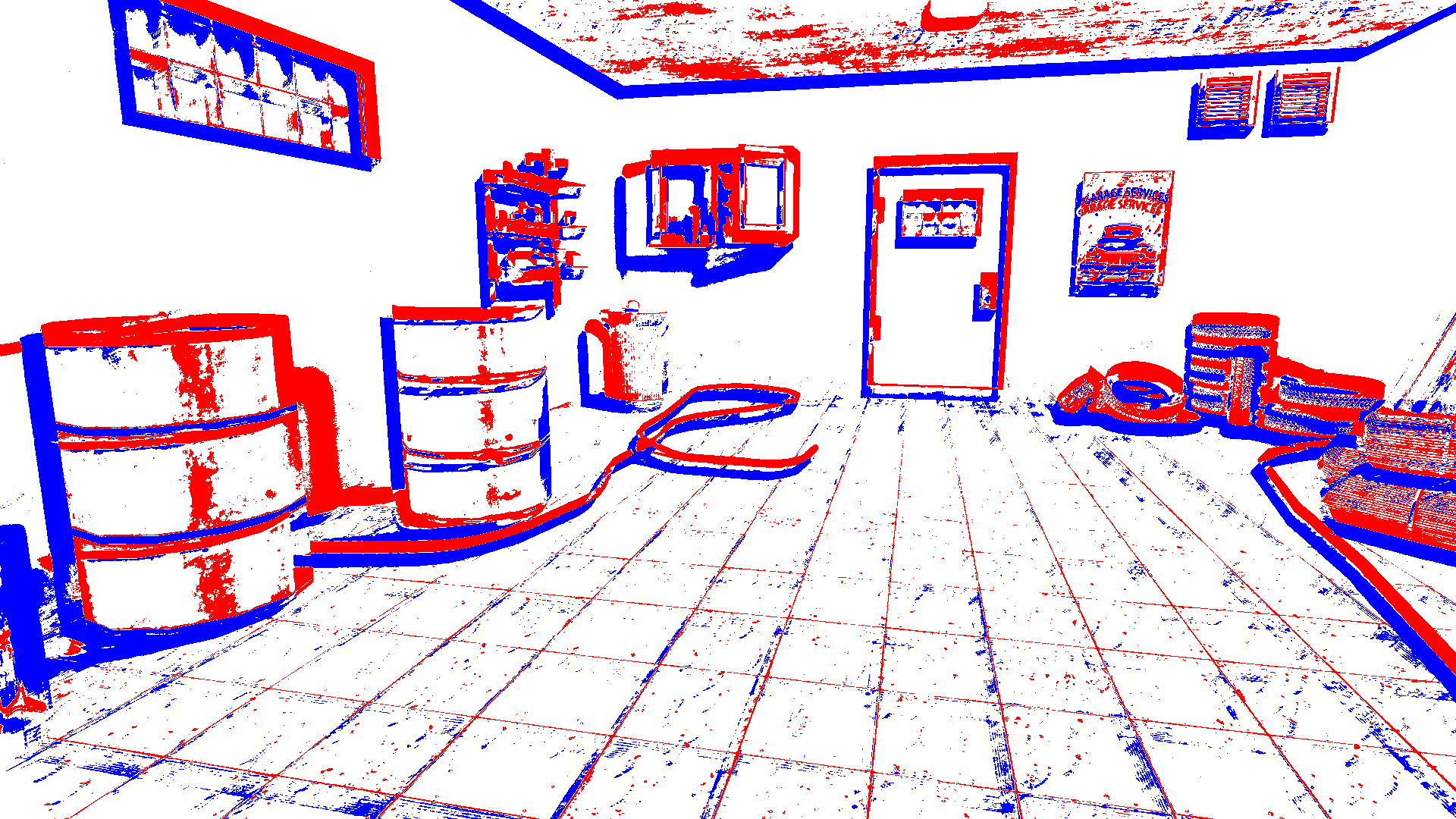} & \includegraphics[valign=c,width=\sz\linewidth]{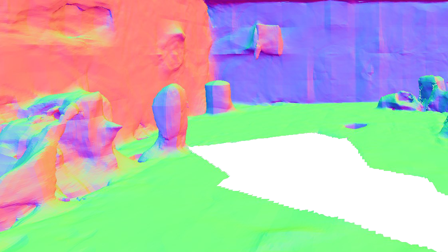} & \includegraphics[valign=c,width=\sz\linewidth]{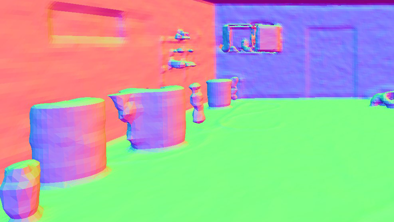} & \includegraphics[valign=c,width=\sz\linewidth]{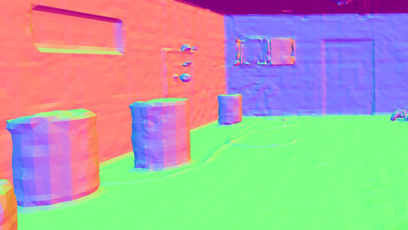} & \includegraphics[valign=c,width=\sz\linewidth]{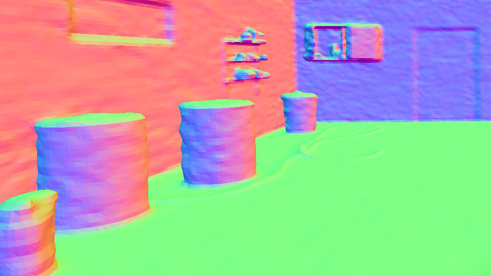} & \includegraphics[valign=c,width=\sz\linewidth]{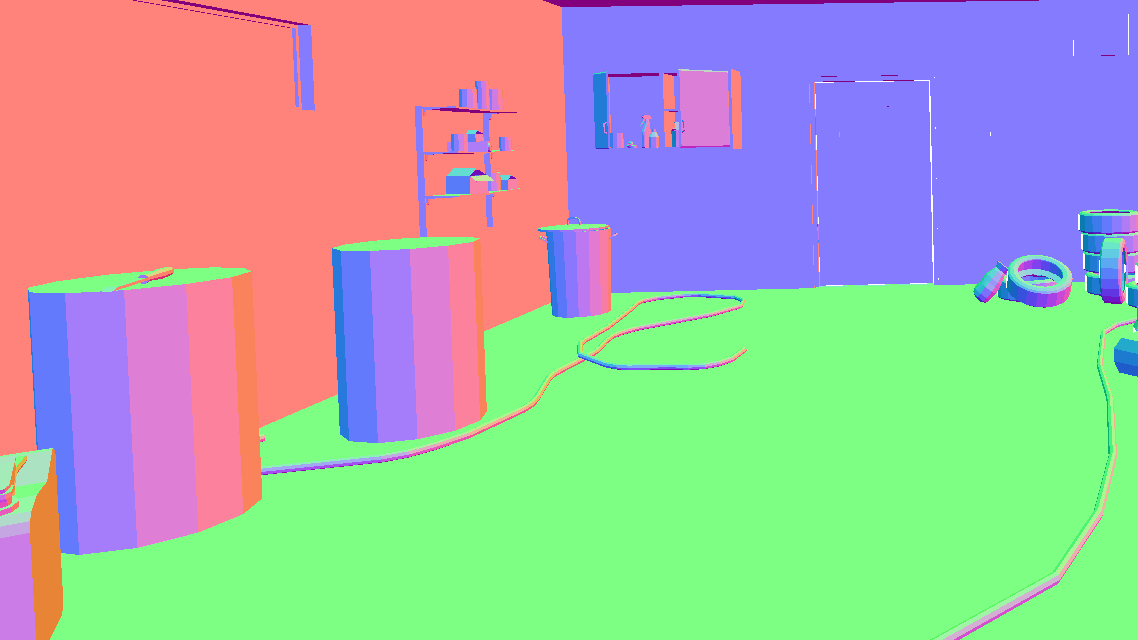} \\
            \rotatebox[origin=c]{90}{{\tiny \texttt{\#Wkp Dark}}} & \includegraphics[valign=c,width=\sz\linewidth]{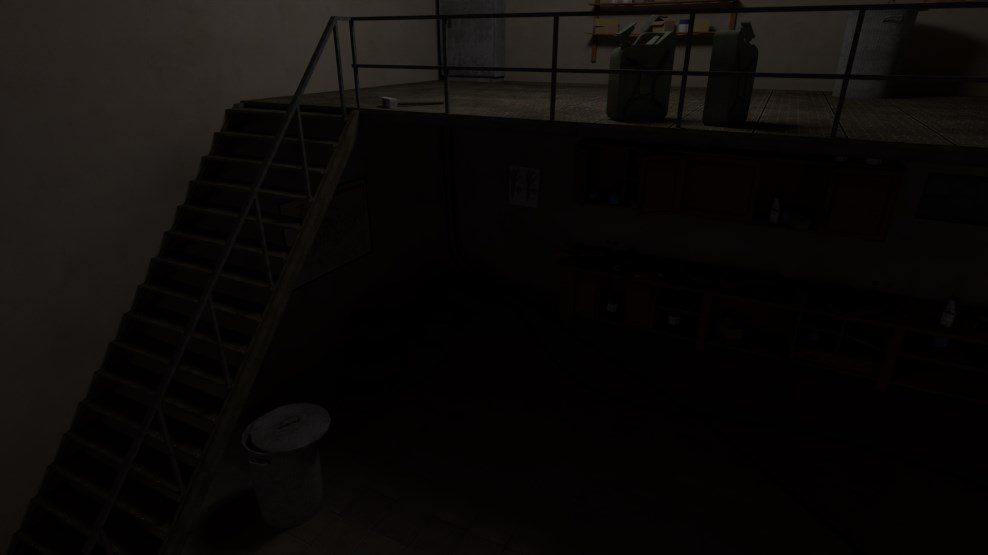}  & \includegraphics[valign=c,width=\sz\linewidth]{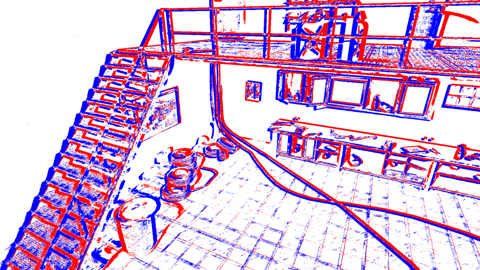} & \includegraphics[valign=c,width=\sz\linewidth]{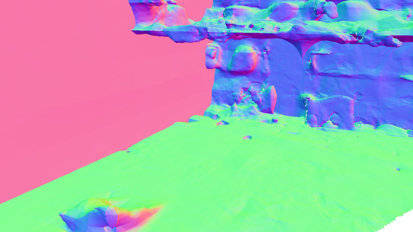} & \includegraphics[valign=c,width=\sz\linewidth]{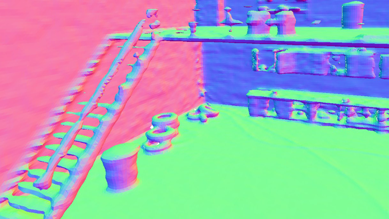} & \includegraphics[valign=c,width=\sz\linewidth]{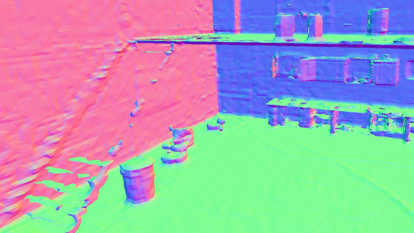} & \includegraphics[valign=c,width=\sz\linewidth]{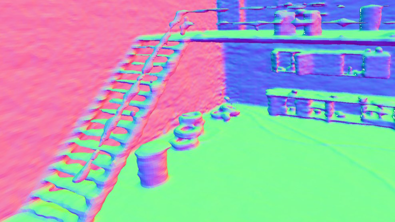} & \includegraphics[valign=c,width=\sz\linewidth]{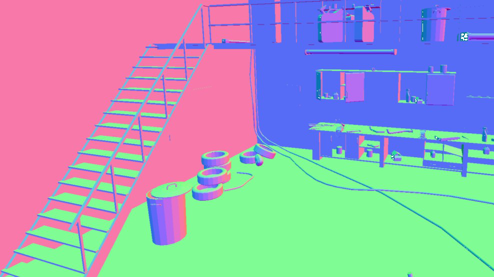} \\                                                                                                                                                                                                                                                                                                                                                                            \\
        \end{tabular}
    }
    \caption{\textbf{Reconstruction Performance on DEV-Indoors.} \ours achieves, on average, more precise reconstruction details than existing methods in motion blur and lighting varying environments with the assistance of high-quality event streams.}
    \label{fig:mesh_devindoors}
    \vspace{-1ex}
\end{figure*}

\textbf{Evaluation on DEV-Indoors.} We report the trajectory accuracy and reconstruction quality in Tab.~\ref{tab:tracking_devindoors} and Tab.~\ref{tab:reconstruction_devindoors}. As shown in Tab.~\ref{tab:tracking_devindoors}, \ours performs the best in all 9 scenes and remains stable (fluctuation below 0.8) under all the blur and dark sequences. While the others perform sensitive and unstable when facing non-ideal environments, especially iMAP~\cite{Sucar2021iMAPIM} and NICE-SLAM~\cite{Zhu2021NICESLAMNI}, which completely crushes in the blur and dark scenes. ESLAM~\cite{Zhu2021NICESLAMNI} and CoSLAM~\cite{Wang2023CoSLAMJC} exhibit similar trends but achieve relatively stable results. Specifically, in \texttt{\#room} sequences, they achieved 1.02 and 1.45 times the error on the blur subset and 2.49 and 3.77 times the error on the dark subset, respectively. The \textbf{reconstruction quality} in \cref{tab:reconstruction_devindoors,fig:mesh_devindoors} show that \ours performs more accurately and robustly than the other methods. Specifically, our method reduces the error by 2.37, 0.48, and 1.90 in ACC, Comp, and Depth L1 compared with the second ESLAM~\cite{Johari2022ESLAMED}. \cref{fig:mesh_devindoors} also shows that our method reconstructs the details of the scenes more accurately and produces fewer artifacts.

\begin{table}[t]
    \centering
    \footnotesize
    \caption{\textbf{Reconstruction Performance [cm]} of the proposed method vs. the state-of-the-art methods on \textbf{DEV-Indoors} dataset.}
    \label{tab:reconstruction_devindoors}
    \setlength{\tabcolsep}{1pt}
    \renewcommand{\arraystretch}{0.8}
    \resizebox{\columnwidth}{!}{
        \begin{tabular}{clcccccccccc}
            \toprule
            Method                                                                                       & Metric               & \begin{tabular}[c]{@{}c@{}}\texttt{\#Rm}\\ \textbf{\texttt{norm}}\end{tabular} & \begin{tabular}[c]{@{}c@{}}\texttt{\#Rm}\\ \textbf{\texttt{blur}}\end{tabular} & \begin{tabular}[c]{@{}c@{}}\texttt{\#Rm}\\ \textbf{\texttt{dark}}\end{tabular} & \begin{tabular}[c]{@{}c@{}}\texttt{\#Apt}\\ \textbf{\texttt{norm}}\end{tabular} & \begin{tabular}[c]{@{}c@{}}\texttt{\#Apt}\\ \textbf{\texttt{blur}}\end{tabular} & \begin{tabular}[c]{@{}c@{}}\texttt{\#Apt}\\ \textbf{\texttt{dark}}\end{tabular} & \begin{tabular}[c]{@{}c@{}}\texttt{\#Wkp}\\ \textbf{\texttt{norm}}\end{tabular} & \begin{tabular}[c]{@{}c@{}}\texttt{\#Wkp}\\ \textbf{\texttt{blur}}\end{tabular} & \begin{tabular}[c]{@{}c@{}}\texttt{\#Wkp}\\ \textbf{\texttt{dark}}\end{tabular} & \begin{tabular}[c]{@{}c@{}}\texttt{\#all}\\ \textbf{\texttt{avg}}\end{tabular} \\
            \midrule
            \multirow{4}{*}{\begin{tabular}[l]{@{}c@{}}iMAP\\~\cite{Sucar2021iMAPIM}\end{tabular}}       & Acc$\downarrow$      & 37.16                                                                                   & 37.60                                                                                   & 34.30                                                                                   & 25.55                                                                                    & 44.288                                                                                   & 54.47                                                                                    & 56.40                                                                                    & 51.74                                                                                    & 38.12                                                                                    & 42.18                                                                                    \\
                                                                                                         & Comp$\downarrow$     & 48.69                                                                                   & 55.97                                                                                   & 33.76                                                                                   & 19.28                                                                                    & 23.94                                                                                    & 49.61                                                                                    & 65.40                                                                                    & 27.18                                                                                    & 47.11                                                                                    & 41.22                                                                                    \\
                                                                                                         & Comp Ratio$\uparrow$ & 37.56                                                                                   & 37.76                                                                                   & 39.11                                                                                   & 60.03                                                                                    & 46.27                                                                                    & 42.56                                                                                    & 12.14                                                                                    & 48.44                                                                                    & 40.41                                                                                    & 40.48                                                                                    \\
                                                                                                         & Depth L1$\downarrow$ & 79.93                                                                                   & 97.98                                                                                   & 64.96                                                                                   & 40.36                                                                                    & 128.45                                                                                   & 140.00                                                                                   & 131.79                                                                                   & 115.07                                                                                   & 124.69                                                                                   & 102.58                                                                                   \\
            \midrule
            \multirow{4}{*}{\begin{tabular}[l]{@{}c@{}}NICESL\\AM~\cite{Zhu2021NICESLAMNI}\end{tabular}} & Acc$\downarrow$      & 18.49                                                                                   & 18.86                                                                                   & 16.69                                                                                   & 21.27                                                                                    & 16.51                                                                                    & 19.17                                                                                    & 26.35                                                                                    & 22.09                                                                                    & 28.40                                                                                    & 20.87                                                                                    \\
                                                                                                         & Comp$\downarrow$     & 20.26                                                                                   & 21.93                                                                                   & 21.43                                                                                   & 20.67                                                                                    & 18.70                                                                                    & 21.29                                                                                    & 28.04                                                                                    & 49.82                                                                                    & 77.19                                                                                    & 31.04                                                                                    \\
                                                                                                         & CompRatio $\uparrow$ & 60.40                                                                                   & 59.03                                                                                   & 58.14                                                                                   & 59.60                                                                                    & 63.49                                                                                    & 56.55                                                                                    & 50.91                                                                                    & 46.00                                                                                    & 35.58                                                                                    & 54.41                                                                                    \\
                                                                                                         & Depth L1$\downarrow$ & 40.59                                                                                   & 42.09                                                                                   & 40.26                                                                                   & 41.54                                                                                    & 24.00                                                                                    & 34.89                                                                                    & 62.48                                                                                    & 104.20                                                                                   & 106.09                                                                                   & 55.13                                                                                    \\
            \midrule
            \multirow{4}{*}{\begin{tabular}[l]{@{}c@{}}CoSLA\\M~\cite{Wang2023CoSLAMJC}\end{tabular}}    & Acc$\downarrow$      & 10.66                                                                                   & 11.36                                                                                   & 12.77                                                                                   & 15.47                                                                                    & 16.42                                                                                    & 30.71                                                                                    & 13.02                                                                                    & 17.59                                                                                    & 19.85                                                                                    & 16.43                                                                                    \\
                                                                                                         & Comp$\downarrow$     & 13.24                                                                                   & \nd 12.44                                                                               & 12.23                                                                                   & 14.09                                                                                    & 15.36                                                                                    & 21.67                                                                                    & 13.92                                                                                    & 18.26                                                                                    & 19.46                                                                                    & 15.63                                                                                    \\
                                                                                                         & Comp Ratio$\uparrow$ & 69.22                                                                                   & \nd 76.87                                                                               & 77.26                                                                                   & 70.61                                                                                    & 66.75                                                                                    & 55.26                                                                                    & 67.92                                                                                    & 61.26                                                                                    & 60.70                                                                                    & 67.3                                                                                     \\
                                                                                                         & Depth L1$\downarrow$ & 24.78                                                                                   & \nd 20.91                                                                               & 20.65                                                                                   & 32.29                                                                                    & 35.90                                                                                    & 64.14                                                                                    & \nd 28.69                                                                                & 39.17                                                                                    & 42.85                                                                                    & 34.38                                                                                    \\
            \midrule
            \multirow{4}{*}{\begin{tabular}[l]{@{}c@{}}ESLA\\M~\cite{Johari2022ESLAMED}\end{tabular}}    & Acc$\downarrow$      & \nd 9.48                                                                                & \fs \textbf{8.58}                                                                       & \nd 11.81                                                                               & \nd 12.86                                                                                & \nd 16.25                                                                                & \nd 14.85                                                                                & \fs \textbf{9.01}                                                                        & \nd 10.01                                                                                & \nd 10.02                                                                                & \nd 11.43                                                                                \\
                                                                                                         & Comp$\downarrow$     & \nd 7.94                                                                                & \fs \textbf{7.54}                                                                       & \fs \textbf{7.08}                                                                       & \nd 8.511                                                                                & \nd 12.37                                                                                & \nd 10.40                                                                                & \fs \textbf{8.89}                                                                        & \nd 10.95                                                                                & \nd 10.44                                                                                & \fs \textbf{9.35}                                                                        \\
                                                                                                         & Comp Ratio$\uparrow$ & \fs \textbf{84.60}                                                                      & \fs \textbf{85.69}                                                                      & \fs \textbf{86.70}                                                                      & \nd 82.93                                                                                & \nd 71.53                                                                                & \nd 80.90                                                                                & \fs \textbf{83.17}                                                                       & \nd 80.02                                                                                & \nd 81.64                                                                                & \nd 81.91                                                                                \\
                                                                                                         & Depth L1$\downarrow$ & \nd 15.34                                                                               & \fs \textbf{12.27}                                                                      & \nd 19.08                                                                               & \nd 11.07                                                                                & \nd 27.80                                                                                & \nd 15.50                                                                                & 30.03                                                                                    & \nd 29.06                                                                                & \nd 28.02                                                                                & \nd 20.91                                                                                \\
            \midrule
            \rowcolor[HTML]{EFEFEF}
            \rowcolor[HTML]{EFEFEF}                                                                      & Acc$\downarrow$      & \fs \textbf{7.48}                                                                       & \nd 10.53                                                                               & \fs \textbf{7.07}                                                                       & \fs \textbf{9.46}                                                                        & \fs \textbf{9.91}                                                                        & \fs \textbf{9.34}                                                                        & \nd 9.23                                                                                 & \fs \textbf{9.28}                                                                        & \fs \textbf{9.25}                                                                        & \fs \textbf{9.06}                                                                        \\
            \rowcolor[HTML]{EFEFEF}
            \rowcolor[HTML]{EFEFEF}                                                                      & Comp$\downarrow$     & \fs \textbf{7.70}                                                                       & 12.51                                                                                   & \nd 7.70                                                                                & \fs \textbf{9.87}                                                                        & \fs \textbf{9.28}                                                                        & \fs \textbf{9.61}                                                                        & \nd 9.95                                                                                 & \fs \textbf{9.92}                                                                        & \fs \textbf{9.92}                                                                        & \nd 9.61                                                                                 \\
            \rowcolor[HTML]{EFEFEF}
            \rowcolor[HTML]{EFEFEF}                                                                      & Comp Ratio$\uparrow$ & \nd 83.00                                                                               & 74.48                                                                                   & \nd 84.26                                                                               & \fs \textbf{85.36}                                                                       & \fs \textbf{83.40}                                                                       & \fs \textbf{84.01}                                                                       & \nd 82.27                                                                                & \fs \textbf{82.38}                                                                       & \fs \textbf{82.35}                                                                       & \fs \textbf{82.39}                                                                       \\
            \rowcolor[HTML]{EFEFEF}
            \multirow{-4}{*}{\cellcolor[HTML]{EFEFEF} Ours}                                              & Depth L1$\downarrow$ & \fs \textbf{15.10}                                                                      & 23.36                                                                                   & \fs \textbf{11.92}                                                                      & \fs \textbf{19.86}                                                                       & \fs \textbf{11.94}                                                                       & \fs \textbf{19.21}                                                                       & \fs \textbf{23.16}                                                                       & \fs \textbf{23.14}                                                                       & \fs \textbf{23.39}                                                                       & \fs \textbf{19.01}                                                                       \\
            \bottomrule
        \end{tabular}
    }
\end{table}

\boldparagraph{Evaluation on DEV-Reals.} \cref{tab:tracking_devreals} illustrates the tracking performance of our method and the state-of-the-art methods on the DEV-Reals dataset. Our method achieves the best performance in all the scenes (18.97), and the average error is 1.63 times lower than the second-best CoSLAM~\cite{Wang2023CoSLAMJC} (30.95). Note that DEV-Reals is challenging due to the large motion and varying light, leading to the crushes of ORB-SLAM~\cite{MurArtal2016ORBSLAM2AO} and NICE-SLAM~\cite{Zhu2021NICESLAMNI}.

\begin{table}[t]
    \centering
    \caption{\textbf{Tracking comparison (ATE median [cm])} of the proposed method vs. the state-of-the-art methods on \textbf{DEV-Reals}.}
    \label{tab:tracking_devreals}
    \resizebox{\columnwidth}{!}{
        \begin{tabular}{rccccccccc}
            \toprule
            Method                             & \texttt{Pio1}     & \texttt{Pio2}      & \texttt{Gre1}      & \texttt{Gre2}      & \texttt{dorm1}     & \texttt{dorm2}     & \texttt{dorm3}     & \texttt{dorm4}     & \texttt{\textbf{avg}} \\
            \midrule
            ORBSLAM~\cite{EDS}                 & \blackx63\%       & \blackx63\%        & \blackx63\%        & \blackx63\%        & \blackx63\%        & \blackx63\%        & \blackx63\%        & \blackx63\%        & \blackx63\%           \\
            NICE-SLAM~\cite{Zhu2021NICESLAMNI} & 13.21             & 23.35              & \blackx63\%        & \blackx25\%        & 24.69              & \nd 10.68          & 18.44              & 44.04              & \blackx22.40          \\
            COSLAM~\cite{Wang2023CoSLAMJC}     & \nd 11.14         & \nd 19.83          & 82.52              & 40.16              & \nd 15.99          & 15.42              & 30.12              & \nd 32.45          & \nd 30.95             \\
            ESLAM~\cite{Johari2022ESLAMED}     & 11.28             & 21.42              & \nd 63.65          & \nd 30.75          & 37.94              & 31.04              & \nd 16.19          & 37.91              & 31.27                 \\
            \midrule
            Ours                               & \fs \textbf{8.94} & \fs \textbf{19.05} & \fs \textbf{43.63} & \fs \textbf{21.18} & \fs \textbf{11.26} & \fs \textbf{11.91} & \fs \textbf{16.00} & \fs \textbf{19.78} & \fs \textbf{18.97}    \\
            \bottomrule
        \end{tabular}
    }
    \vspace{-4ex}
\end{table}

\begin{table}[t]
    \centering
    \caption{\textbf{Tracking comparison (ATE mean [cm])} of the proposed method vs. the Event-based SLAM system on \textbf{Vector}\cite{Gao2022VECtorAV} dataset.}
    \label{tab:tracking_vector_details}
    \resizebox{\columnwidth}{!}{
        \begin{tabular}{rccccccccc}
            \toprule
            Method                                      & \begin{tabular}[c]{@{}c@{}}\texttt{robot}\\ \texttt{norm}\end{tabular} & \begin{tabular}[c]{@{}c@{}}\texttt{robot}\\ \texttt{fast}\end{tabular} & \begin{tabular}[c]{@{}c@{}}\texttt{desk}\\\texttt{norm}\end{tabular} & \begin{tabular}[c]{@{}c@{}}\texttt{desk}\\ \texttt{fast}\end{tabular} & \begin{tabular}[c]{@{}c@{}}\texttt{sofa}\\ \texttt{norm}\end{tabular} & \begin{tabular}[c]{@{}c@{}}\texttt{sofa}\\\texttt{fast}\end{tabular} & \begin{tabular}[c]{@{}c@{}}\texttt{hdr}\\\texttt{norm}\end{tabular} & \begin{tabular}[c]{@{}c@{}}\texttt{hdr}\\\texttt{fast}\end{tabular} & \begin{tabular}[c]{@{}c@{}}\texttt{\textbf{\#all}}\\\texttt{\textbf{avg}}\end{tabular} \\
            \midrule
            EVO~\cite{rebecq2016evo}                    & 3.25                                                                   & \blackx                                                                & \blackx                                                              & \blackx                                                               & \blackx                                                               & \blackx                                                              & \blackx                                                             & \blackx                                                             & 3.25                                                                                                       \\
            ESVO~\cite{ESVO}                            & \blackx                                                                & \blackx                                                                & \blackx                                                              & \blackx                                                               & \fs \textbf{1.77}                                                     & \blackx                                                              & \blackx                                                             & \blackx                                                             & \nd 1.77                                                                                                   \\
            USLAM~\cite{vidal2018ultimate}~\small(EVIO) & 1.18                                                                   & \fs \textbf{1.65}                                                      & 2.24                                                                 & \fs \textbf{1.08}                                                     & 5.74                                                                  & \nd 2.54                                                             & 5.69                                                                & 2.61                                                                & 2.84                                                                                                       \\
            \midrule
            CoSLAM~\cite{Wang2023CoSLAMJC}~{\small(DV)} & \fs \textbf{1.00}                                                      & 124.69                                                                 & \fs \textbf{1.76}                                                    & 97.65                                                                 & \nd 1.74                                                              & 77.89                                                                & \nd 1.47                                                            & \nd 1.42                                                            & 38.45                                                                                                      \\
            ESLAM~\cite{Johari2022ESLAMED}~{\small(DV)} & 1.39                                                                   & 3.30                                                                   & 2.54                                                                 & 3.64                                                                  & 7.99                                                                  & 19.03                                                                & 7.38                                                                & 12.23                                                               & 7.19                                                                                                       \\
            Ours~{\small(EDV)}                          & \nd 1.06                                                               & \nd 1.73                                                               & \fs \textbf{1.76}                                                    & \nd  2.69                                                             & 2.02                                                                  & \fs \textbf{1.84}                                                    & \fs 1.03                                                            & \fs 1.22                                                            & \fs \textbf{1.67}                                                                                          \\
            \bottomrule
        \end{tabular}
    }
    \footnotesize{\tiny E: event, V: RGB or gray image, D: depth, I: IMU, S: stereo, O: odometry, \blackx: crashes.}
\end{table}

\subsection{Runtime Analysis}
\label{sec:runtime}
We evaluate all the frameworks on an NVIDIA RTX 4090 GPU and report average tracking and mapping iterations spending, FPS, and parameters number of the model in~\cref{tab:runtime}. The experimental results indicate that our method is fast, with an average of 17 FPS, comparable to the currently most efficient ESLAM~\cite{Johari2022ESLAMED}. Meanwhile, our method remains lightweight, with only 1.95M parameters, yet achieves the best accuracy.

\begin{table}[t]
    \centering
    \footnotesize
    \caption{\textbf{Run-time comparison} on DEV-Indoors. \ours is comparable to the most efficient ESLAM and keeps lightweight.}
    \label{tab:runtime}
    \setlength{\tabcolsep}{10pt}
    \renewcommand{\arraystretch}{1}
    \resizebox{\columnwidth}{!}{
        \begin{tabular}{rcccc}
            \toprule
            Method                             & Tracking [ms$\times$it] $\downarrow$ & Mapping [ms$\times$it] $\downarrow$ & FPS $\uparrow$     & \#parama. \\
            \midrule
            iMAP~\cite{Sucar2021iMAPIM}        & 24.73$\times$50                      & 41.18$\times$300                    & 0.36               & 0.22 M    \\
            NICE-SLAM~\cite{Zhu2021NICESLAMNI} & 6.46$\times$16                       & 26.42$\times$120                    & 1.55               & 5.86 M    \\
            CoSLAM\cite{Wang2023CoSLAMJC}      & 6.08$\times$15                       & 13.52$\times$15                     & 11.26              & 1.71 M    \\
            ESLAM~\cite{Johari2022ESLAMED}     & 5.20$\times$13                       & 16.68$\times$10                     & \nd 14.77          & 7.85 M    \\
            Ours                               & 5.75$\times$10                       & 13.16$\times$10                     & \fs \textbf{17.40} & 1.95 M    \\
            \bottomrule
        \end{tabular}
    }
    \vspace{-5ex}
\end{table}

\subsection{Evaluation of Rendering}
\label{sec:rendering}
\begin{figure*}[t]
    \vspace{-6ex}
    \centering
    {\footnotesize
        \setlength{\tabcolsep}{7.5pt}
        \begin{tabular}{cc}
            \resizebox{0.3\linewidth}{!}
            {
                \begin{subfigure}{0.4\linewidth}
                    \centering
                    \footnotesize
                    \setlength{\tabcolsep}{0.55pt}
                    \renewcommand{\arraystretch}{1.3}
                    \begin{tabular}{clccccccccc}
                        \toprule
                        Method                                                                                       & Metric & \begin{tabular}[c]{@{}c@{}}\texttt{\#Rm}\\ \textbf{\texttt{norm}}\end{tabular} & \begin{tabular}[c]{@{}c@{}}\texttt{\#Rm}\\ \textbf{\texttt{blur}}\end{tabular} & \begin{tabular}[c]{@{}c@{}}\texttt{\#Rm}\\ \textbf{\texttt{dark}}\end{tabular} & \begin{tabular}[c]{@{}c@{}}\texttt{\#Apt}\\ \textbf{\texttt{norm}}\end{tabular} & \begin{tabular}[c]{@{}c@{}}\texttt{\#Apt}\\ \textbf{\texttt{blur}}\end{tabular} & \begin{tabular}[c]{@{}c@{}}\texttt{\#Apt}\\ \textbf{\texttt{dark}}\end{tabular} & \begin{tabular}[c]{@{}c@{}}\texttt{\#Wkp}\\ \textbf{\texttt{norm}}\end{tabular} & \begin{tabular}[c]{@{}c@{}}\texttt{\#Wkp}\\ \textbf{\texttt{blur}}\end{tabular} & \begin{tabular}[c]{@{}c@{}}\texttt{\#Wkp}\\ \textbf{\texttt{dark}}\end{tabular} \\
                        \midrule
                        \multirow{3}{*}{\begin{tabular}[l]{@{}c@{}}NICESL\\AM~\cite{Zhu2021NICESLAMNI}\end{tabular}} & PSNR   & 13.65                                                                                    & 18.24                                                                                    & 28.09                                                                                    & 14.23                                                                                     & 17.50                                                                                     & 28.37                                                                                     &                                                                                           &                                                                                           &                                                                                           \\
                                                                                                                     & SSIM   & 0.457                                                                                    & 0.623                                                                                    & 0.828                                                                                    & 0.445                                                                                     & 0.573                                                                                     & 0.853                                                                                     & \blackx                                                                                   & \blackx                                                                                   & \blackx                                                                                   \\
                                                                                                                     & LPIPS  & 0.646                                                                                    & 0.485                                                                                    & 0.349                                                                                    & 0.673                                                                                     & 0.552                                                                                     & 0.325                                                                                     &                                                                                           &                                                                                           &                                                                                           \\
                        \midrule
                        \multirow{3}{*}{\begin{tabular}[l]{@{}c@{}}CoSLA\\M~\cite{Wang2023CoSLAMJC}\end{tabular}}    & PSNR   & 23.16                                                                                    & 24.86                                                                                    & 31.22                                                                                    & 22.79                                                                                     & 23.85                                                                                     & \fs 32.45                                                                                 & \fs 24.12                                                                                 & \fs 25.11                                                                                 & \nd 39.13                                                                                 \\
                                                                                                                     & SSIM   & 0.785                                                                                    & 0.830                                                                                    & 0.883                                                                                    & 0.768                                                                                     & 0.799                                                                                     & \fs 0.925                                                                                 & \fs 0.821                                                                                 & \nd 0.846                                                                                 & \nd 0.962                                                                                 \\
                                                                                                                     & LPIPS  & 0.487                                                                                    & 0.428                                                                                    & 0.392                                                                                    & 0.515                                                                                     & 0.523                                                                                     & 0.289                                                                                     & \fs 0.462                                                                                 & \nd 0.451                                                                                 & \nd 0.183                                                                                 \\
                        \midrule
                        \multirow{3}{*}{\begin{tabular}[l]{@{}c@{}}ESLA\\M~\cite{Johari2022ESLAMED}\end{tabular}}
                                                                                                                     & PSNR   & 19.52                                                                                    & 20.70                                                                                    & 28.48                                                                                    & 18.68                                                                                     & 15.11                                                                                     & 31.15                                                                                     & 16.38                                                                                     & \nd 18.35                                                                                 & 31.08                                                                                     \\
                                                                                                                     & SSIM   & 0.670                                                                                    & 0.715                                                                                    & 0.841                                                                                    & 0.614                                                                                     & 0.518                                                                                     & 0.895                                                                                     & 0.519                                                                                     & 0.603                                                                                     & 0.905                                                                                     \\
                                                                                                                     & LPIPS  & 0.522                                                                                    & 0.487                                                                                    & 0.414                                                                                    & 0.606                                                                                     & 0.836                                                                                     & \fs 0.285                                                                                 & 0.688                                                                                     & 0.664                                                                                     & 0.255                                                                                     \\
                        \midrule
                        \multirow{3}{*}{Ours}                                                                        & PSNR   & \fs 23.72                                                                                & \fs 25.11                                                                                & \fs 32.64                                                                                & \fs 23.08                                                                                 & \fs 24.53                                                                                 & \nd 31.26                                                                                 & \nd 23.83                                                                                 & \fs 25.11                                                                                 & \fs 39.38                                                                                 \\
                                                                                                                     & SSIM   & \fs 0.808                                                                                & \fs 0.840                                                                                & \fs 0.911                                                                                & \fs 0.777                                                                                 & \fs 0.821                                                                                 & \nd 0.909                                                                                 & \nd 0.810                                                                                 & \fs 0.848                                                                                 & \fs 0.963                                                                                 \\
                                                                                                                     & LPIPS  & \fs 0.468                                                                                & \fs 0.423                                                                                & \fs 0.349                                                                                & \fs 0.510                                                                                 & \fs 0.493                                                                                 & 0.358                                                                                     & \nd 0.481                                                                                 & \fs 0.448                                                                                 & \fs 0.182                                                                                 \\
                        \bottomrule
                    \end{tabular}
                    \subcaption*{}
                \end{subfigure}
            }
             &
            \resizebox{0.7\linewidth}{!}
            {
                \begin{subfigure}{0.7\linewidth} 
                    \centering
                    {\footnotesize
                        \setlength{\tabcolsep}{0pt}
                        \newcommand{\sz}{0.1825}
                        \renewcommand{\arraystretch}{0}
                        \begin{tabular}{cccccccc}
                                                                            & RGB Input                                                                                  & COSLAM~\cite{Wang2023CoSLAMJC}                                                              & ESLAM~\cite{Johari2022ESLAMED}                                                             & Ours.RGB                                                                                  & Ours.luminance                                                                                 \\
                            \rotatebox[origin=c]{90}{{\texttt{\#Rm~blur}}}  & \includegraphics[valign=c,width=\sz\linewidth]{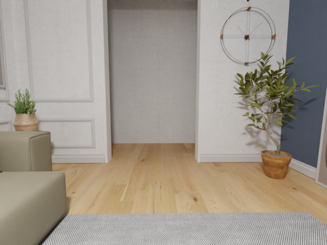}  & \includegraphics[valign=c,width=\sz\linewidth]{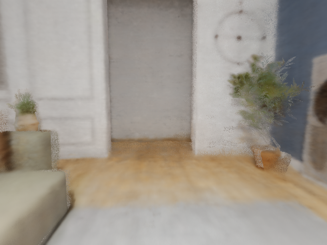}  & \includegraphics[valign=c,width=\sz\linewidth]{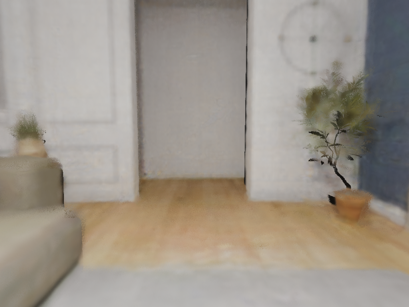}  & \includegraphics[valign=c,width=\sz\linewidth]{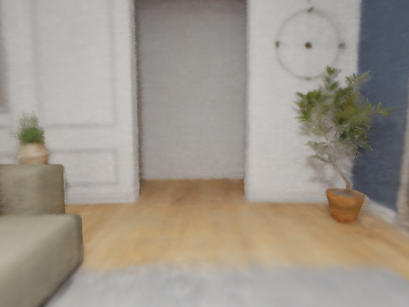}  & \includegraphics[valign=c,width=\sz\linewidth]{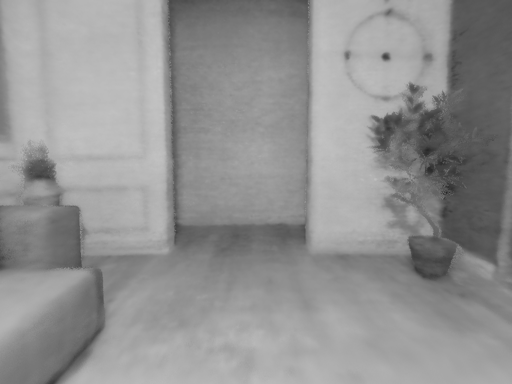}  \\
                            \rotatebox[origin=c]{90}{{\texttt{\#Rm Dark}}}  & \includegraphics[valign=c,width=\sz\linewidth]{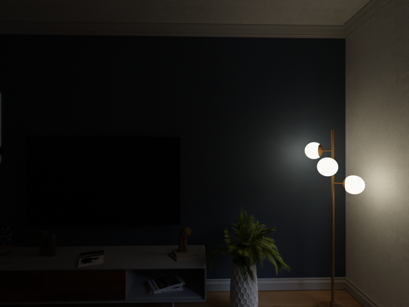}  & \includegraphics[valign=c,width=\sz\linewidth]{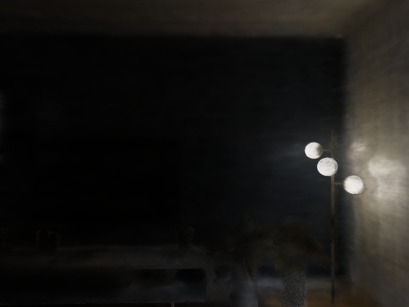}   & \includegraphics[valign=c,width=\sz\linewidth]{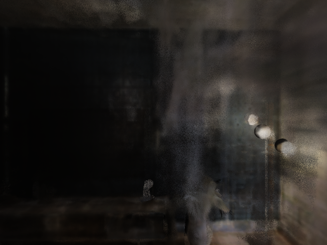}   & \includegraphics[valign=c,width=\sz\linewidth]{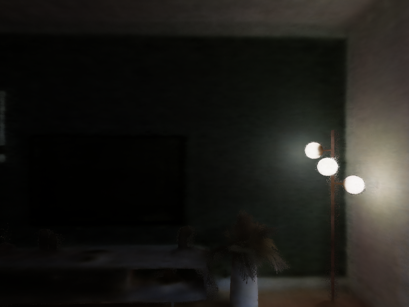}   & \includegraphics[valign=c,width=\sz\linewidth]{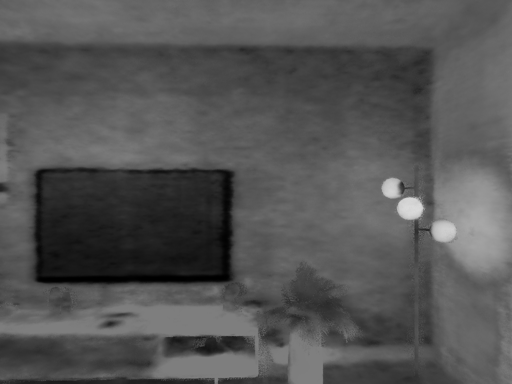}   \\
                            \rotatebox[origin=c]{90}{{\texttt{\#Wkp Dark}}} & \includegraphics[valign=c,width=\sz\linewidth]{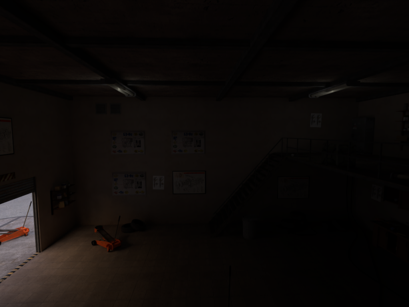} & \includegraphics[valign=c,width=\sz\linewidth]{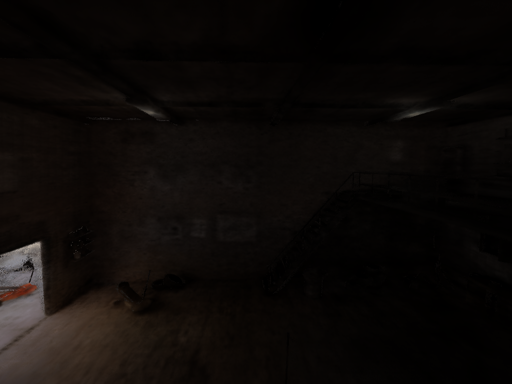} & \includegraphics[valign=c,width=\sz\linewidth]{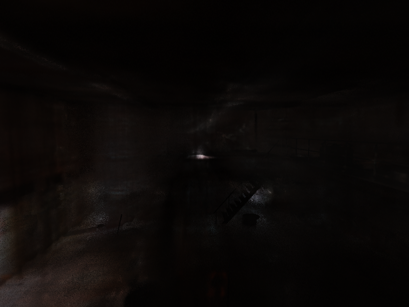} & \includegraphics[valign=c,width=\sz\linewidth]{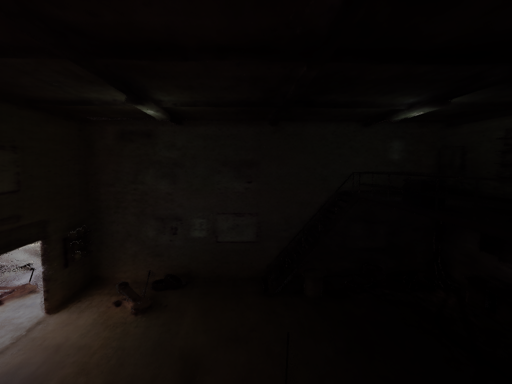} & \includegraphics[valign=c,width=\sz\linewidth]{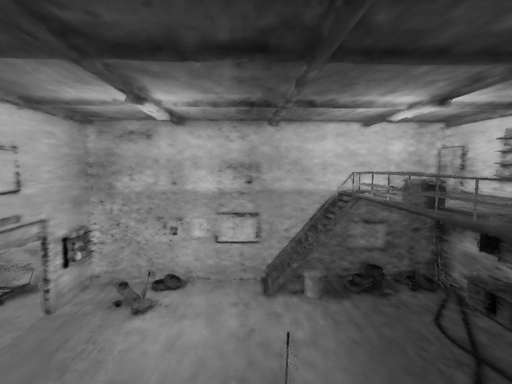} \\                                                                                                                                                                                                                                                                                                                                                                            \\
                        \end{tabular}
                    }
                    \subcaption*{}
                \end{subfigure}
            }
        \end{tabular}
    }
    \vspace{-2ex}
    \label[figure]{fig:render_devindoors}
    \caption{\textbf{Rendering Performance on DEV-Indoors.} left): Our method outperforms most previous works in image quality evaluation under non-ideal environments. right): \ours achieves more precise rendering details on average than previous methods.}
    \vspace{-1ex}
\end{figure*}

We compare the rendering performance in~\cref{fig:render_devindoors}~(left), \ours outperforms most SOTA works in image quality. The thumbnails in \cref{fig:render_devindoors}~(right) show that \ours achieves more precise rendering details than previous methods. Specifically, on \texttt{\#Rm~Blur}, \ours yields more refined results, while CoSLAM and ESLAM exhibit ghosting. Note that in \texttt{\#Rm~Dark} and \texttt{\#Wkp~Dark}, all the RGB rendering is dark and blurred, while our method can still generate high-quality luminance results with the assistance of the HDR event stream.

\begin{figure}[t]
    \centering
    \resizebox{0.47\textwidth}{!}{
        \setlength{\tabcolsep}{2.5pt}
        \begin{tabular}{ccccccccc}
            \toprule
            Tracking                             & Mapping     & \multicolumn{4}{c}{\texttt{\#Rm~blur}} &                   & \multicolumn{2}{c}{\texttt{\#Dorm2}}                                                                          \\ \cline{3-6} \cline{8-9}
            Event                                & Event       & ATE$\downarrow$                        & ACC$\downarrow$   & Comp$\downarrow$                     & Comp ratio$\uparrow$ &       & Median$\downarrow$ & RSME$\downarrow$   \\
            \midrule
            \redx                                & \redx       & 11.89                                  & 8.61              & 10.98                                & 76.31                &       & 14.46              & 18.75              \\
            \redx                                & \greencheck & \nd 10.73                              & 8.32              & \nd 9.53                             & \nd 81.83            &       & \nd 14.17          & \nd 17.07          \\
            \greencheck                          & \redx       & 11.68                                  & \nd 8.28          & 10.28                                & 79.05                &       & 16.09              & 19.72              \\
            \greencheck                          & \greencheck & \fs \textbf{9.61}                      & \fs \textbf{7.88} & \fs \textbf{7.59}                    & \fs \textbf{83.51}   &       & \fs \textbf{11.91} & \fs \textbf{15.47} \\
            \midrule
            \multicolumn{2}{c}{RGB 1st-2nd only} & 10.92       & 9.02                                   & 9.15              & 82.81                                &                      & 13.52 & 17.80                                   \\
            \multicolumn{2}{c}{W/o RGB}          & 12.07       & 11.12                                  & 11.05             & 76.27                                &                      & 22.50 & 26.48                                   \\
            \bottomrule
        \end{tabular}
    }\\[0.2em]
    {\footnotesize
    \setlength{\tabcolsep}{0.pt}
    \renewcommand{\arraystretch}{1}
    \newcommand{\sz}{0.33}
    \begin{tabular}{ccc}
        \includegraphics[height=\sz\linewidth]{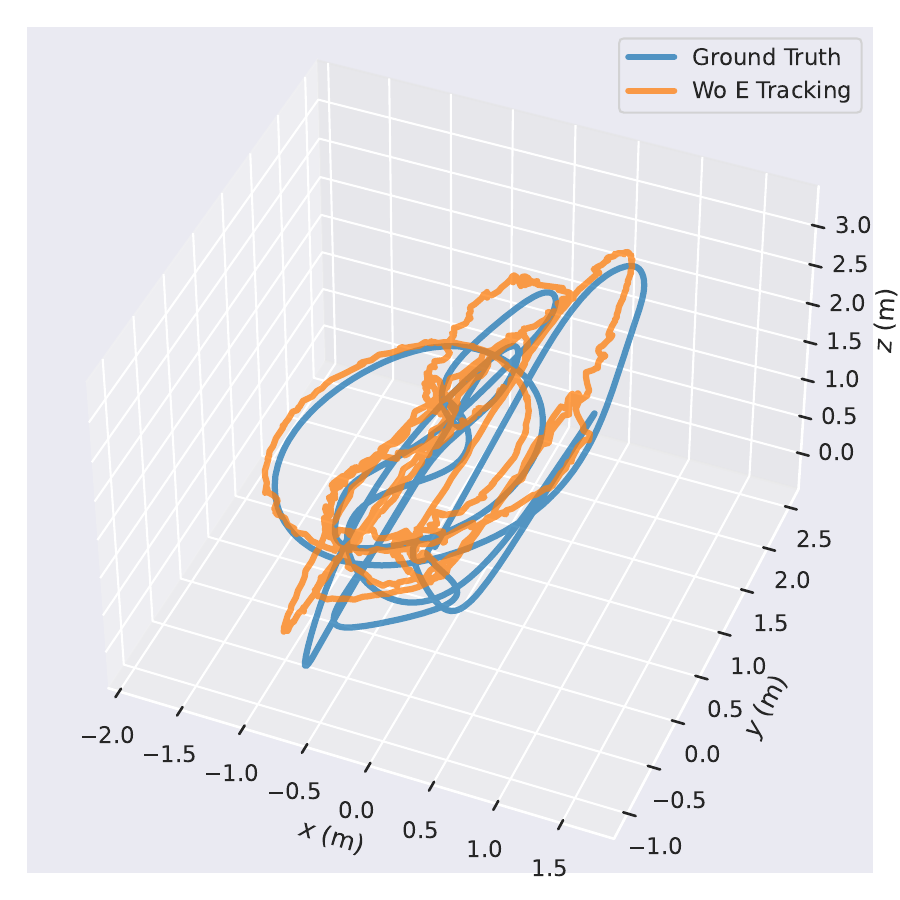} & \includegraphics[height=\sz\linewidth]{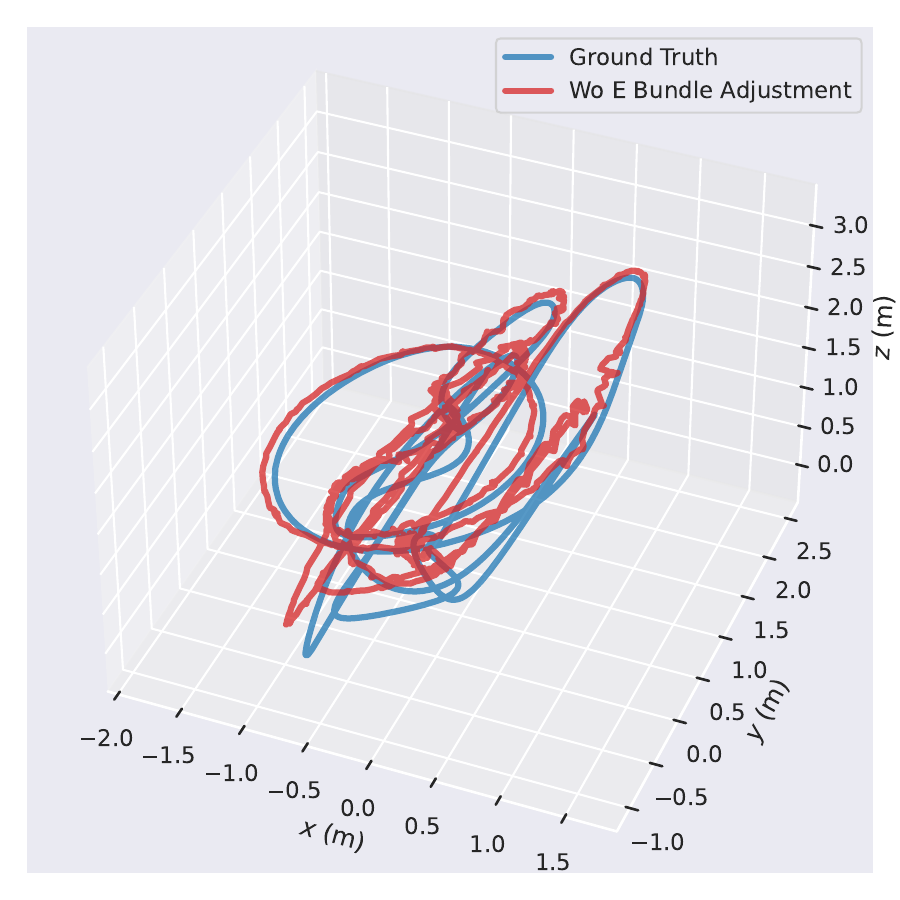} & \includegraphics[height=\sz\linewidth]{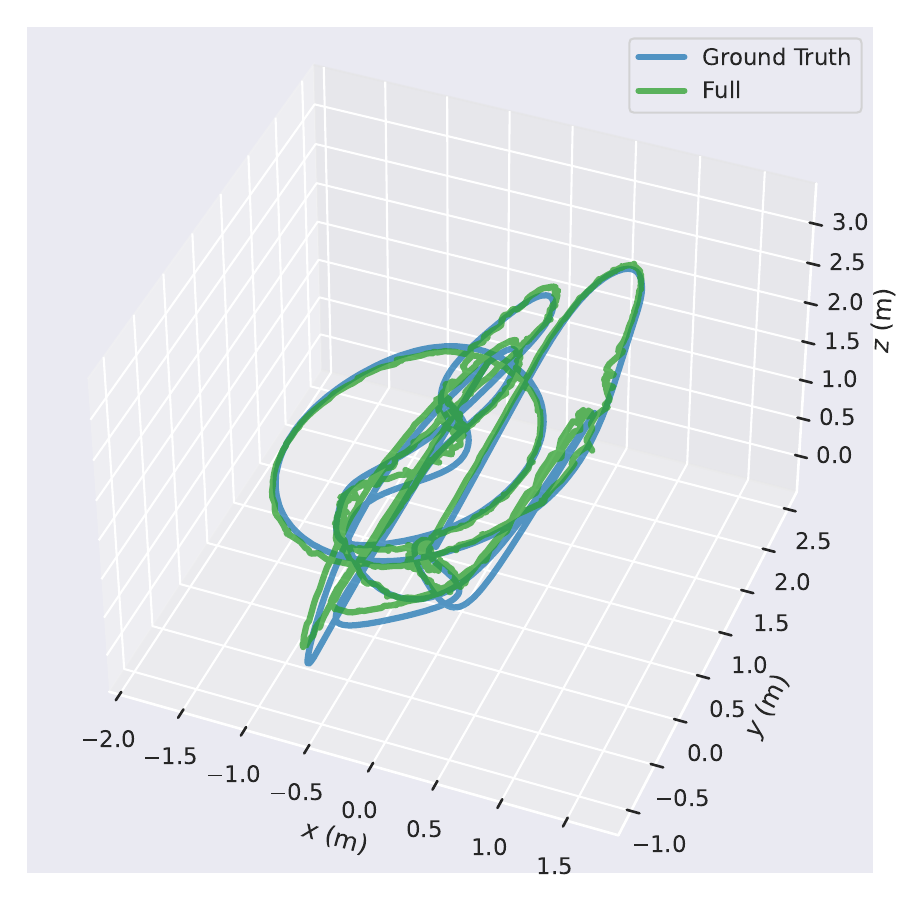} \\
        {\tiny W/o E Mapping. ATE: 11.68}                                        & {\tiny W/o E Tracking. ATE: 10.73}                                 & {\tiny FULL. ATE: 9.61}                                           \\
    \end{tabular}
    }
    \caption{\textbf{Ablation study of modalities} on the \texttt{\#Rm~blur} and \texttt{\#Drom2} subset of DEV-Indoors and DEV-Reals (15 iterations).}
    \label{tab:modalities}
    \vspace{-2ex}
\end{figure}

\subsection{Ablation Study}
\label{sec:ablation}
\noindent{\textbf{Effect of event and RGB modalities.}}
\cref{tab:modalities} illustrates quantitative evaluation using ETA in tracking and mapping. Our full model achieve lower tracking error of 9.61 and 15.47 than the model w/o ETA in tracking (10.73 and 17.07) and mapping (11.68 and 19.72) on the \texttt{\#Rm~blur} and \texttt{\#Dorm2}, respectively. The results also show that the full model surpasses the model w/o ETA by 0.73 and 3.39\% in ACC and completion. In addition, the RGB is also critical, but with the initialization of RGB in 2 frames, the performance is significantly improved, benefiting from the CRF.

\noindent{\textbf{Effect of CRF and probability-weighted sampling.}}
\cref{tab:modalities} show the performance of differentiable CRF and probability-weighted sampling strategy (PWS). The system without CRF might suffer from the distinct event and RGB imaging process, resulting in fluctuating training and poor performance, especially in real datasets. It significantly reduces the tracking ATE RSME from $30.18$ to $15.47$ in \texttt{\#Drom2} and reconstruction completion from $9.78$ to $7.59$ in \texttt{\#Rm~blur}. The results also show that the full model surpasses the model w/o PWS by 0.25 and 1.9\% in ATE and completion on \texttt{\#Rm~blur}. A visualization of CRF is shown in \cref{fig:ablation_crf}, on dark scene \texttt{\#Drom2}, the model with CRF renders HDR luminance results and accurate mesh benefiting from events. In contrast, the model w/o CRF suffers from the low dynamic range RGB input.
\begin{table}[t]
    \centering
    \scriptsize
    \resizebox{0.47\textwidth}{!}{
        \setlength{\tabcolsep}{2.5pt}
        \begin{tabular}{lccccccc}
            \toprule
            \multirow{2}{*}{Setting} & \multicolumn{4}{c}{\texttt{\# Rm~blur}} &                   & \multicolumn{2}{c}{\texttt{\#Dorm2}}                                                                     \\ \cline{2-5} \cline{7-8}
                                     & ATE$\downarrow$                         & ACC$\downarrow$   & Comp$\downarrow$                     & Comp ratio$\uparrow$ &  & Median$\downarrow$ & RSME$\downarrow$   \\
            \midrule
            w/o CRF                  & 12.12                                   & 8.29              & 9.78                                 & \nd 83.57            &  & 27.67              & 30.18              \\
            w/o PWS                  & \nd 9.86                                & \nd 7.88          & \nd 9.49                             & 81.04                &  & 16.59              & 19.78              \\
            Full model               & \fs \textbf{9.61}                       & \fs \textbf{7.88} & \fs \textbf{7.59}                    & \fs \textbf{83.51}   &  & \fs \textbf{11.91} & \fs \textbf{15.47} \\
            \bottomrule
        \end{tabular}
    }
    \caption{\textbf{Ablation study of CRF and PWS} on the \texttt{\#Rm~blur} and \texttt{\#Drom2} subset of DEV-Indoors and DEV-Reals (15 iterations).}
    \label{tab:ablation_event}
    \vspace{-2ex}
\end{table}
\begin{figure}[t]
    \centering
    {\footnotesize
        \setlength{\tabcolsep}{0.25pt}
        \renewcommand{\arraystretch}{0}
        \newcommand{\sz}{0.248}
        \begin{tabular}{cccc}
            {\tiny RGB Input}                                               & {\tiny Event Input}                                               & {\tiny W/o CRF Color Rendering}                                   & {\tiny FULL Luminance Rendering}                                 \\
            \includegraphics[width=\sz\linewidth]{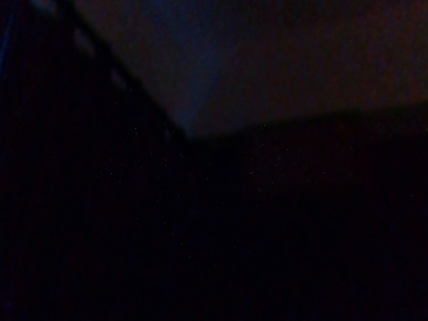} & \includegraphics[width=\sz\linewidth]{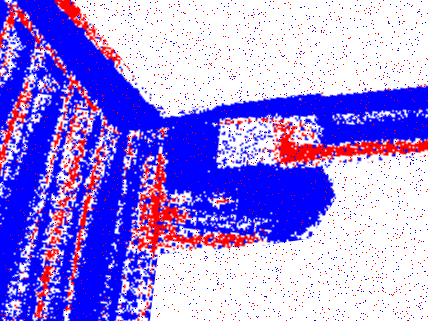} & \includegraphics[width=\sz\linewidth]{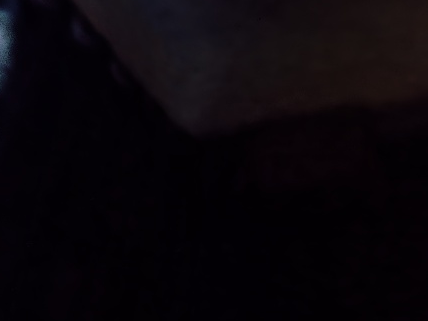} & \includegraphics[width=\sz\linewidth]{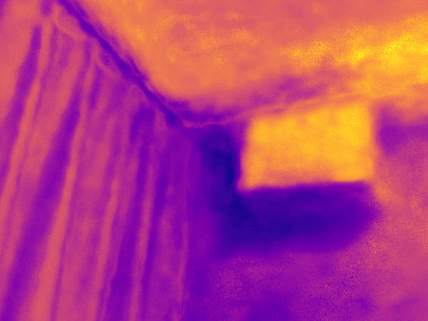}
        \end{tabular}
    }
    \\[0.0em]
    {\footnotesize
    \setlength{\tabcolsep}{0.pt}
    \renewcommand{\arraystretch}{0.5}
    \newcommand{\sz}{0.5}
    \begin{tabular}{cc}
        \includegraphics[width=\sz\linewidth]{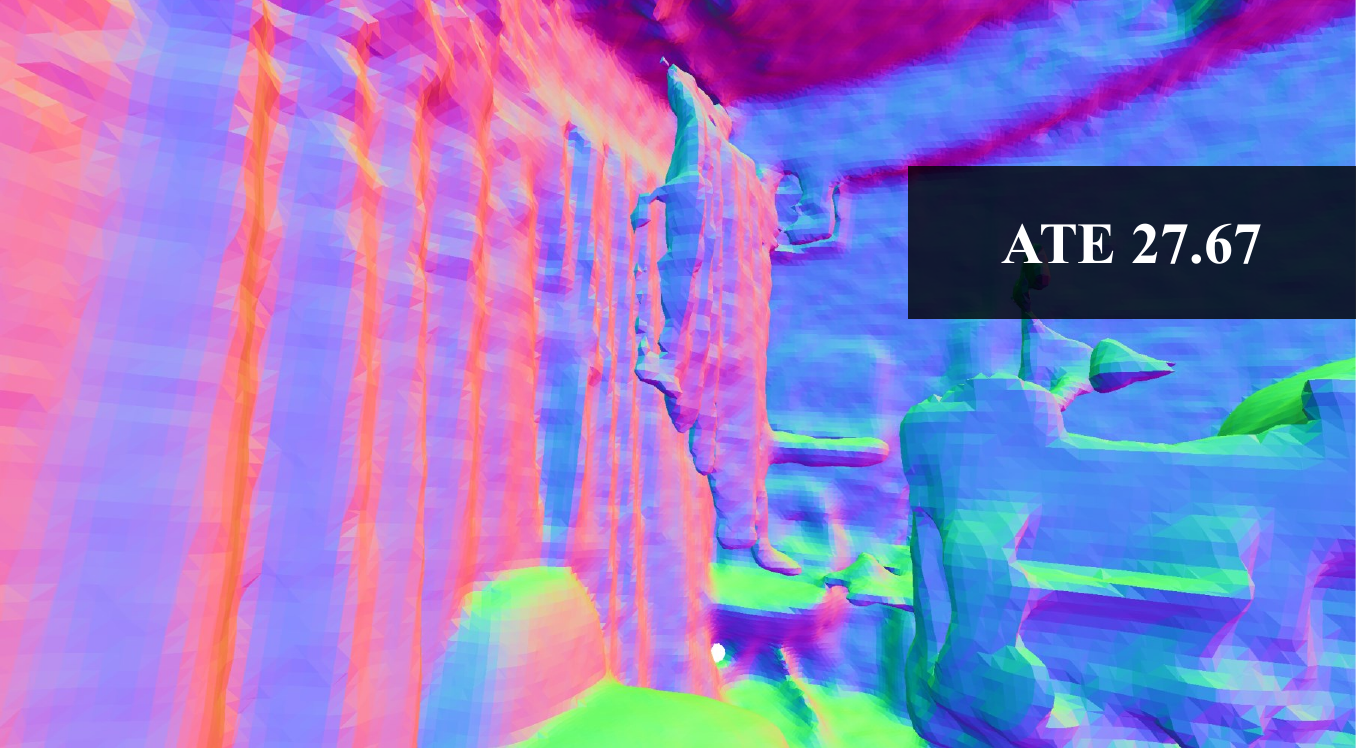} & \includegraphics[width=\sz\linewidth]{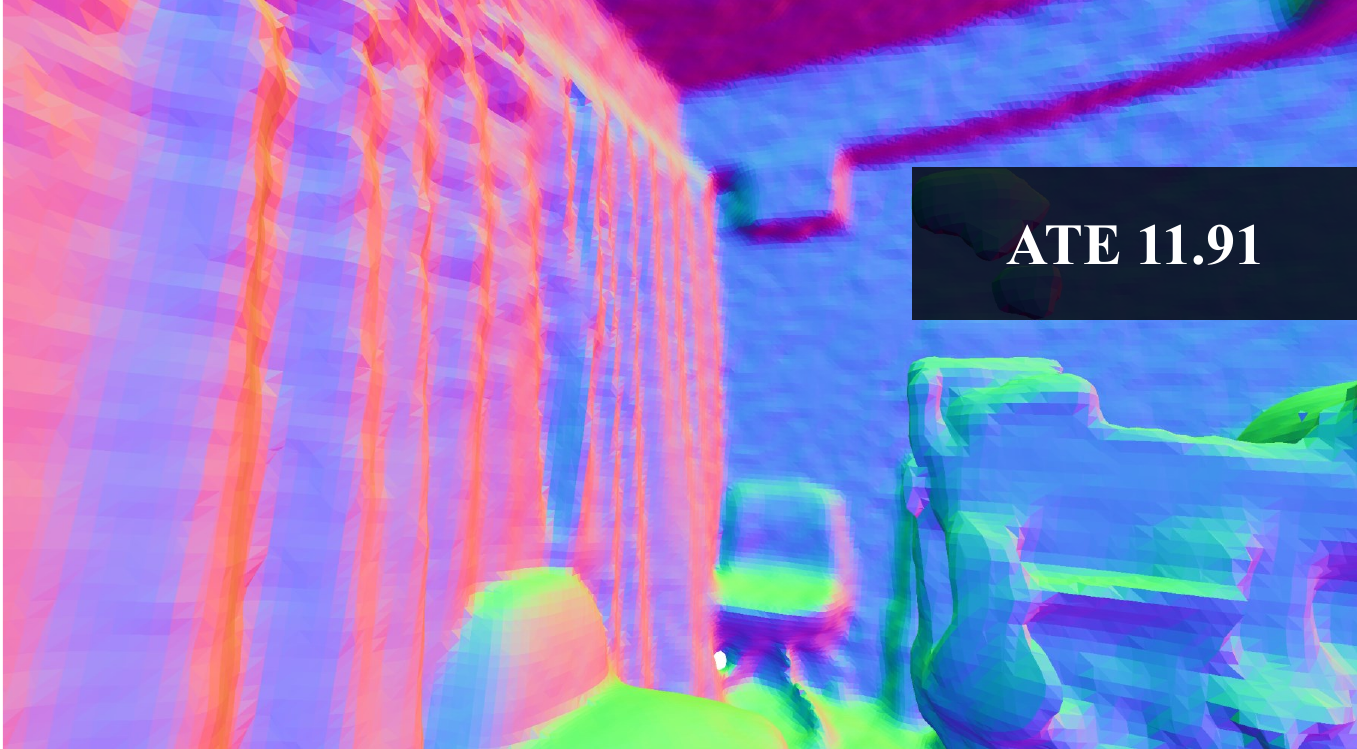} \\
        {\tiny W/o CRF. Mesh}                                                  & {\tiny FULL. Mesh}                                                   \\
    \end{tabular}
    }
    \caption{\textbf{CRF ablation} on the \texttt{\#Dorm2} of DEV-Reals.}
    \label{fig:ablation_crf}
    \vspace{-2ex}
\end{figure}
\section{Conclusion and Limitation}
\label{sec:limitation_conclusion}
This paper first integrates the event stream into the implicit neural SLAM framework to overcome challenges in scenes with motion blur and lighting variation. A differentiable CRF rendering technique that maps the unified representation to color and luminance is proposed to address the significant distinction between event and RGB. An event temporal aggregating optimization strategy that capitalizes the consecutive difference constraints of events is presented to enhance the optimization. We construct \textbf{DEV-Indoors} and \textbf{DEV-Reals} datasets to evaluate the effectiveness of \ours under various environments. However, \ours relies on depth input, which might be unavailable in some scenarios. Besides, \ours focuses on indoor scenes and might face challenges in boundless long trajectories. In future work, we aim to extend it to large-scale outdoor environments and enhance the generalization capability.

{\small

\boldparagraph{Acknowledgements.}
This work is supported by the Shanghai AI Laboratory, National Key R\&D Program of China (2022ZD0160101), the National Natural Science Foundation of China (62376222), and Young Elite Scientists Sponsorship Program by CAST (2023QNRC001).
}

\clearpage
\setcounter{page}{1}
\maketitlesupplementary

\begin{abstract}
    \noindent This supplementary material accompanies the main paper by providing more details for reproducibility as well as additional evaluations and qualitative results to to verify the effectiveness and robustness of \ours:\\
    \noindent $\triangleright$ \textbf{\cref{sec:devindoors}}: Configurations of DEV-Indoors dataset, including scene assets, event generation, evaluation dataset, ground truth mesh production, and sequence visualization. \\
    \noindent $\triangleright$ \textbf{\cref{sec:devreals}}: Configurations of DEV-Reals dataset, including capture system specifications and sequence visualization. \\
    \noindent $\triangleright$ \textbf{\cref{sec:add_implement_detail}}: Additional implementation details. \\
    \noindent $\triangleright$ \textbf{\cref{sec:add_exp}}: Additional experimental results, including more ablation studies, detailed tracking comparison, and mapping reconstruction visualization. \\
    \noindent $\triangleright$ \textbf{\cref{sec:video}}: Video demonstration.
\end{abstract}
\section{Configurations of DEV-Indoors dataset}
\label{sec:devindoors}
\begin{table}[htbp]
    \centering
    \footnotesize
    \caption{Comparison of different event-centric datasets. We focus on the availability of event data, color images, depth, and ground truth mesh. \textbf{I} denotes indoor scenes. \textbf{O} denotes outdoor scenes.}
    \label{tab:dataset_cmp}
    \setlength{\tabcolsep}{3pt}
    \renewcommand{\arraystretch}{1}
    \resizebox{\columnwidth}{!}{
        \begin{tabular}{rcccccccc}
            \toprule
            \multirow{2}{*}{Dataset}        & event       & RGB/gray    & Depth       & GT          & challenging & lighting    & indoors /    & Synthetic /  \\ \cline{2-4}
                                            & data        & image       & image       & mesh        & motion blur & change      & outdoors     & Real         \\
            \midrule
            ECDS~\cite{Mueggler2016TheED}   & \greencheck & \graycheck  & \redx       & \redx       & \greencheck & \greencheck & \textbf{I+O} & \textbf{S+R} \\
            RPG~\cite{ESVO}                 & \greencheck & \graycheck  & \redx       & \redx       & \greencheck & \greencheck & \textbf{I}   & \textbf{S}   \\
            MVSEC~\cite{Zhu2018TheMS}       & \greencheck & \graycheck  & \greencheck & \redx       & \redx       & \greencheck & \textbf{I+O} & \textbf{R}   \\
            UZH-FPV	~\cite{Delmerico19icra} & \greencheck & \graycheck  & \redx       & \redx       & \greencheck & \redx       & \textbf{I+O} & \textbf{R}   \\
            DSEC~\cite{Gehrig21ral}         & \greencheck & \greencheck & \redx       & \redx       & \redx       & \greencheck & \textbf{O}   & \textbf{R}   \\
            TUM-VIE	~\cite{klenk2021tumvie} & \greencheck & \graycheck  & \redx       & \redx       & \greencheck & \greencheck & \textbf{I}   & \textbf{R}   \\
            EDS~\cite{EDS}                  & \greencheck & \greencheck & \redx       & \redx       & \greencheck & \greencheck & \textbf{I}   & \textbf{R}   \\
            Vector~\cite{Gao2022VECtorAV}   & \greencheck & \graycheck  & \greencheck & \redx       & \greencheck & \greencheck & \textbf{I}   & \textbf{R}   \\
            M2DGR~\cite{M2DGR}              & \greencheck & \greencheck & \redx       & \redx       & \greencheck & \greencheck & \textbf{I+O} & \textbf{R}   \\
            VICON~\cite{guan2022monocular}  & \greencheck & \redx       & \redx       & \redx       & \redx       & \redx       & \textbf{O}   & \textbf{R}   \\
            ViVID++~\cite{lee2022vivid++}   & \greencheck & \greencheck & \greencheck & \redx       & \redx       & \greencheck & \textbf{O}   & \textbf{R}   \\
            VISTA 2.0~\cite{amini2022vista} & \greencheck & \greencheck & \greencheck & \redx       & \redx       & \greencheck & \textbf{O}   & \textbf{S}   \\
            \midrule
            DEV-Indoors (ours)              & \greencheck & \greencheck & \greencheck & \greencheck & \greencheck & \greencheck & \textbf{I}   & \textbf{S}   \\
            DEV-Reals (ours)                & \greencheck & \greencheck & \greencheck & \redx       & \greencheck & \greencheck & \textbf{I}   & \textbf{R}   \\
            \bottomrule
        \end{tabular}
    }
    \vspace{-2ex}
\end{table}
\noindent \cref{tab:dataset_cmp} presents a comparison of the prevalent event-centric datasets available today. In this work, we focus on addressing challenges associated with motion blur and lighting variations within indoor settings rather than ground robot navigation or SLAM from a UAV perspective. A pervasive issue with current datasets is the absence of ground truth depth~\cite{Mueggler2016TheED,ESVO,Delmerico19icra,Gehrig21ral,klenk2021tumvie,EDS,guan2022monocular,M2DGR} or mesh data~\cite{Zhu2018TheMS,Gao2022VECtorAV,amini2022vista,lee2022vivid++}, which are essential for the operation and evaluation of NeRF-based SLAM methods. In addition, many outdoor datasets are geared towards large-scale navigation~\cite{Gehrig21ral,M2DGR,lee2022vivid++,amini2022vista} and lack significant motion blur and lighting variation, making them unsuitable for our intended purposes. Besides, most datasets are synthetic~\cite{Mueggler2016TheED,ESVO,amini2022vista}, which are not representative of real-world scenarios or provide sample motion~\cite{Mueggler2016TheED,Bryner19icra}. To address the existing limitations, we introduce the synthetic dataset DEV-Reals and DEV-Indoors, which consist of 6 scenes and 17 sequences with practical motion blur and lighting changes.

\begin{figure*}[t]
    \vspace{-4ex}
    \centering
    {\footnotesize
        \setlength{\tabcolsep}{1pt}
        \newcommand{\sz}{0.09}
        \begin{tabular}{cccc}
                                                        & \texttt{\#Room}                                                                    & \texttt{\#Apartment}                                                                & \texttt{\#Workshop}                                                                \\
            \rotatebox[origin=c]{90}{Modele in Blender} & \includegraphics[valign=c,height=\sz\linewidth]{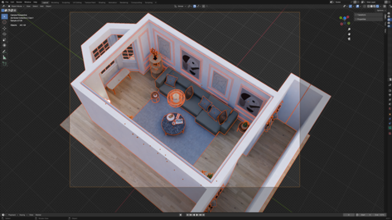} & \includegraphics[valign=c,height=\sz\linewidth]{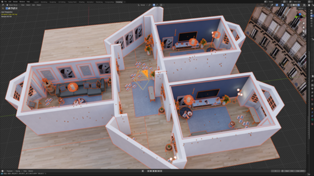} & \includegraphics[valign=c,height=\sz\linewidth]{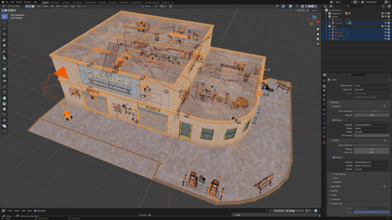} \\
            \addlinespace[2pt]
            \rotatebox[origin=c]{90}{Trajectory}        & \includegraphics[valign=c,height=\sz\linewidth]{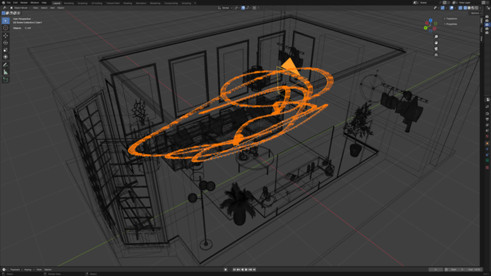}  & \includegraphics[valign=c,height=\sz\linewidth]{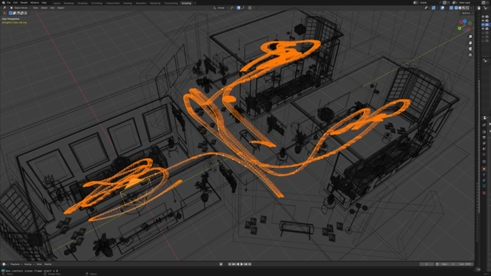}  & \includegraphics[valign=c,height=\sz\linewidth]{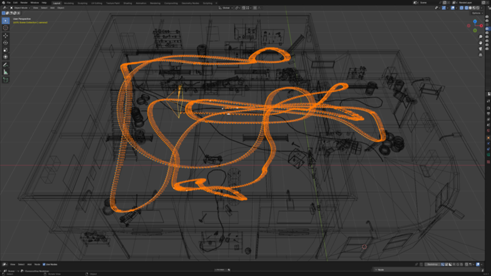} \\
        \end{tabular}
    }
    \caption{\textbf{The models and trajectories} of the DEV-Indoors dataset in Blender~\cite{blender}, including \texttt{\#room}, \texttt{\#apartment}, and \texttt{\#workshop}.}
    \label{fig:devindoors_models}
\end{figure*}

\boldparagraph{Scene Assets of DEV-Indoors.} We use the Blender~\cite{blender} to construct the synthetic DEV-Indoors dataset, including three high-quality models: \texttt{\#Room}, \texttt{\#Apartment} and \texttt{\#Workshop}. \cref{fig:devindoors_models} illustrates the blender models and corresponding camera trajectories. Unlike the camera motion on the Replica dataset~\cite{straub2019replica}, our camera trajectory is six degrees of freedom (6-DOF), and the motion is highly complex. The camera trajectory is obtained through manual manipulation of position and orientation and further refined through smoothing operations.

\boldparagraph{Event Data Generation.} The simulated event data in DEV-indoors are obtained via the following three steps: first, we render high-quality RGB captures covering norm, motion blur, and dark scenarios by varying the scene lighting and camera exposure time. Second, we perform a video frame interpolation algorithm FILM~\cite{reda2022film} to convert the rendered images into ultra-high frequency RGB frames. Finally, We use the event camera simulator~\cite{Gehrig_2020_CVPR} to generate synthetic event data.
\begin{figure}[t]
    \centering
    {\footnotesize
        \setlength{\tabcolsep}{0pt}
        \newcommand{\sz}{0.11}
        \begin{tabular}{ccc}
            \includegraphics[valign=c,height=\sz\linewidth]{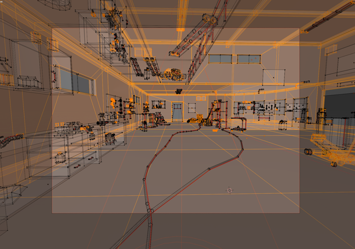} & \includegraphics[valign=c,height=\sz\linewidth]{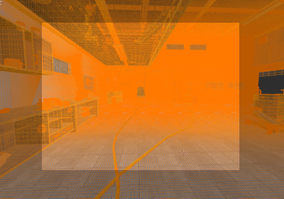} & \includegraphics[valign=c,height=\sz\linewidth]{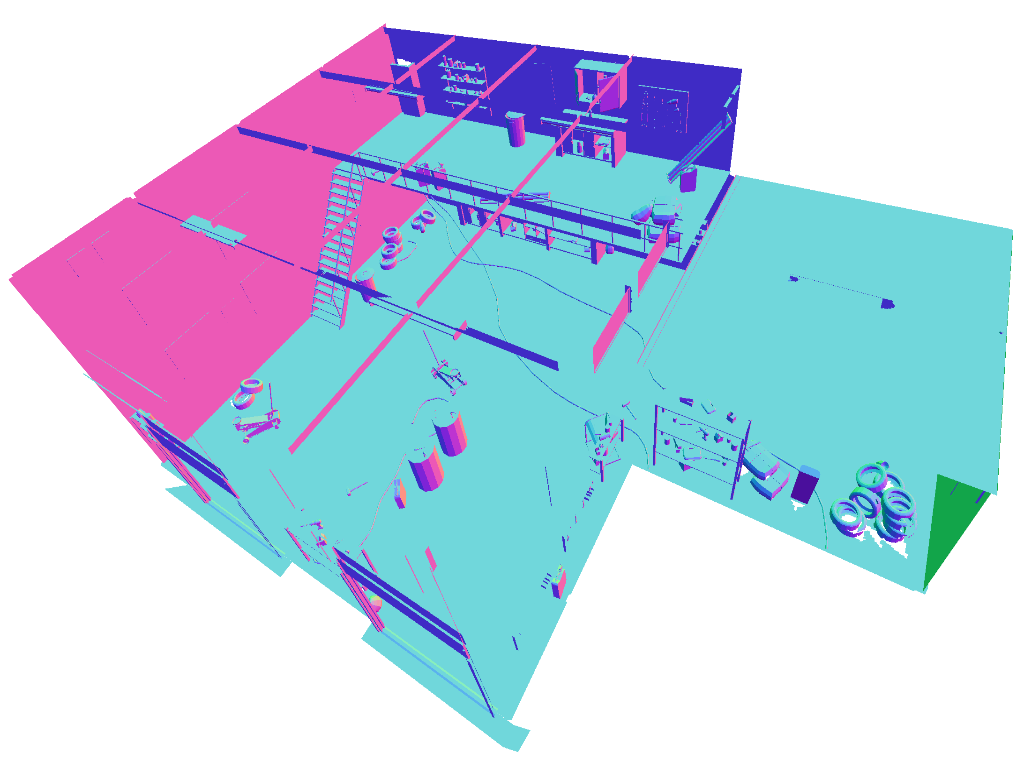} \\
            Sparse Modeling                                                                      & Dense Triangulation                                                                 & Final Mesh
        \end{tabular}
    }
    \caption{\textbf{The ground truth mesh generation process} of \texttt{\#workshop} in DEV-Indoors dataset.}
    \label{fig:gt_esh_process}
\end{figure}

\begin{figure}[htbp]
    \vspace{-2ex}
    \centering
    {\footnotesize
        \setlength{\tabcolsep}{0pt}
        \newcommand{\sz}{0.105}
        \begin{tabular}{ccc}
            \includegraphics[valign=c,height=\sz\linewidth]{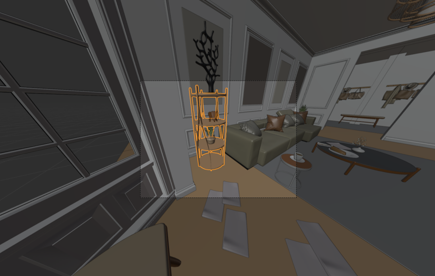} & \includegraphics[valign=c,height=\sz\linewidth]{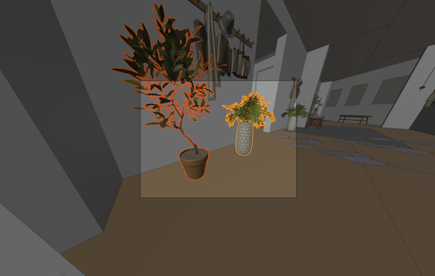} & \includegraphics[valign=c,height=\sz\linewidth]{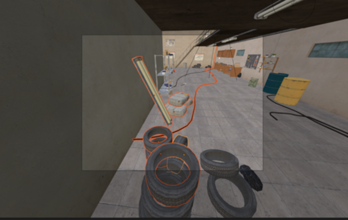} \\
            \texttt{\#Room}                                                                          & \texttt{\#Apartment}                                                                 & \texttt{\#Workshop}
        \end{tabular}
    }
    \caption{\textbf{Extra virtual views} of \texttt{\#Room}, \texttt{\#Apartment} and \texttt{\#Workshop} models in DEV-Indoors dataset.}
    \label{fig:eval_dataset}
\end{figure}

\boldparagraph{Ground Truth Mesh.} As shown in~\cref{fig:gt_esh_process}, to obtain a dense mesh that can apply to algorithm reconstruction, we perform detailed and dense triangulation on the models and use the sampling algorithm of Open3D~\footnote{\href{http://www.open3d.org/docs/0.7.0/python_api/open3d.geometry.simplify_vertex_clustering.html}{open3d.geometry.simplify\_vertex\_clustering}} to uniformly sample them to avoid points gathering on the surface of small objects. Then, we further use the mesh culling in~\cite{Wang2023CoSLAMJC} to remove the unseen vertices of the models. This process ultimately yields a high-quality mesh that can be used for evaluation. Note that although Blender can directly export point cloud files in PLY format, they cannot be directly used for reconstruction evaluation. The reason is that the models created in Blender are highly structured and sparsely connected, where a face may only be covered by a few vertices.

\boldparagraph{Evaluation Datasets.} To construct the evaluation subsets, we use frustum + occlusion + virtual cameras that introduce extra virtual views to cover the occluded parts inside the region of interest in CoSLAM~\cite{Wang2023CoSLAMJC}. The evaluation datasets are generated by randomly conducting 2000 poses and depths in Blender for each scene. We further manually add extra virtual views to cover all scenes, as shown in~\cref{fig:eval_dataset}. This process helps to evaluate the view synthesis and hole-filling capabilities of the algorithm.

\boldparagraph{Dataset Sequence Visualization.} We show the visualization details in~\cref{fig:devindoors_overview}, including 9 subsets: \#Room Norm, \#Room Blur, \#Room Dark, \#Apartment Norm, \#Apartment Blur, \#Apartment Dark, \#Workshop Norm, \#Workshop Blur, and \#Workshop Dark, with corresponding RGB frames, event data, and depth images.

\section{Configurations of DEV-Reals dataset}
\label{sec:devreals}
\begin{figure}[t]
    \begin{center}
        \includegraphics[width=0.9\linewidth]{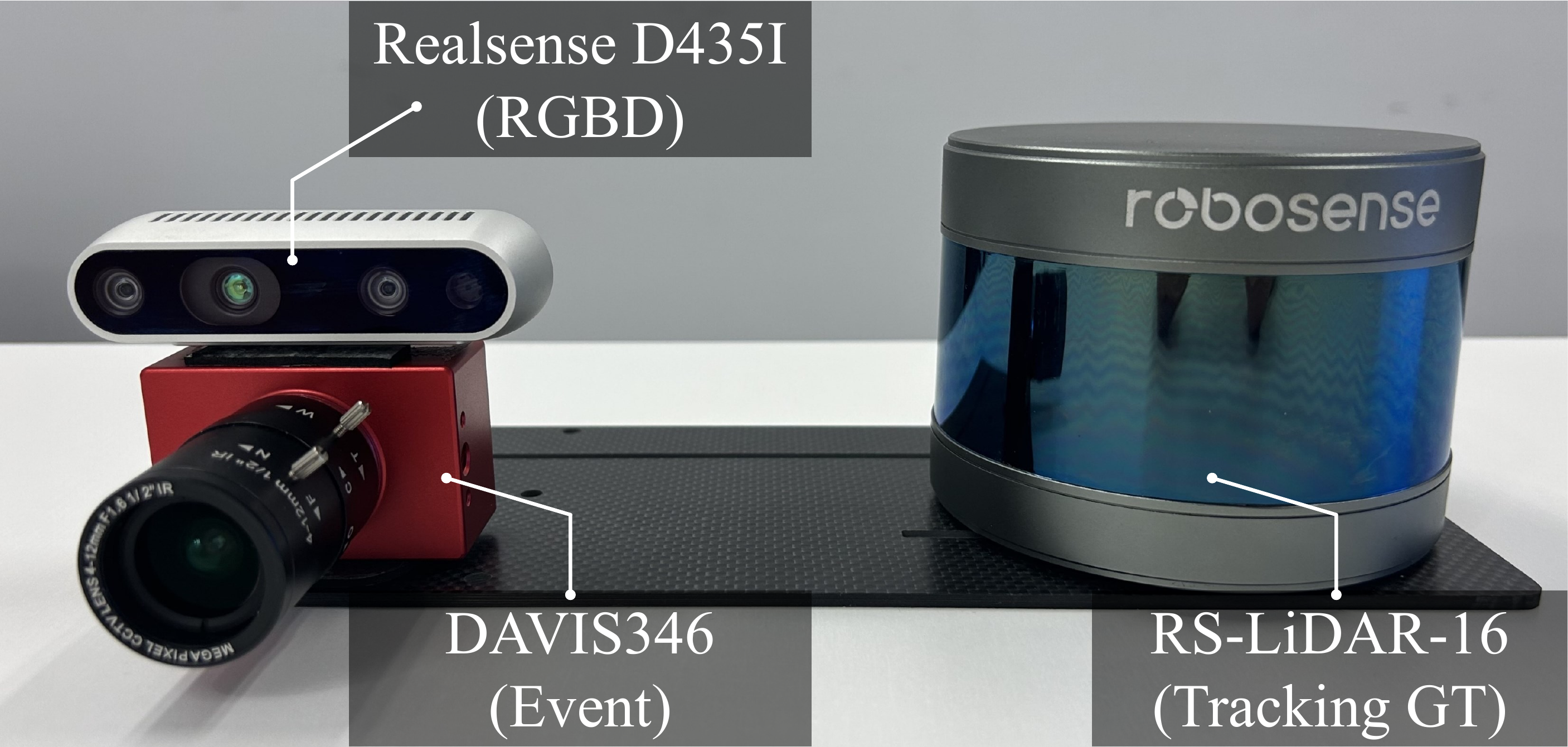}
    \end{center}
    \caption{Illustration of the DEV-Reals capture configuration.}
    \label{fig:capture_config}
\end{figure}

\boldparagraph{Capture System.} As shown in~\cref{fig:capture_config}, our capture system comprises a LiDAR (for ground truth pose), a Realsense D435I RGBD camera, and a DAVIS346 event camera. Besides, we report the hardware specifications of our capture system in~\cref{tab:specifications}. All data sequences are recorded on a PC running Ubuntu 18.04 LTS on an Intel Core i7 CPU. We use the Kalibr toolkit to calibrate the extrinsic parameters between IMUs of DAVIS346 and Realsense D435I. The ground truth trajectories are obtained using the advanced implementation of LOAM~\cite{Zhang2014LOAMLO} algorithm. Time calibration across all sensors is synchronized to a \textbf{millisecond level}, and spatial calibration accuracy is in \textbf{millimeters level}.
\begin{table}[htbp]
    \centering
    \footnotesize
    \caption{Capture System Sensors Specifications of DEV-Reals.}
    \label{tab:specifications}
    \setlength{\tabcolsep}{10pt}
    \renewcommand{\arraystretch}{1}
    \resizebox{\columnwidth}{!}{
        \begin{tabular}{ccc}
            \toprule
            Sensors         & Rate / Bandwidth & Specifications                                       \\
            \midrule
                            &                  & 1920 $\times$ 1080 pixels,                           \\
            Realsense D435I & 90 / 30 fps      & Depth: 69$^{\circ}$H / 42$^{\circ}$V, Stereoscopic,  \\
                            &                  & RGB: 87$^{\circ}$H / 58$^{\circ}$V, Rolling Shutter. \\
            \midrule
                            &                  & 346 $\times$ 260 pixels,                             \\
            DAVIS346        & 12 MEvents / s   & DVS: 120 dB, APS: 56.7 dB,                           \\
                            &                  & f/2.1-12, FoV: 125$^{\circ}$D / 97.7$^{\circ}$V.     \\
            \midrule
            RS-LiDAR-16     & 10 hz            & 6 DoF ground truth trajectory.                       \\
            \bottomrule
        \end{tabular}
    }
    \vspace{-2ex}
\end{table}

\boldparagraph{Dataset Sequence Visualization.} The dataset is captured in three challenging scenarios: \#Pioffice, \#Garage, and \#Dormitory by changing the lighting conditions and camera movement speed in the environment.  We report the visualization details in~\cref{fig:devreals_overview}, including 8 subsets: \#Pioffice1, \#Pioffice2, \#Garage1, \#Garage2, \#Dormitory1, \#Dormitory2, \#Dormitory3 and \#Dormitory4, with corresponding RGB frames, event data, and depth images. Compared with the synthetic DEV-Indoors dataset, the DEV-Reals dataset is more challenging and realistic, containing depth and event noise, which is more suitable for evaluating the robustness of the algorithm.

\section{Additional implementation details}
\label{sec:add_implement_detail}
\begin{table*}[t]
    \centering
    \footnotesize
    \vspace{-2ex}
    \caption{\textbf{Tracking (RSME)} and \textbf{run-time} comparison with detailed iteration setting on \textbf{DEV-Indoors} dataset. Our method outperforms previous works in both accuracy and efficiency in most subsets, demonstrating its robustness under motion blur and luminance variation.}
    \label{tab:tracking_devindoors_iter}
    \resizebox{1.0\textwidth}{!}{
        \setlength{\tabcolsep}{8pt}
        \renewcommand{\arraystretch}{0.8}
        \begin{tabular}{clllllllllll}
            \toprule
            Method                                                                                        & Metric                   & \begin{tabular}[c]{@{}c@{}}\texttt{\#Rm}\\ \textbf{\texttt{norm}}\end{tabular} & \begin{tabular}[c]{@{}c@{}}\texttt{\#Rm}\\ \textbf{\texttt{blur}}\end{tabular} & \begin{tabular}[c]{@{}c@{}}\texttt{\#Rm}\\ \textbf{\texttt{dark}}\end{tabular} & \begin{tabular}[c]{@{}c@{}}\texttt{\#Apt}\\ \textbf{\texttt{norm}}\end{tabular} & \begin{tabular}[c]{@{}c@{}}\texttt{\#Apt}\\ \textbf{\texttt{blur}}\end{tabular} & \begin{tabular}[c]{@{}c@{}}\texttt{\#Apt}\\ \textbf{\texttt{dark}}\end{tabular} & \begin{tabular}[c]{@{}c@{}}\texttt{\#Wkp}\\ \textbf{\texttt{norm}}\end{tabular} & \begin{tabular}[c]{@{}c@{}}\texttt{\#Wkp}\\ \textbf{\texttt{blur}}\end{tabular} & \begin{tabular}[c]{@{}c@{}}\texttt{\#Wkp}\\ \textbf{\texttt{dark}}\end{tabular} & \begin{tabular}[c]{@{}c@{}}\texttt{\#all}\\ \textbf{\texttt{avg}}\end{tabular} \\
            \midrule
            \multirow{4}{*}{\begin{tabular}[l]{@{}c@{}}iMAP\\~\cite{Sucar2021iMAPIM}\end{tabular}}        & ATE RMSE (cm)            & 41.08                                                                                   & 50.58                                                                                   & 70.77                                                                                   & 25.75                                                                                    & 14.41                                                                                    & 1.06$e^5$                                                                                & 276.91                                                                                   & 891.86                                                                                   & 345.21                                                                                   & 214.57                                                                                   \\
                                                                                                          & Tracking (ms) $\uparrow$ & 24.72$\times$50                                                                         & 24.66$\times$50                                                                         & 24.66$\times$50                                                                         & 24.76$\times$50                                                                          & 24.79$\times$50                                                                          & 24.70$\times$50                                                                          & 24.78$\times$50                                                                          & 24.75$\times$50                                                                          & 24.73$\times$50                                                                          & 24.73$\times$50                                                                          \\
                                                                                                          & Mapping (ms) $\uparrow$  & 45.97$\times$300                                                                        & 45.72$\times$300                                                                        & 45.68$\times$300                                                                        & 45.33$\times$300                                                                         & 45.50$\times$300                                                                         & 45.34$\times$300                                                                         & 45.34$\times$300                                                                         & 45.94$\times$300                                                                         & 45.76$\times$300                                                                         & 41.18$\times$300                                                                         \\
                                                                                                          & FPS $\uparrow$           & 0.07                                                                                    & 0.07                                                                                    & 0.07                                                                                    & 0.07                                                                                     & 0.07                                                                                     & 0.07                                                                                     & 0.07                                                                                     & 0.07                                                                                     & 0.07                                                                                     & 0.07                                                                                     \\
            \midrule
            \multirow{4}{*}{\begin{tabular}[l]{@{}c@{}}NICE-SLAM\\~\cite{Zhu2021NICESLAMNI}\end{tabular}} & ATE RMSE (cm)            & 17.06                                                                                   & 29.54                                                                                   & 30.53                                                                                   & 25.17                                                                                    & 44.22                                                                                    & 48.28                                                                                    & \blackx 94 \%                                                                            & \blackx 33 \%                                                                            & \blackx 33\%                                                                             & 32.47                                                                                    \\
                                                                                                          & Tracking (ms) $\uparrow$ & 7.70$\times$10                                                                          & 7.31$\times$10                                                                          & 7.44$\times$10                                                                          & 5.88$\times$20                                                                           & 5.82$\times$20                                                                           & 5.93$\times$20                                                                           & 5.88$\times$20                                                                           & 5.91$\times$20                                                                           & 5.89$\times$20                                                                           & 6.46$\times$16                                                                           \\
                                                                                                          & Mapping (ms) $\uparrow$  & 27.65$\times$120                                                                        & 26.31$\times$120                                                                        & 26.45$\times$120                                                                        & 26.10$\times$120                                                                         & 25.43$\times$120                                                                         & 26.53$\times$120                                                                         & 26.07$\times$120                                                                         & 26.65$\times$120                                                                         & 26.59$\times$120                                                                         & 26.42$\times$120                                                                         \\
                                                                                                          & FPS $\uparrow$           & 0.30                                                                                    & 0.32                                                                                    & 0.32                                                                                    & 0.32                                                                                     & 0.33                                                                                     & 0.31                                                                                     & 0.31                                                                                     & 0.31                                                                                     & 0.31                                                                                     & 0.31                                                                                     \\
            \midrule
            \multirow{4}{*}{\begin{tabular}[l]{@{}c@{}}CoSLAM\\~\cite{Wang2023CoSLAMJC}\end{tabular}}     & ATE RMSE (cm)            & 10.71                                                                                   & 10.88                                                                                   & 26.64                                                                                   & 10.02                                                                                    & 13.03                                                                                    & 30.75                                                                                    & 7.96                                                                                     & 14.37                                                                                    & 17.88                                                                                    & 15.80                                                                                    \\
                                                                                                          & Tracking (ms) $\uparrow$ & 5.46$\times$15                                                                          & 5.50$\times$15                                                                          & 5.39$\times$15                                                                          & 5.51$\times$15                                                                           & 5.09$\times$15                                                                           & 5.15$\times$15                                                                           & 7.55$\times$15                                                                           & 7.48$\times$15                                                                           & 7.62$\times$15                                                                           & 6.08$\times$15                                                                           \\
                                                                                                          & Mapping (ms) $\uparrow$  & 9.76$\times$15                                                                          & 12.84$\times$15                                                                         & 12.38$\times$15                                                                         & 11.29$\times$15                                                                          & 11.47$\times$15                                                                          & 14.07$\times$15                                                                          & 16.63$\times$15                                                                          & 16.61$\times$15                                                                          & 16.65$\times$15                                                                          & 13.52$\times$15                                                                          \\
                                                                                                          & FPS $\uparrow$           & 12.22                                                                                   & 12.12                                                                                   & 12.37                                                                                   & 12.09                                                                                    & 13.11                                                                                    & 12.95                                                                                    & 8.83                                                                                     & 8.91                                                                                     & 8.75                                                                                     & 11.26                                                                                    \\
            \midrule
            \multirow{4}{*}{\begin{tabular}[l]{@{}c@{}}ESLAM\\~\cite{Johari2022ESLAMED}\end{tabular}}     & ATE RMSE (cm)            & 10.72                                                                                   & 15.55                                                                                   & 40.42                                                                                   & 9.99                                                                                     & 12.79                                                                                    & 12.39                                                                                    & 7.01                                                                                     & 15.07                                                                                    & 7.97                                                                                     & 14.66                                                                                    \\
                                                                                                          & Tracking (ms) $\uparrow$ & 2.90$\times$10                                                                          & 5.34$\times$10                                                                          & 5.30$\times$10                                                                          & 5.18$\times$15                                                                           & 5.16$\times$15                                                                           & 5.30$\times$15                                                                           & 5.33$\times$15                                                                           & 5.22$\times$15                                                                           & 5.32$\times$15                                                                           & 5.20$\times$13                                                                           \\
                                                                                                          & Mapping (ms) $\uparrow$  & 17.95$\times$10                                                                         & 17.83$\times$10                                                                         & 18.20$\times$10                                                                         & 15.00$\times$10                                                                          & 15.02$\times$10                                                                          & 15.11$\times$10                                                                          & 17.07$\times$10                                                                          & 17.02$\times$10                                                                          & 16.92$\times$10                                                                          & 16.68$\times$10                                                                          \\
                                                                                                          & FPS $\uparrow$           & \textbf{19.18}                                                                          & \textbf{18.71}                                                                          & \textbf{18.86}                                                                          & 12.87                                                                                    & 12.92                                                                                    & 12.58                                                                                    & 12.51                                                                                    & 12.76                                                                                    & 12.53                                                                                    & 14.77                                                                                    \\
            \midrule

                                                                                                          & Acc (cm) $\downarrow$    & \textbf{9.62}                                                                           & \textbf{9.72}                                                                           & \textbf{9.94}                                                                           & \textbf{8.62}                                                                            & \textbf{8.77}                                                                            & \textbf{9.21}                                                                            & \textbf{6.74}                                                                            & \textbf{7.51}                                                                            & \textbf{6.94}                                                                            & \textbf{8.56}                                                                            \\

                                                                                                          & Tracking (ms) $\uparrow$ & 5.64$\times$10                                                                          & 5.83$\times$10                                                                          & 5.65$\times$10                                                                          & 5.76$\times$10                                                                           & 5.69$\times$10                                                                           & 5.77$\times$10                                                                           & 5.96$\times$10                                                                           & 5.80$\times$10                                                                           & 5.63$\times$10                                                                           & 5.75$\times$10                                                                           \\

                                                                                                          & Mapping (ms) $\uparrow$  & 13.02$\times$10                                                                         & 13.33$\times$10                                                                         & 13.07$\times$10                                                                         & 13.16$\times$10                                                                          & 13.23$\times$10                                                                          & 13.21$\times$10                                                                          & 13.36$\times$10                                                                          & 13.04$\times$10                                                                          & 12.98$\times$10                                                                          & 13.16$\times$10                                                                          \\

            \multirow{-4}{*}{Ours}                                                                        & FPS $\uparrow$           & 17.71                                                                                   & 17.15                                                                                   & 17.71                                                                                   & \textbf{17.37}                                                                           & \textbf{17.59}                                                                           & \textbf{17.34}                                                                           & \textbf{16.77}                                                                           & \textbf{17.23}                                                                           & \textbf{17.76}                                                                           & \textbf{17.40}                                                                           \\
            \bottomrule
        \end{tabular}
    }
    \vspace{-4ex}
\end{table*}

\boldparagraph{Hyperparameters.}
\ours run at $17$ FPS and sample $1024$ and $2048$ rays in tracking and BA stages with 10 iterations by default. The event joint global BA is performed every 5 frames with 5\% of pixels from all keyframes. The model is trained using Adam optimizer with learning rate $lr_{rot} = 1e^{-3}, lr_{trans} = 1e^{-3}$, and loss weights $\lambda_{ev} = 0.05, \lambda_{rgb} = 5.0, \lambda_{d} = 0.1, \lambda_{sdf} = 1000.0, \lambda_{sf} = 10$. The adaptive event forward query window $w_d$ and neighborhood window $w_k$ are set as $10$ and $5$ in DEV-Indoors, DEV-Indoors, and the fast subsets of DEV-Reals. Loss threshold $\mathcal{L}_s$ is set as $0.08$ by default and $0.1$ for DEV-Reals. The patch size of probability-weighted sampling is set as $32 \times 32$ for both RGB and event cameras. The event threshold $C$ is set as $0.2$ for the synthetic DEV-Indoors dataset and performs a normalization for real datasets DEV-Reals and Vector. For the camera distortion, we do not perform a pixel-wised undistortion but remove the distortion for each ray of both the RGBD camera and event camera.

We use Realsense RGB frames in DEV-Real for higher resolution compared to DAVIS. The pseudo-exposure is a \textbf{equivalent exposure time} of the event CRF rendering model. \ours renders logarithmic brightness in~\cref{eq:joint_event,eq:event_loss} at $t_\alpha$ and $t_\beta$ rather than all events between $t_\alpha$ and $t_\beta$. Thus, we do not focus on the intrinsic exposure of the event camera but on the equivalent exposure time for volume rendering and training. 

For DEV-Reals capture, we enable the auto-exposure to obtain a suitable exposure time and fixed it in a constant, \ie, 7.5 ms for normal scenes and 30 ms for the dark, to ensure the data match the algorithm inputs and support the validation. However, we enable the auto-gain and model the differentiable ISP through neural networks, as mentioned in~\cref{sec:decomposition} and \cite{Szeliski2010ComputerVA,Huang2021HDRNeRFHD}.

\section{Additional Experimental Results}
\label{sec:add_exp}
\subsection{More Ablation Studies}
\noindent{\textbf{Effect of the Event Temporal Aggregating Optimization Strategy.}}
To evaluate the effect of each component of the event temporal aggregating optimization strategy (ETA), we conduct an ablation study on the \texttt{\#Rm~blur} subset of DEV-Indoors and \texttt{\#Dorm2} subset of DEV-Reals. We investigate the performance using a constant interval of 5 frames and 10 frames for forward query, as well as utilize the proposed adaptive query in~\cref{tab:ETA}. The results show that the query interval is critical for \ours. The adaptive query strategy can significantly reduce the tracking ATE by 1.5 cm on \texttt{\#Rm~blur} and 16.08 cm on \texttt{\#Dorm2}, compared with the constant query interval of 5, respectively. In addition, the implementation with \#10 interval is better than \#5 interval by providing a longer time window constraint for the event temporal aggregating optimization, but still worse than the adaptive query strategy. The reason is that the event temporal aggregating optimization is sensitive, and the adaptive query strategy can adaptively select events to participate in optimization based on the loss, providing more robust local constraints thus reducing the impact of noise on optimization. Besides, \cref{tab:ETA} also shows that the full model surpasses the model w/o PWS by 0.25 and 1.9\% in ATE and completion on \texttt{\#Rm~blur}. For the effectiveness of ETA, our full model achieves lower tracking errors of 9.61 and 15.47 than the model w/o ETA on the \texttt{\#Rm~blur} and \texttt{\#Dorm2}, respectively.

\begin{table}[htbp]
    \centering
    \footnotesize
    \scriptsize
    \caption{\textbf{Ablation study of ETA} on the \texttt{\#Rm~blur} and \texttt{\#Drom2} subset of DEV-Indoors and DEV-Reals (15 iterations).}
    \label{tab:ETA}
    \resizebox{0.48\textwidth}{!}{
        \begin{tabular}{lccccccc}
            \toprule
            \multirow{2}{*}{Setting} & \multicolumn{4}{c}{\texttt{\#Rm~blur}} &                   & \multicolumn{2}{c}{\texttt{\#Dorm2}}                                                                     \\ \cline{2-5} \cline{7-8}
                                     & ATE$\downarrow$                        & ACC$\downarrow$   & Comp$\downarrow$                     & Comp ratio$\uparrow$ &  & Median$\downarrow$ & RSME$\downarrow$   \\
            \midrule
            Forward Query \#5        & 11.11                                  & 8.54              & \nd 8.51                             & \nd 83.21            &  & 27.99              & 28.99              \\
            Forward Query \#0        & 10.45                                  & 8.23              & 8.60                                 & 82.62                &  & \nd 12.50          & \nd 14.15          \\
            w/o PWS                  & \nd 9.86                               & \nd 7.88          & 9.49                                 & 81.04                &  & 16.59              & 19.78              \\
            w/o ETA                  & 11.89                                  & 8.61              & 10.98                                & 76.31                &  & 14.46              & 18.75              \\
            Full ETA                 & \fs \textbf{9.61}                      & \fs \textbf{7.88} & \fs \textbf{7.59}                    & \fs \textbf{83.51}   &  & \fs \textbf{11.91} & \fs \textbf{15.47} \\
            \bottomrule
        \end{tabular}
    }
\end{table}

\subsection{More Detailed Tracking Comparison}
\label{sec:tracking_iteration}
In this section, we further provide the accuracy of tracking and its corresponding iteration settings, as well as the runtime. Note that it is unrealistic to strictly control all the iterations or FPS to be the same. Therefore, all the methods are compared under similar runtimes. Besides, we must emphasize that we had to increase the iteration number for certain methods to avoid crashes. Nevertheless, \ours still achieves superior accuracy with less time-consuming.

\begin{table*}[t]
    \centering
    \vspace{-2ex}
    \setlength{\tabcolsep}{8pt}
    \renewcommand{\arraystretch}{0.9}
    \caption{\textbf{Tracking (ATE median [cm]) and run-time comparison} with detailed iteration setting of the proposed method vs. the SOTA methods on \textbf{DEV-Reals}. Our method achieves better performance in comparison to NICE-SLAM~\cite{Zhu2021NICESLAMNI}, CoSLAM~\cite{Wang2023CoSLAMJC} and ESLAM~\cite{Johari2022ESLAMED}.}
    \label{tab:tracking_devreals_iter}
    \resizebox{1.0\linewidth}{!}{
        \begin{tabular}{clrrrrrrrrr}
            \toprule
            Method                                                                                        & Metric                     & \texttt{\#Pio1}   & \texttt{\#Pio2}   & \texttt{\#Gre1}    & \texttt{\#Gre2}    & \texttt{\#dorm1}  & \texttt{\#dorm2}  & \texttt{\#dorm3}  & \texttt{\#dorm4}  & \texttt{\textbf{\#avg}} \\
            \midrule
            \multirow{4}{*}{\begin{tabular}[l]{@{}c@{}}NICE-SLAM\\~\cite{Zhu2021NICESLAMNI}\end{tabular}} & ATE RMSE (cm) $\downarrow$ & 13.21             & 23.35             & \blackx63\%        & \blackx25\%        & 24.69             & \textbf{10.68}    & 18.44             & 44.04             & \blackx22.40            \\
                                                                                                          & Tracking (ms) $\uparrow$   & 3.08$\times$100   & 3.61$\times$100   & \blackx$\times$100 & \blackx$\times$100 & 3.08$\times$100   & 3.15$\times$100   & 3.18$\times$100   & 3.17$\times$100   & 3.21$\times$100         \\
                                                                                                          & Mapping (ms) $\uparrow$    & 2.97$\times$60    & 2.57$\times$60    & \blackx$\times$60  & \blackx$\times$60  & 3.86$\times$60    & 3.97 $\times$60   & 3.27$\times$60    & 3.20$\times$60    & 3.31$\times$60          \\
                                                                                                          & FPS $\uparrow$             & 0.28              & 0.28              & \blackx            & \blackx            & 0.31              & 0.32              & 0.32              & 0.32              & 0.31                    \\
            \midrule
            \multirow{4}{*}{\begin{tabular}[l]{@{}c@{}}CoSLAM\\~\cite{Wang2023CoSLAMJC}\end{tabular}}     & ATE RMSE (cm) $\downarrow$ & 11.14             & 19.83             & 82.52              & 40.16              & 15.99             & 15.42             & 30.12             & 32.45             & 30.95                   \\
                                                                                                          & Tracking (ms) $\uparrow$   & 8.87$\times$20    & 8.90$\times$20    & 8.96$\times$20     & 8.89$\times$20     & 8.87$\times$20    & 9.09$\times$20    & 9.03$\times$20    & 9.08$\times$20    & 8.96$\times$20          \\
                                                                                                          & Mapping (ms) $\uparrow$    & 14.86 $\times$ 20 & 14.84 $\times$ 20 & 14.97 $\times$ 20  & 14.71 $\times$ 20  & 15.33 $\times$ 20 & 14.83 $\times$ 20 & 16.09 $\times$ 20 & 15.41$\times$20   & 15.13$\times$20         \\
                                                                                                          & FPS $\uparrow$             & 5.64              & 5.62              & 5.58               & 5.63               & 5.64              & 5.50              & 5.54              & 5.51              & 5.58                    \\
            \midrule
            \multirow{4}{*}{\begin{tabular}[l]{@{}c@{}}ESLAM\\~\cite{Johari2022ESLAMED}\end{tabular}}     & ATE RMSE (cm) $\downarrow$ & 11.28             & 21.42             & 63.65              & 30.75              & 37.94             & 31.04             & 16.19             & 37.91             & 31.27                   \\
                                                                                                          & Tracking (ms) $\uparrow$   & 5.11 $\times$ 20  & 5.15 $\times$ 20  & 5.08 $\times$ 20   & 5.16 $\times$ 20   & 4.84 $\times$ 20  & 4.93 $\times$ 20  & 4.92 $\times$ 20  & 4.84 $\times$ 20  & 5.00$\times$20          \\
                                                                                                          & Mapping (ms) $\uparrow$    & 17.85$\times$ 20  & 17.6$\times$ 20   & 17.4$\times$ 20    & 18.4$\times$ 20    & 17.$\times$ 20    & 19.05$\times$ 20  & 16.2$\times$ 20   & 16.46 $\times$ 20 & 17.50$\times$20         \\
                                                                                                          & FPS $\uparrow$             & 9.76              & 9.70              & 9.83               & 9.68               & 10.31             & 10.13             & 10.15             & 10.33             & 9.99                    \\
            \midrule

            \multirow{4}{*}{\begin{tabular}[l]{@{}c@{}}ENSLAM\\(Ours)\end{tabular}}                       & ATE RMSE (cm) $\downarrow$ & \textbf{8.94}     & \textbf{19.05}    & \textbf{43.63}     & \textbf{21.18}     & \textbf{11.26}    & 11.91             & \textbf{16.00}    & \textbf{19.78}    & \textbf{18.97}          \\
                                                                                                          & Tracking (ms) $\uparrow$   & 5.75$\times$15    & 5.88$\times$15    & 5.59$\times$15     & 5.91$\times$15     & 5.34$\times$15    & 5.78$\times$15    & 5.77$\times$15    & 6.44$\times$15    & 5.81$\times$15          \\
                                                                                                          & Mapping (ms) $\uparrow$    & 14.00$\times$15   & 14.70$\times$15   & 14.97$\times$15    & 14.23$\times$15    & 14.90$\times$15   & 13.79$\times$15   & 14.35$\times$15   & 15.32$\times$15   & 14.53$\times$15         \\
                                                                                                          & FPS $\uparrow$             & \textbf{11.59}    & \textbf{11.33}    & \textbf{11.92}     & \textbf{11.28}     & \textbf{12.48}    & \textbf{11.53}    & \textbf{11.55}    & \textbf{10.35}    & \textbf{11.50}          \\
            \bottomrule
        \end{tabular}
    }
\end{table*}

\begin{table*}[htbp]
    \centering
    \setlength{\tabcolsep}{8pt}
    \renewcommand{\arraystretch}{0.8}
    \caption{\textbf{Tracking (ATE mean [cm])} with detailed iteration setting of the proposed method vs. the SOTA NeRF-based methods on \textbf{Vector}\cite{Gao2022VECtorAV} dataset. \ours achieves better accuracy and efficiency compared with CoSLAM~\cite{Wang2023CoSLAMJC} and ESLAM~\cite{Johari2022ESLAMED} in most scenes.}
    \label{tab:tracking_vector_iter}
    \resizebox{1.0\linewidth}{!}{
        \begin{tabular}{clrrrrrrrrr}
            \toprule
            Method                                                                                    & Metric                     & \begin{tabular}[c]{@{}c@{}}\texttt{robot}\\ \texttt{norm}\end{tabular} & \begin{tabular}[c]{@{}c@{}}\texttt{robot}\\ \texttt{fast}\end{tabular} & \begin{tabular}[c]{@{}c@{}}\texttt{desk}\\\texttt{norm}\end{tabular} & \begin{tabular}[c]{@{}c@{}}\texttt{desk}\\ \texttt{fast}\end{tabular} & \begin{tabular}[c]{@{}c@{}}\texttt{sofa}\\ \texttt{norm}\end{tabular} & \begin{tabular}[c]{@{}c@{}}\texttt{sofa}\\\texttt{fast}\end{tabular} & \begin{tabular}[c]{@{}c@{}}\texttt{hdr}\\\texttt{norm}\end{tabular} & \begin{tabular}[c]{@{}c@{}}\texttt{hdr}\\\texttt{fast}\end{tabular} & \begin{tabular}[c]{@{}c@{}}\texttt{\textbf{\#all}}\\\texttt{\textbf{avg}}\end{tabular} \\
            \midrule
            \multirow{4}{*}{\begin{tabular}[l]{@{}c@{}}CoSLAM\\~\cite{Wang2023CoSLAMJC}\end{tabular}} & ATE RMSE (cm) $\downarrow$ & \textbf{1.00}                                                          & 124.69                                                                 & \textbf{1.76}                                                        & 97.65                                                                 & \textbf{1.74}                                                         & 77.89                                                                & 1.47                                                                & 1.42                                                                & 38.45                                                                                                      \\
                                                                                                      & Tracking (ms) $\uparrow$   & 59.74 $\times$ 10                                                      & 5.99 $\times$ 10                                                       & 5.51 $\times$ 10                                                     & 5.67 $\times$ 10                                                      & 5.55 $\times$ 10                                                      & 5.47 $\times$ 10                                                     & 5.55 $\times$ 10                                                    & 5.80 $\times$ 10                                                    & 5.69                                                                                                       \\
                                                                                                      & Mapping (ms) $\uparrow$    & 11.44 $\times$ 10                                                      & 11.18 $\times$ 10                                                      & 10.41 $\times$ 10                                                    & 11.18 $\times$ 10                                                     & 12.12 $\times$ 10                                                     & 16.90 $\times$ 10                                                    & 14.32 $\times$ 10                                                   & 11.15 $\times$ 10                                                   & 12.34                                                                                                      \\
                                                                                                      & FPS $\uparrow$             & 16.74                                                                  & 16.69                                                                  & 18.16                                                                & 17.63                                                                 & 18.02                                                                 & 18.29                                                                & 18.03                                                               & 17.24                                                               & 17.60                                                                                                      \\
            \midrule
            \multirow{4}{*}{\begin{tabular}[l]{@{}c@{}}ESLAM\\~\cite{Johari2022ESLAMED}\end{tabular}} & ATE RMSE (cm) $\downarrow$ & 1.39                                                                   & 3.30                                                                   & 2.54                                                                 & 3.64                                                                  & 7.99                                                                  & 19.03                                                                & 7.38                                                                & 12.23                                                               & 7.19                                                                                                       \\
                                                                                                      & Tracking (ms) $\uparrow$   & 4.94 $\times$ 20                                                       & 4.96 $\times$ 20                                                       & 4.96 $\times$ 20                                                     & 4.67 $\times$ 20                                                      & 4.85 $\times$ 20                                                      & 5.00 $\times$ 20                                                     & 5.10 $\times$ 20                                                    & 4.91 $\times$ 20                                                    & 4.93 $\times$ 20                                                                                           \\
                                                                                                      & Mapping (ms) $\uparrow$    & 18.68$\times$20                                                        & 19.49$\times$20                                                        & 17.07$\times$20                                                      & 18.69$\times$20                                                       & 17.97$\times$20                                                       & 17.57$\times$20                                                      & 18.16$\times$20                                                     & 18.08$\times$20                                                     & 18.22 $\times$ 20                                                                                          \\
                                                                                                      & FPS $\uparrow$             & 10.11                                                                  & 10.06                                                                  & 10.07                                                                & 10.69                                                                 & 10.30                                                                 & 9.98                                                                 & 9.79                                                                & 10.16                                                               & 10.15                                                                                                      \\
            \midrule
            \multirow{4}{*}{\begin{tabular}[l]{@{}c@{}}ENSLAM\\(Ours)\end{tabular}}                   & ATE RMSE (cm) $\downarrow$ & 1.06                                                                   & \textbf{1.73}                                                          & \textbf{1.76}                                                        & \textbf{2.69}                                                         & 2.02                                                                  & \textbf{1.84}                                                        & \textbf{1.03}                                                       & \textbf{1.22}                                                       & \textbf{1.67}                                                                                              \\
                                                                                                      & Tracking (ms) $\uparrow$   & 5.58 $\times$ 10                                                       & 5.91 $\times$ 10                                                       & 5.81 $\times$ 10                                                     & 6.01 $\times$ 10                                                      & 5.74 $\times$ 10                                                      & 6.01 $\times$ 10                                                     & 5.76 $\times$ 10                                                    & 6.12 $\times$ 10                                                    & 5.87                                                                                                       \\
                                                                                                      & Mapping (ms) $\uparrow$    & 19.05 $\times$ 10                                                      & 17.07 $\times$ 10                                                      & 18.05 $\times$ 10                                                    & 16.28 $\times$ 10                                                     & 13.91 $\times$ 10                                                     & 13.22 $\times$ 10                                                    & 13.42 $\times$ 10                                                   & 13.76                                                               & 15.60                                                                                                      \\
                                                                                                      & FPS $\uparrow$             & \textbf{17.92}                                                         & \textbf{16.92}                                                         & \textbf{17.21}                                                       & \textbf{16.63}                                                        & \textbf{17.42}                                                        & \textbf{16.63}                                                       & \textbf{17.36}                                                      & \textbf{16.33}                                                      & \textbf{17.05}                                                                                             \\
            \bottomrule
        \end{tabular}
    }
    \vspace{-2ex}
\end{table*}

\boldparagraph{Tracking Comparison on DEV-Indoors.} We provide the detailed iterations and corresponding FPS of the tracking evaluation on the DEV-Indoors dataset in~\cref{tab:tracking_devindoors_iter}. The results show that our method is more efficient and accurate than existing NeRF-based SLAM methods. Specifically, our method reduces the tracking ATE by 23.9, 7.24, and 6.1 cm, compared with the SOTA methods NICE-SLAM~\cite{Zhu2021NICESLAMNI}, CoSLAM~\cite{Wang2023CoSLAMJC} and ESLAM~\cite{Johari2022ESLAMED}, respectively. In addition, all the other methods face significant challenges from \#norm subsets to \#blur and \#dark scenarios, with a serious decline in accuracy. Hence, we must increase the tracking or mapping iteration times for some baselines to avoid crushes but slow down the FPS. In contrast, our method uses the invariant iterations 10 times for both tracking and mapping and maintains fast, robust, and accurate results.

\boldparagraph{Tracking Comparison on DEV-Reals.} In the main paper, we only report the final tracking ATE. Hence, we further show the detailed performance with tracking and mapping iterations in~\cref{tab:tracking_devreals_iter}. \ours uses 15 iterations for both tracking and mapping and achieves the best performance in accuracy and efficiency in the challenging DEV-Reals dataset. In contrast, the other methods perform worse with an event larger iteration number.

\boldparagraph{Tracking Comparison on Vector.}~\cref{tab:tracking_vector_iter} illustrates the tracking ATE and iterations on Vector~\cite{Gao2022VECtorAV} dataset. \ours, CoSLAM~\cite{Wang2023CoSLAMJC} and ESLAM~\cite{Johari2022ESLAMED} set the iterations as 10, 20 and 10 in both tracking and mapping, respectively. CoSLAM and \ours perform comparably in the normal subsets, but \ours significantly surpasses CoSLAM on the fast subsets, benefitting from the high-quality event data.

\subsection{Additional Reconstruction Visualization}
\label{sec:add_viz}
\begin{figure*}[t]
    \centering
    {
        \footnotesize
        \setlength{\tabcolsep}{0.7pt}
        \renewcommand{\arraystretch}{0}
        \newcommand{\sz}{0.055}
        \begin{tabular}{cccccc}
                                                 & NICE-SLAM~\cite{Zhu2021NICESLAMNI}                                                                                     & CoSLAM~\cite{Wang2023CoSLAMJC}                                                                                       & ESLAM~\cite{Johari2022ESLAMED}                                                                                      & ENSLAM (Ours)                                                                                                      & Ground Truth Mesh                                                                                                \\
            \rotatebox[origin=c]{90}{\#Rm Blur}  & \includegraphics[valign=c,height=\sz\linewidth]{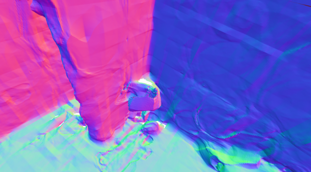}            & \includegraphics[valign=c,height=\sz\linewidth]{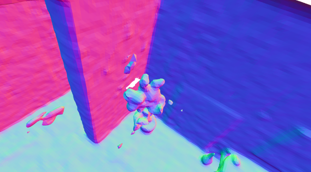}            & \includegraphics[valign=c,height=\sz\linewidth]{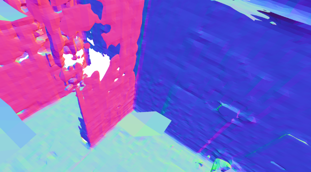}            & \includegraphics[valign=c,height=\sz\linewidth]{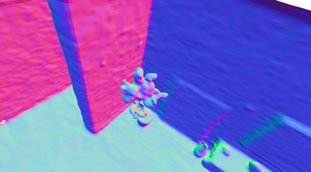}            & \includegraphics[valign=c,height=\sz\linewidth]{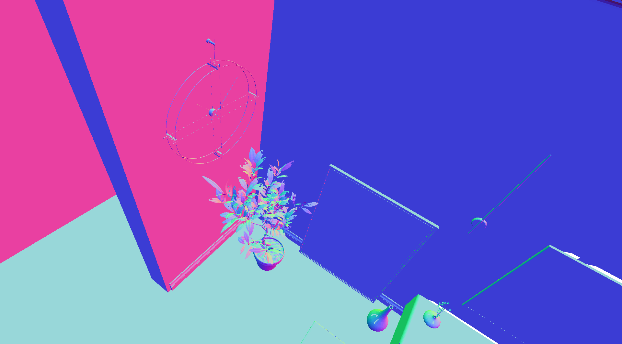}            \\
            \addlinespace[2pt]
            \rotatebox[origin=c]{90}{\#Rm Dark}  & \includegraphics[valign=c,height=\sz\linewidth]{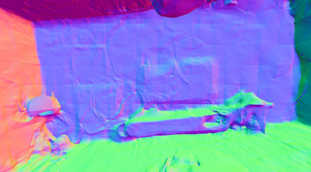}            & \includegraphics[valign=c,height=\sz\linewidth]{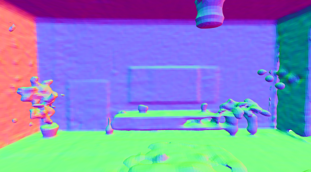}            & \includegraphics[valign=c,height=\sz\linewidth]{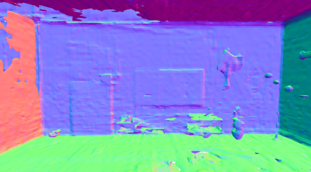}            & \includegraphics[valign=c,height=\sz\linewidth]{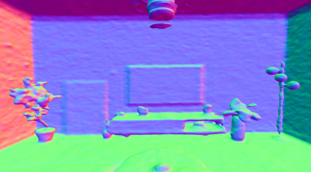}            & \includegraphics[valign=c,height=\sz\linewidth]{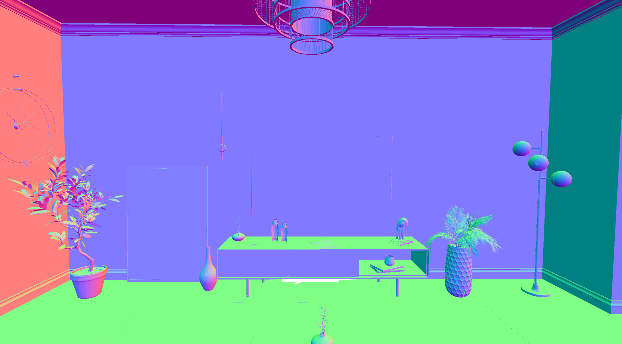}            \\
            \addlinespace[2pt]
            \rotatebox[origin=c]{90}{\#Apt Norm} & \includegraphics[valign=c,height=\sz\linewidth]{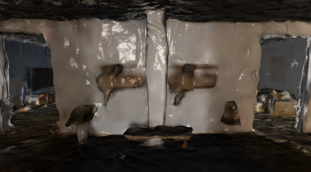} & \includegraphics[valign=c,height=\sz\linewidth]{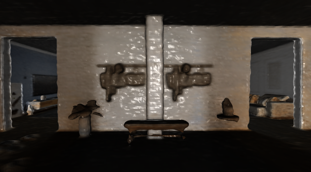} & \includegraphics[valign=c,height=\sz\linewidth]{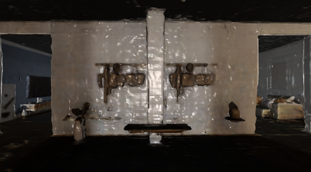} & \includegraphics[valign=c,height=\sz\linewidth]{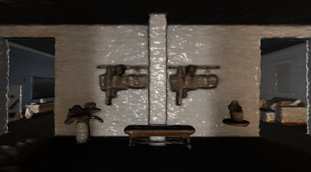} & \includegraphics[valign=c,height=\sz\linewidth]{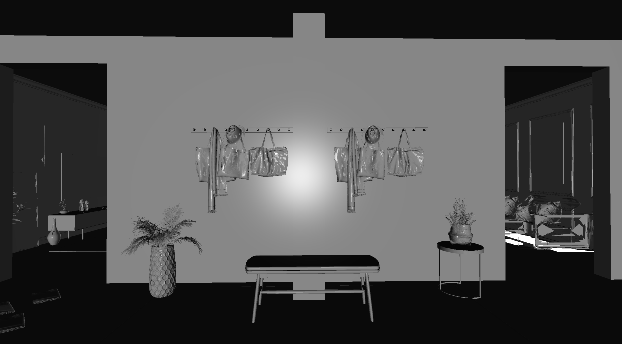} \\
            \addlinespace[2pt]
            \rotatebox[origin=c]{90}{\#Apt Blur} & \includegraphics[valign=c,height=\sz\linewidth]{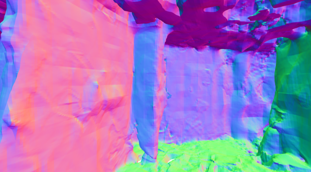}       & \includegraphics[valign=c,height=\sz\linewidth]{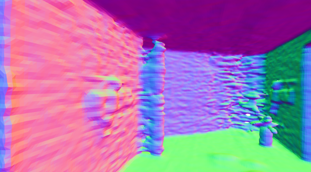}       & \includegraphics[valign=c,height=\sz\linewidth]{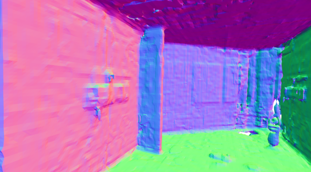}       & \includegraphics[valign=c,height=\sz\linewidth]{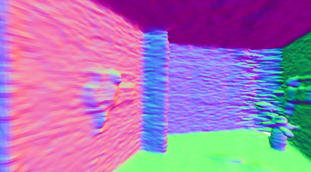}       & \includegraphics[valign=c,height=\sz\linewidth]{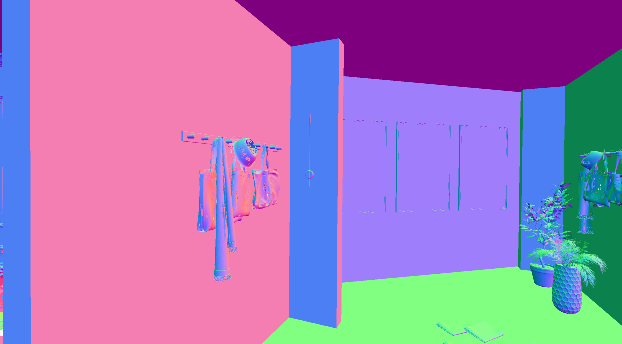}       \\
            \addlinespace[2pt]
            \rotatebox[origin=c]{90}{\#Apt Dark} & \includegraphics[valign=c,height=\sz\linewidth]{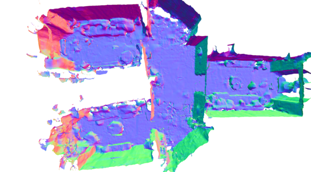}       & \includegraphics[valign=c,height=\sz\linewidth]{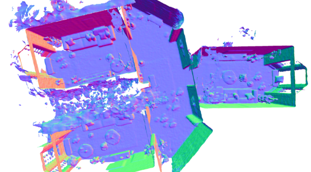}       & \includegraphics[valign=c,height=\sz\linewidth]{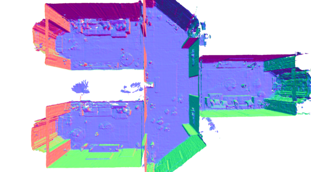}       & \includegraphics[valign=c,height=\sz\linewidth]{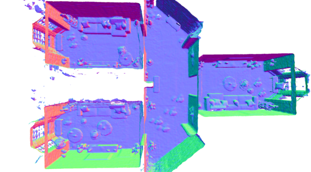}       & \includegraphics[valign=c,height=\sz\linewidth]{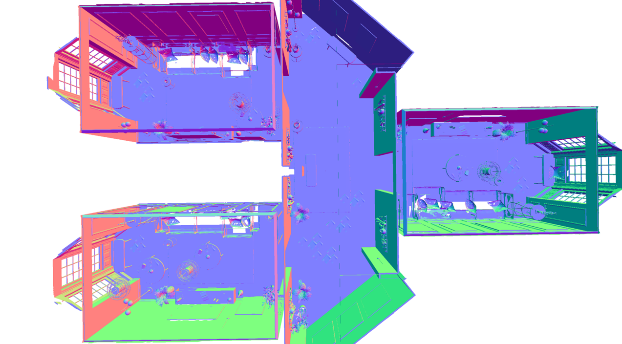}       \\
            \addlinespace[2pt]
            \rotatebox[origin=c]{90}{\#Wkp Norm} & \includegraphics[valign=c,height=\sz\linewidth]{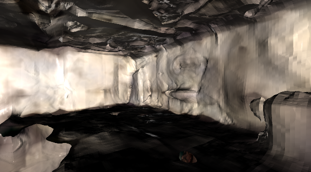}  & \includegraphics[valign=c,height=\sz\linewidth]{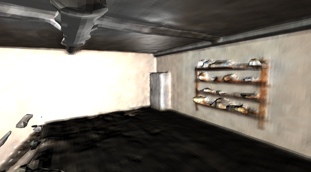}  & \includegraphics[valign=c,height=\sz\linewidth]{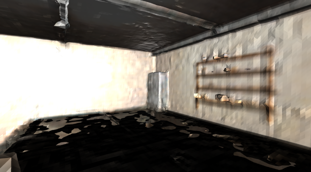}  & \includegraphics[valign=c,height=\sz\linewidth]{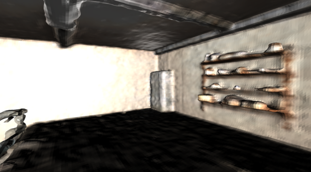}  & \includegraphics[valign=c,height=\sz\linewidth]{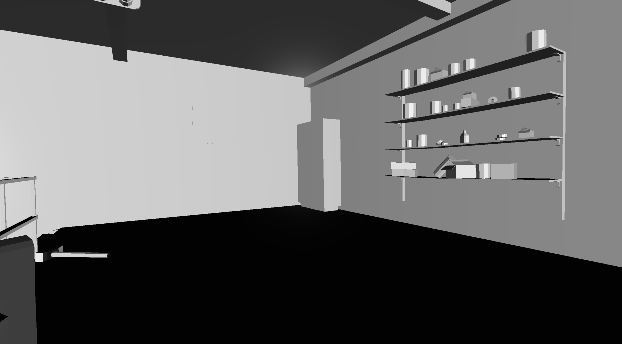}  \\
            \addlinespace[2pt]
            \rotatebox[origin=c]{90}{\#Wkp Blur} & \includegraphics[valign=c,height=\sz\linewidth]{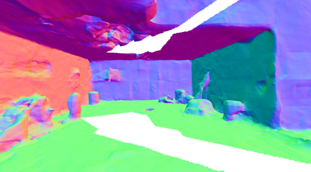}        & \includegraphics[valign=c,height=\sz\linewidth]{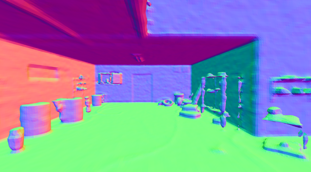}        & \includegraphics[valign=c,height=\sz\linewidth]{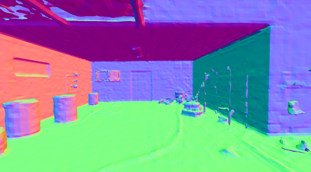}        & \includegraphics[valign=c,height=\sz\linewidth]{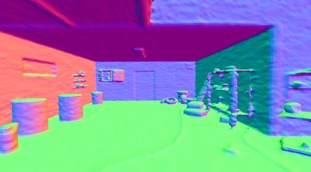}        & \includegraphics[valign=c,height=\sz\linewidth]{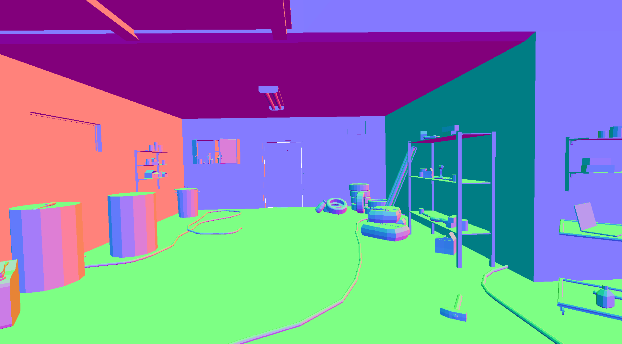}        \\
            \addlinespace[2pt]
            \rotatebox[origin=c]{90}{\#Wkp Dark} & \includegraphics[valign=c,height=\sz\linewidth]{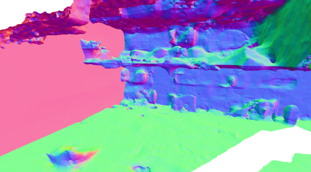}        & \includegraphics[valign=c,height=\sz\linewidth]{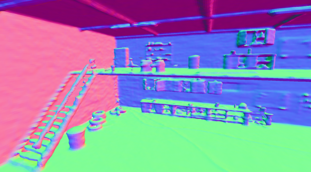}        & \includegraphics[valign=c,height=\sz\linewidth]{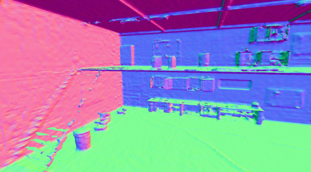}        & \includegraphics[valign=c,height=\sz\linewidth]{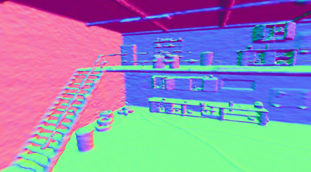}        & \includegraphics[valign=c,height=\sz\linewidth]{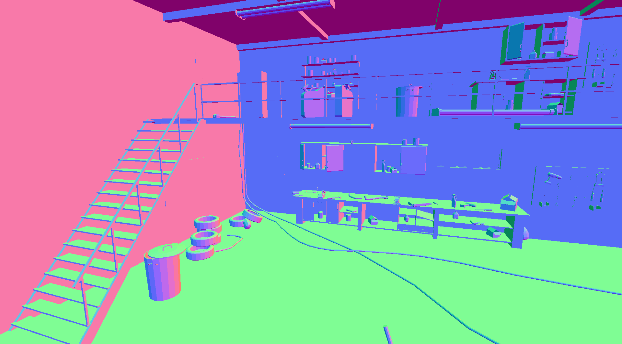}        \\
        \end{tabular}
    }
    \caption{Reconstruction Performance on \textbf{DEV-Indoors}. EN-SLAM achieves, on average, more precise reconstruction details than existing methods in motion blur and lighting-varying environments with the assistance of high-quality event streams.}
    \label{fig:add_viz_devindoors}
\end{figure*}

\boldparagraph{Reconstruction Visualization on DEV-Indoors.} \cref{fig:add_viz_devindoors} provides more mesh reconstruction results in DEV-Indoors dataset. Compared with the other SOTA methods, \ours significantly reduces the presence of holes and ghosting artifacts in reconstructed scenes under blurry scenarios, achieving higher-quality reconstruction results. Under the challenges of dark scenes, \eg, \#Apt Dark, previous methods NICE-SLAM and CoSLAM suffer from the weak supervision of color images, resulting in tracking drift. While \ours maintains robust and accurate.

\begin{figure*}[t]
    \centering
    {\footnotesize
        \setlength{\tabcolsep}{0.7pt}
        \renewcommand{\arraystretch}{0}
        \newcommand{\sz}{0.067}
        \begin{tabular}{ccccc}
                                                   & NICE-SLAM~\cite{Zhu2021NICESLAMNI}                                                                   & CoSLAM~\cite{Wang2023CoSLAMJC}                                                                     & ESLAM~\cite{Johari2022ESLAMED}                                                                    & ENSLAM (Ours)                                                                                    \\
            \rotatebox[origin=c]{90}{\#Pioffice1}  & \includegraphics[valign=c,height=\sz\linewidth]{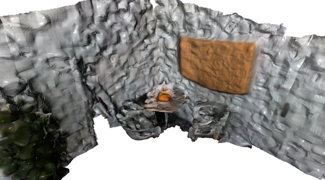} & \includegraphics[valign=c,height=\sz\linewidth]{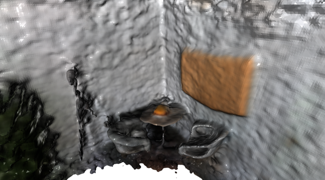} & \includegraphics[valign=c,height=\sz\linewidth]{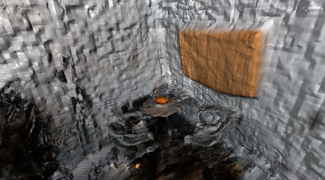} & \includegraphics[valign=c,height=\sz\linewidth]{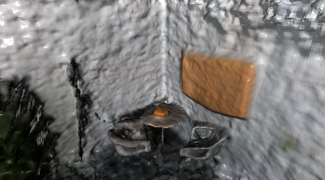} \\
            \addlinespace[2pt]
            \rotatebox[origin=c]{90}{\#Garage1}    & \includegraphics[valign=c,height=\sz\linewidth]{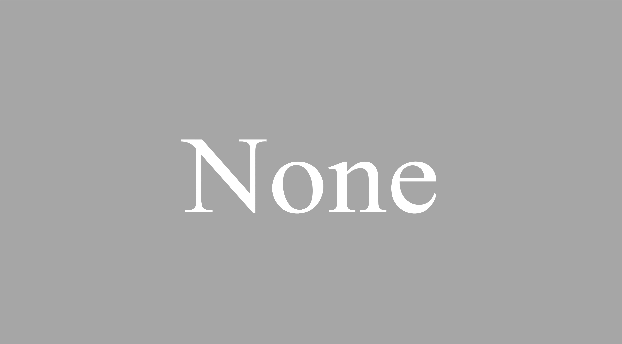} & \includegraphics[valign=c,height=\sz\linewidth]{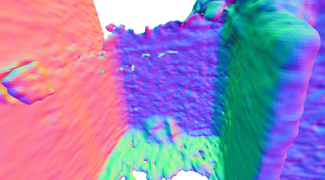} & \includegraphics[valign=c,height=\sz\linewidth]{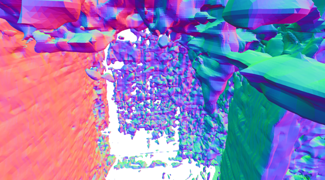} & \includegraphics[valign=c,height=\sz\linewidth]{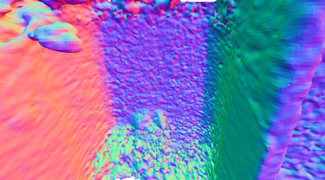} \\
            \addlinespace[2pt]
            \rotatebox[origin=c]{90}{\#Garage1}    & \includegraphics[valign=c,height=\sz\linewidth]{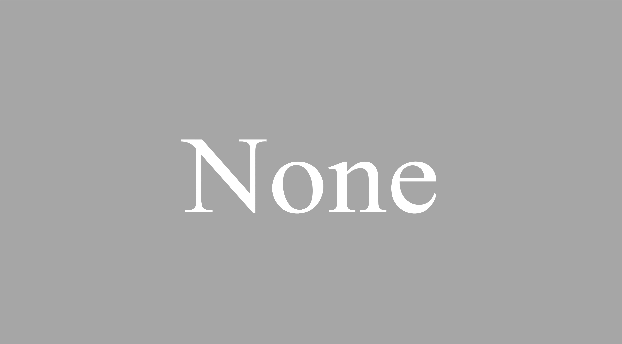} & \includegraphics[valign=c,height=\sz\linewidth]{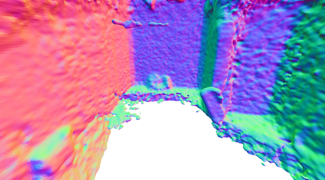} & \includegraphics[valign=c,height=\sz\linewidth]{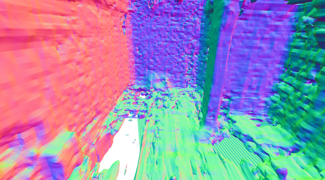} & \includegraphics[valign=c,height=\sz\linewidth]{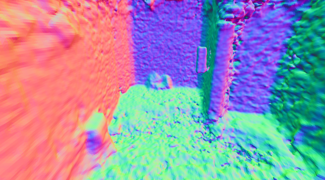} \\
            \addlinespace[2pt]
            \rotatebox[origin=c]{90}{\#Dormitory1} & \includegraphics[valign=c,height=\sz\linewidth]{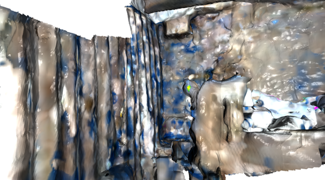} & \includegraphics[valign=c,height=\sz\linewidth]{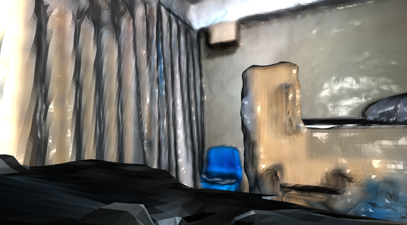} & \includegraphics[valign=c,height=\sz\linewidth]{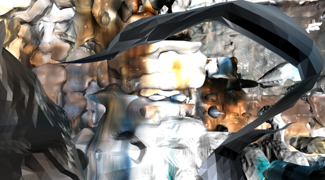} & \includegraphics[valign=c,height=\sz\linewidth]{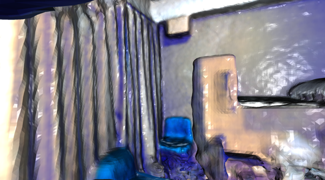} \\
            \addlinespace[2pt]
            \rotatebox[origin=c]{90}{\#Dormitory2} & \includegraphics[valign=c,height=\sz\linewidth]{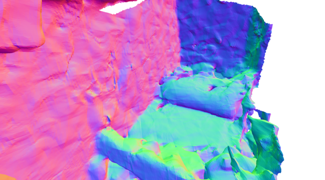} & \includegraphics[valign=c,height=\sz\linewidth]{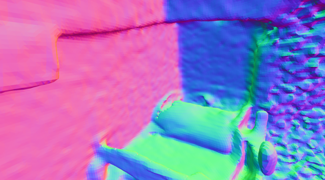} & \includegraphics[valign=c,height=\sz\linewidth]{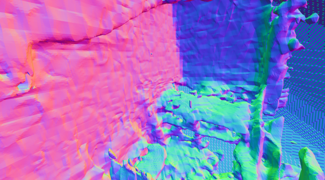} & \includegraphics[valign=c,height=\sz\linewidth]{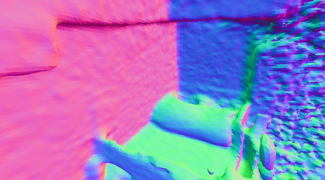} \\
            \addlinespace[2pt]
            \rotatebox[origin=c]{90}{\#Dormitory3} & \includegraphics[valign=c,height=\sz\linewidth]{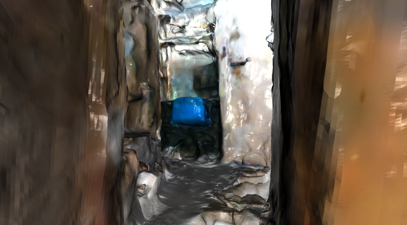} & \includegraphics[valign=c,height=\sz\linewidth]{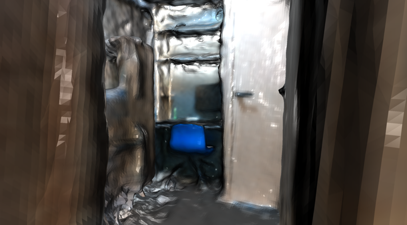} & \includegraphics[valign=c,height=\sz\linewidth]{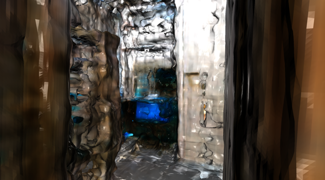} & \includegraphics[valign=c,height=\sz\linewidth]{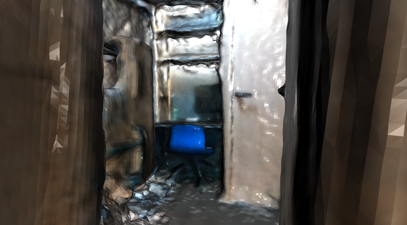} \\
            \addlinespace[2pt]
            \rotatebox[origin=c]{90}{\#Dormitory4} & \includegraphics[valign=c,height=\sz\linewidth]{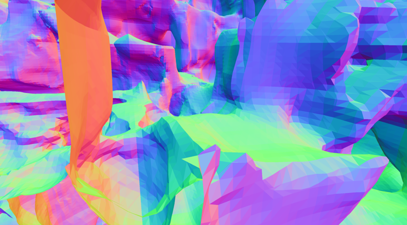} & \includegraphics[valign=c,height=\sz\linewidth]{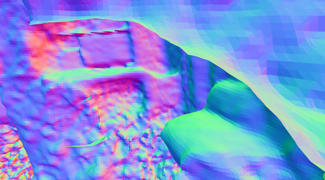} & \includegraphics[valign=c,height=\sz\linewidth]{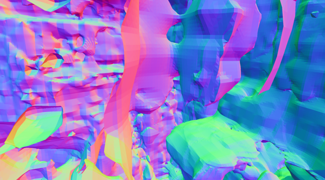} & \includegraphics[valign=c,height=\sz\linewidth]{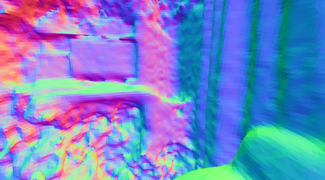} \\
        \end{tabular}
    }
    \caption{Reconstruction Performance on the challenging \textbf{DEV-Reals} dataset. EN-SLAM performs consistently well in all the subsets and obtains more satisfying reconstruction results compared with NICE-SLAM, CoSLAM and ESLAM.}
    \label{fig:add_viz_devreals}
\end{figure*}

\boldparagraph{Reconstruction Visualization on DEV-Reals.} \cref{fig:add_viz_devreals}
\cref{fig:add_viz_devreals} shows the map reconstruction comparison on the challenging DEV-Reals dataset. NICE-SLAM crushes in the \#Garage1 and \#Garage2 subsets due to the low-lighting environments. CoSLAM reconstructs all the scenarios but causes significant holes and artifacts in the mapping results. ESLAM performs relatively well in the \#Pioffice1 and \#Pioffice2 subsets but fails in the low-lighting subsets \#Garage1, \#Dormitory2, and \#Dormitory4 due to the low-quality color and depth images.  In contrast, \ours achieves the best performance in all the subsets, demonstrating its robustness and accuracy in the challenging DEV-Reals dataset.

\begin{figure*}[t]
    \vspace{-4ex}
    \centering
    {\footnotesize
        \setlength{\tabcolsep}{0.7pt}
        \renewcommand{\arraystretch}{0}
        \newcommand{\sz}{0.08}
        \begin{tabular}{cccc}
                                                      & CoSLAM~\cite{Wang2023CoSLAMJC}                                                                                  & ESLAM~\cite{Johari2022ESLAMED}                                                                                 & ENSLAM (Ours)                                                                                                 \\
            \rotatebox[origin=c]{90}{\#Robot normal1} & \includegraphics[valign=c,height=\sz\linewidth]{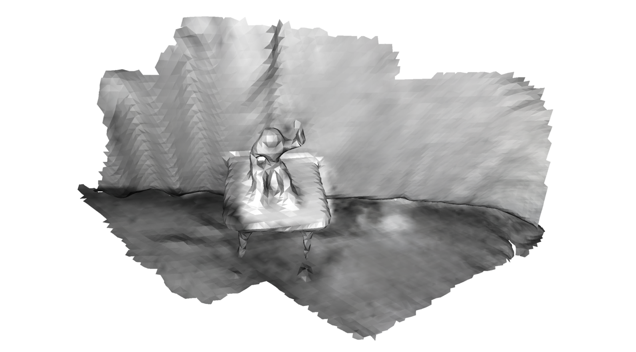} & \includegraphics[valign=c,height=\sz\linewidth]{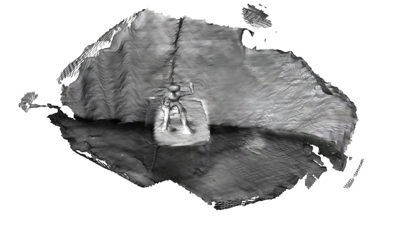} & \includegraphics[valign=c,height=\sz\linewidth]{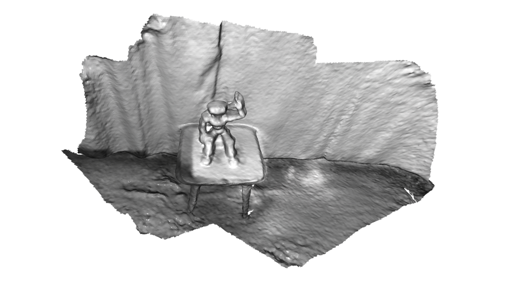} \\
            \rotatebox[origin=c]{90}{\#Robot fast1}   & \includegraphics[valign=c,height=\sz\linewidth]{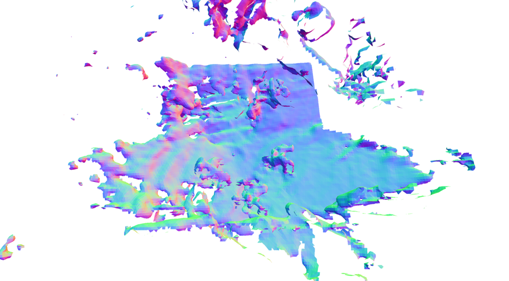}         & \includegraphics[valign=c,height=\sz\linewidth]{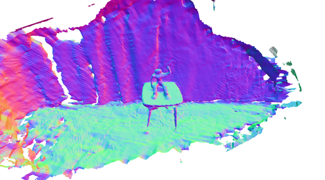}         & \includegraphics[valign=c,height=\sz\linewidth]{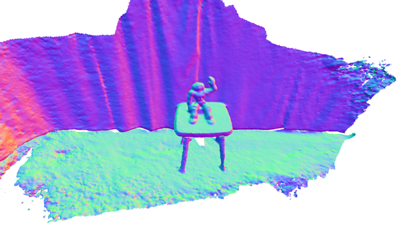}         \\
            \rotatebox[origin=c]{90}{\#Desk normal1}  & \includegraphics[valign=c,height=\sz\linewidth]{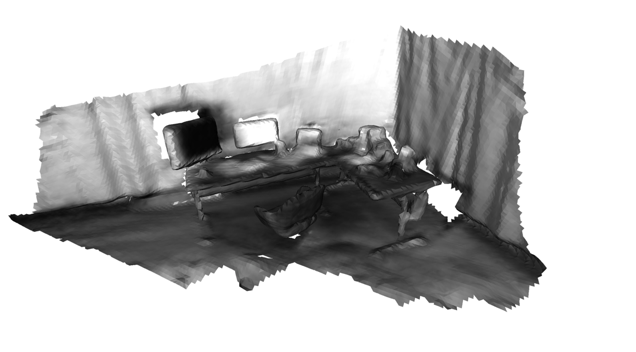}  & \includegraphics[valign=c,height=\sz\linewidth]{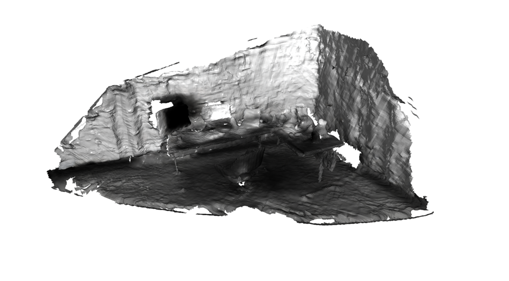}  & \includegraphics[valign=c,height=\sz\linewidth]{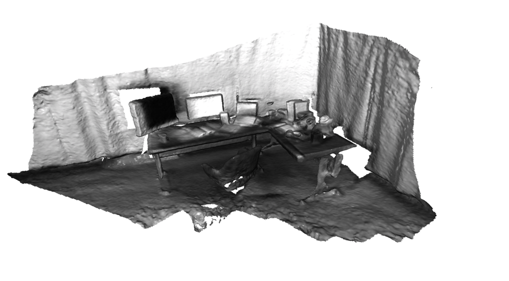}  \\
            \rotatebox[origin=c]{90}{\#Desk fast1}    & \includegraphics[valign=c,height=\sz\linewidth]{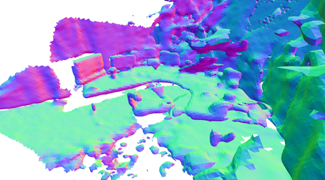}          & \includegraphics[valign=c,height=\sz\linewidth]{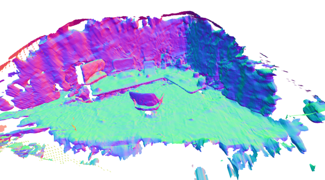}          & \includegraphics[valign=c,height=\sz\linewidth]{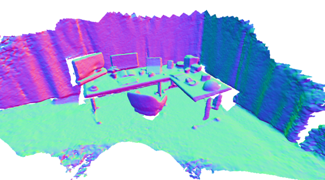}          \\
            \rotatebox[origin=c]{90}{\#Hdr normal1}   & \includegraphics[valign=c,height=\sz\linewidth]{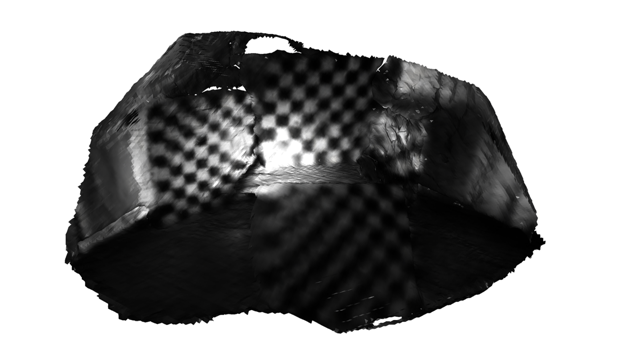}   & \includegraphics[valign=c,height=\sz\linewidth]{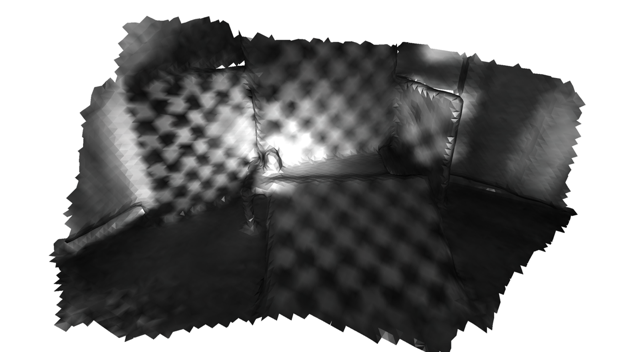}   & \includegraphics[valign=c,height=\sz\linewidth]{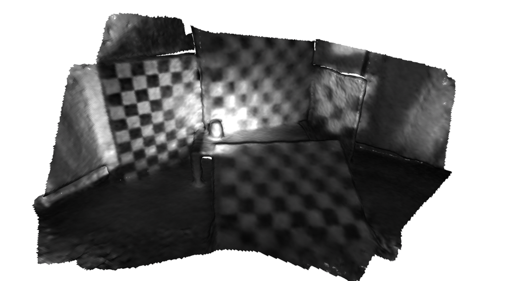}   \\
            \rotatebox[origin=c]{90}{\#Hdr fast1}     & \includegraphics[valign=c,height=\sz\linewidth]{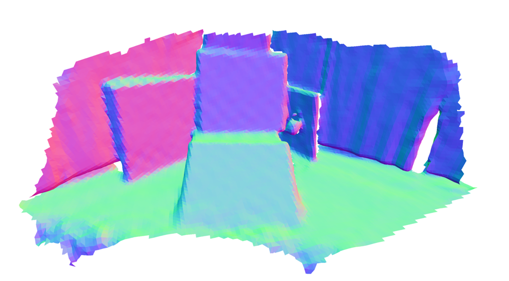}           & \includegraphics[valign=c,height=\sz\linewidth]{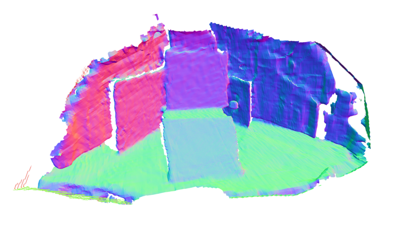}           & \includegraphics[valign=c,height=\sz\linewidth]{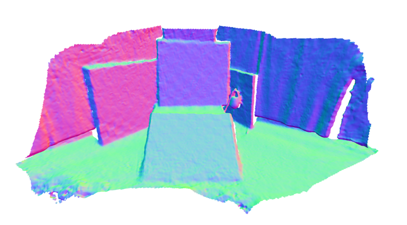}           \\
            \rotatebox[origin=c]{90}{\#Sofa normal1}  & \includegraphics[valign=c,height=\sz\linewidth]{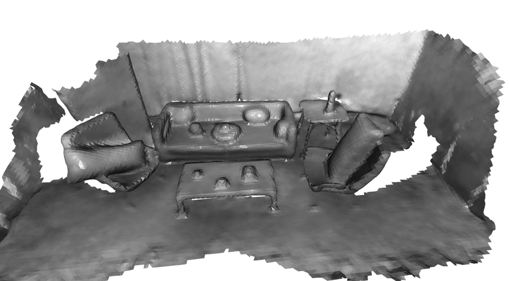}  & \includegraphics[valign=c,height=\sz\linewidth]{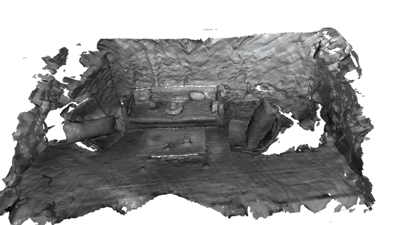}  & \includegraphics[valign=c,height=\sz\linewidth]{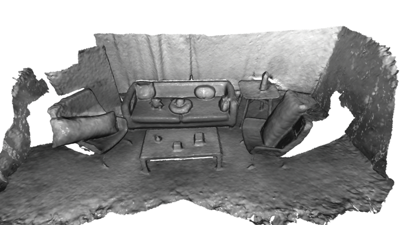}  \\
            \rotatebox[origin=c]{90}{\#Sofa fast1}    & \includegraphics[valign=c,height=\sz\linewidth]{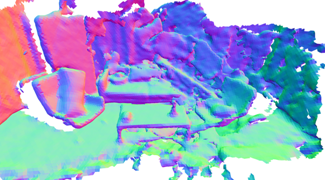}          & \includegraphics[valign=c,height=\sz\linewidth]{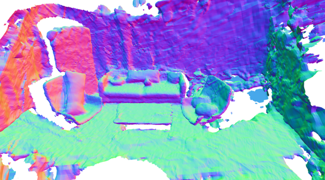}          & \includegraphics[valign=c,height=\sz\linewidth]{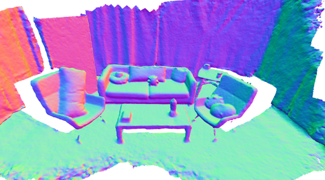}          \\
        \end{tabular}
    }
    \caption{Reconstruction on \textbf{Vector}. All the methods perform comparably in normal subsets, but CoSLAM faces challenges in fast subsets, and ESLAM falls short in precise reconstruction. Our method consistently performs better under both normal and fast movements.}
    \label{fig:add_viz_vector}
\end{figure*}

\boldparagraph{Reconstruction Visualization on Vector.} For the Vector dataset, we show the mesh visualization results in~\cref{fig:add_viz_vector}. All the methods perform comparably in the normal subsets but on the fast subset. All methods show comparable performance on the normal subset. However, in the fast subset, the performance of CoSLAM notably declines, leading to reconstruction ghosting. While ESLAM maintains consistent performance, it falls short in providing detailed reconstruction. Our method achieves consistently excellent performance under both normal and fast camera movements.

\section{Videos Demonstration}
\label{sec:video}
We provide a video of our proposed method \ours along with this document. The video compares \ours with existing state-of-the-art under motion blur and low-lighting environments: \url{./demo.mp4}.

\begin{figure*}[t]
    \vspace{-4ex}
    \centering
    \caption{\textbf{Visualization of the DEV-Indoors dataset}. DEV-Indoors is rendered from Blender models, including 9 subsets containing high-quality color images, depth, meshes, and ground truth trajectories by varying the scene lighting and camera exposure time.}
    {\footnotesize
        \setlength{\tabcolsep}{0.7pt}
        \newcommand{\sz}{0.06}
        \begin{tabular}{|c|ccc|c|c|c}
            \hline
                                                       & \#Room                                                                                            & \#Apartment                                                                                          & \# Workshop                                                                                             & \quad\#Length (frame)~ & \quad\#Duration (second)~ \\
            \hline
            \rotatebox[origin=c]{90}{\#GT Mesh}        & \includegraphics[valign=c,height=\sz\linewidth]{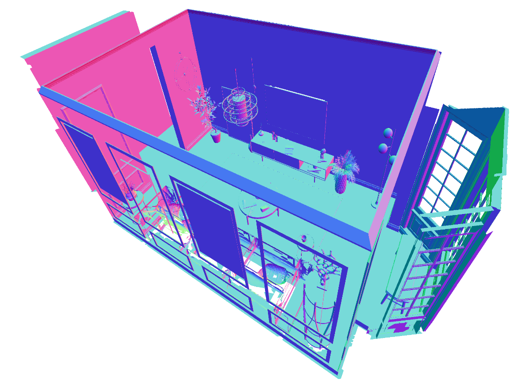}                 & \includegraphics[valign=c,height=\sz\linewidth]{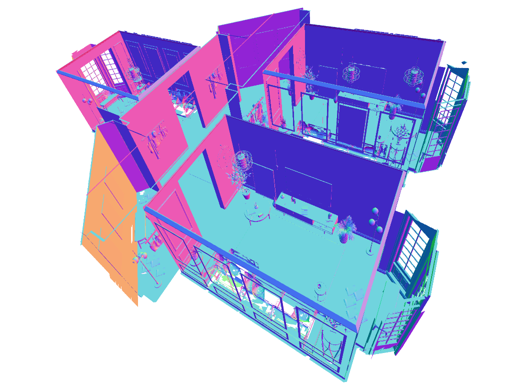}                   & \includegraphics[valign=c,height=\sz\linewidth]{figures/supp/dataset/wkp_mesh.png}                      & ---                    & ---                       \\
            \hline
                                                       & RGB                                                                                               & Event Data                                                                                           & Depth                                                                                                   &                        &                           \\
            \hline
            \rotatebox[origin=c]{90}{\#Room Norm}      & \includegraphics[valign=c,height=\sz\linewidth]{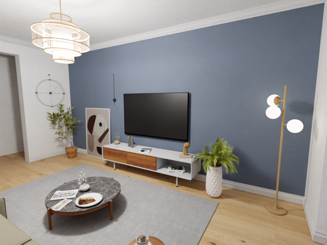}       & \includegraphics[valign=c,height=\sz\linewidth]{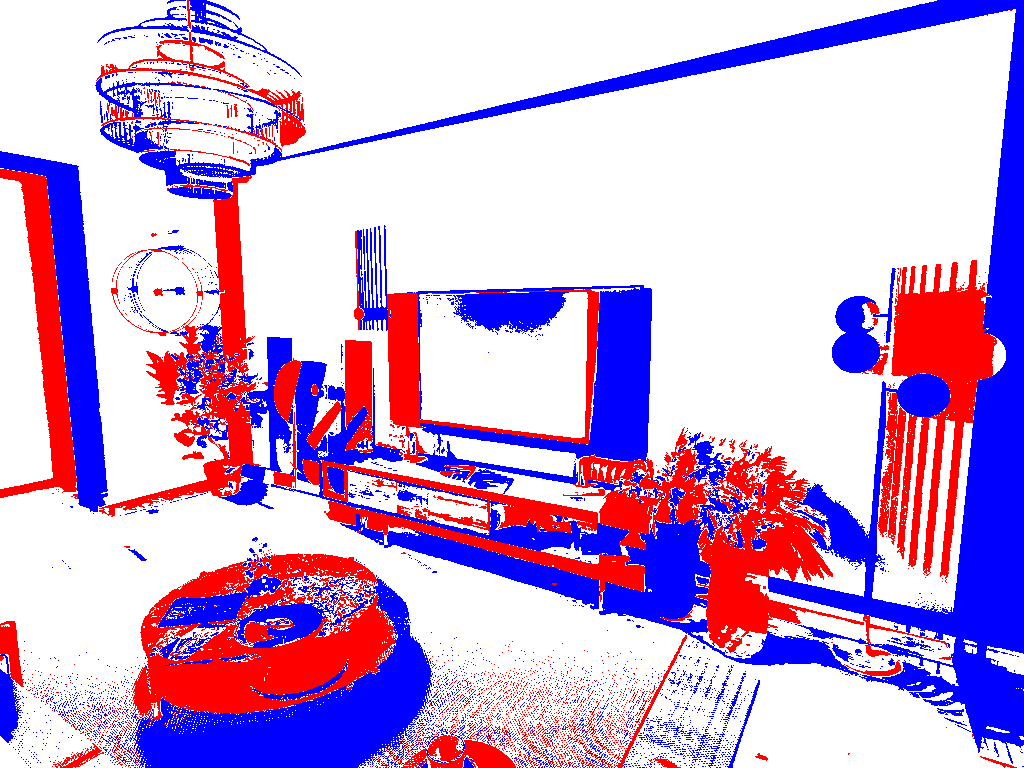}       & \includegraphics[valign=c,height=\sz\linewidth]{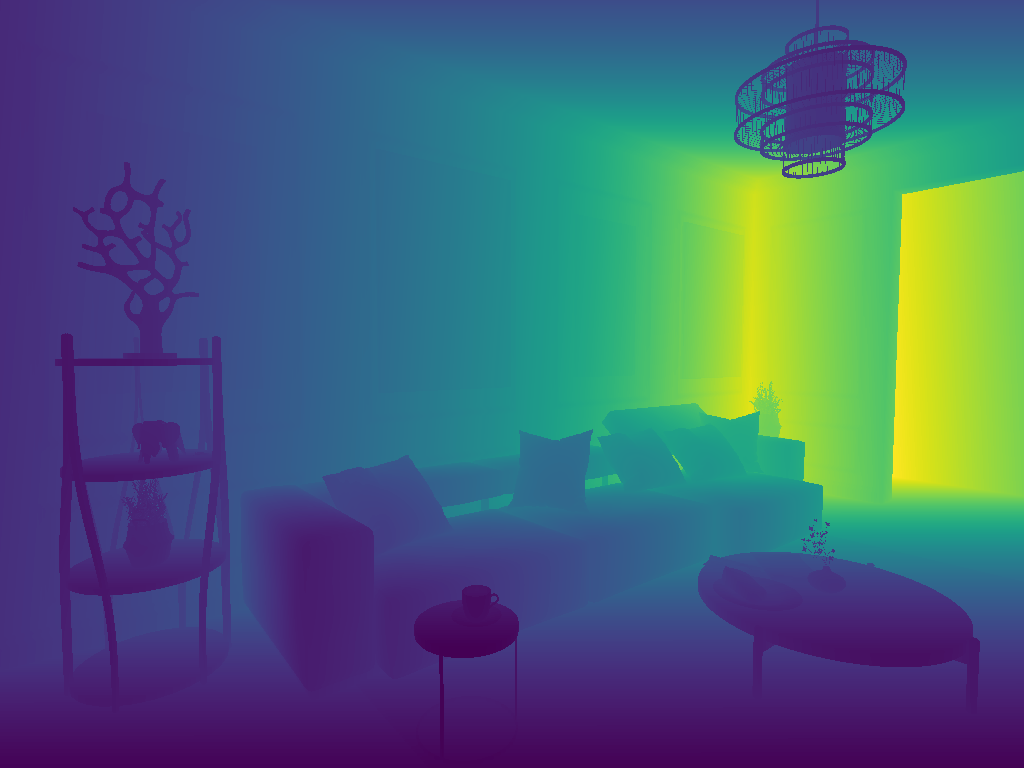}       & 1371                   & 55 s                      \\
            \hline
            \rotatebox[origin=c]{90}{\#Room Blur}      & \includegraphics[valign=c,height=\sz\linewidth]{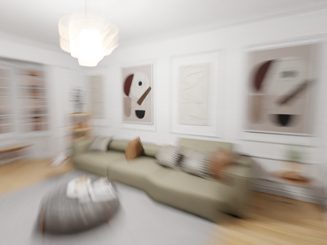}       & \includegraphics[valign=c,height=\sz\linewidth]{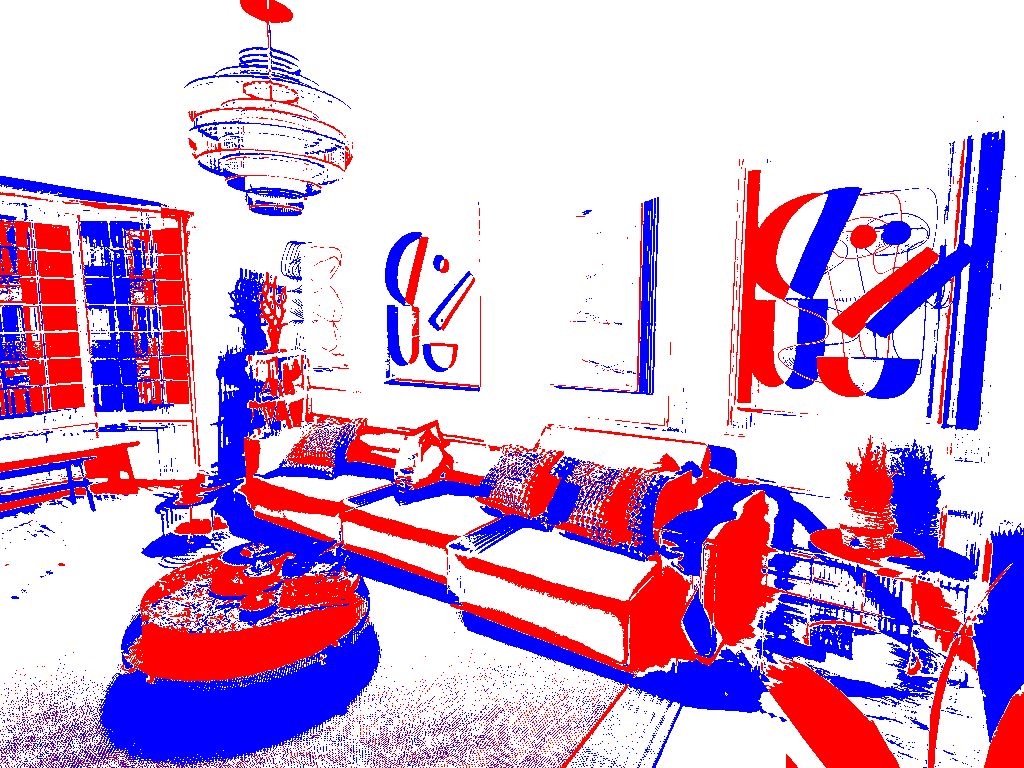}       & \includegraphics[valign=c,height=\sz\linewidth]{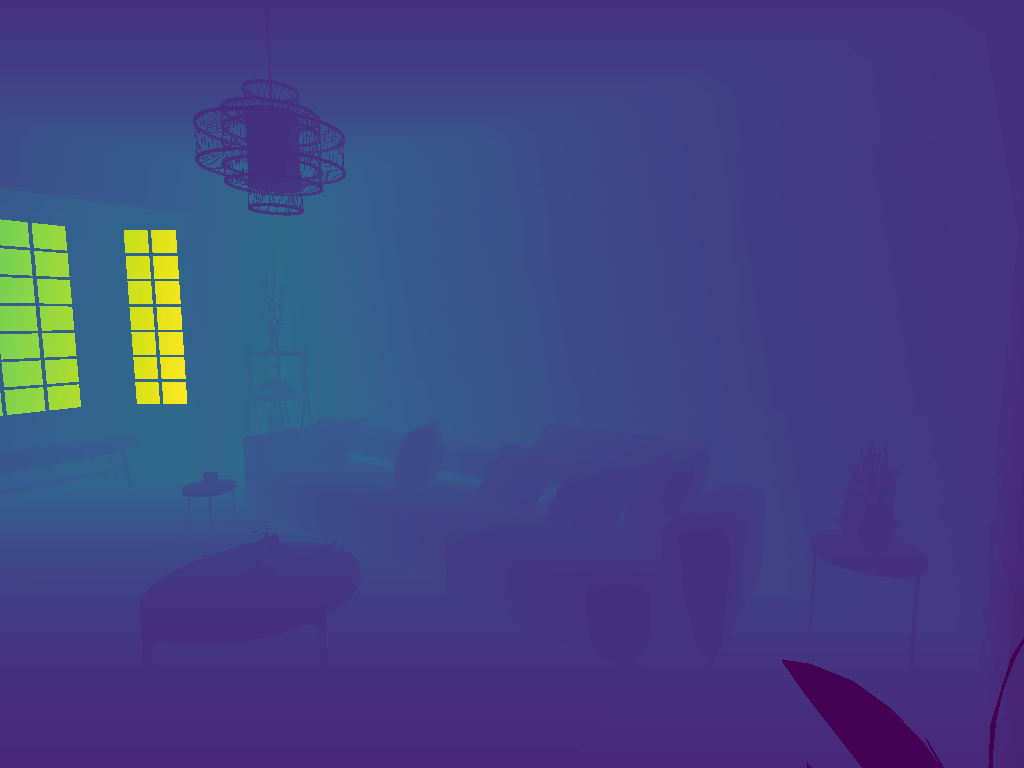}       & 1371                   & 55 s                      \\
            \hline
            \rotatebox[origin=c]{90}{\#Room Dark}      & \includegraphics[valign=c,height=\sz\linewidth]{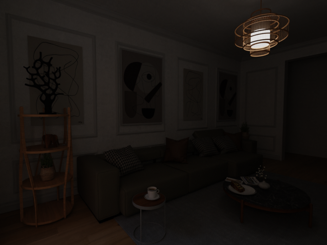}       & \includegraphics[valign=c,height=\sz\linewidth]{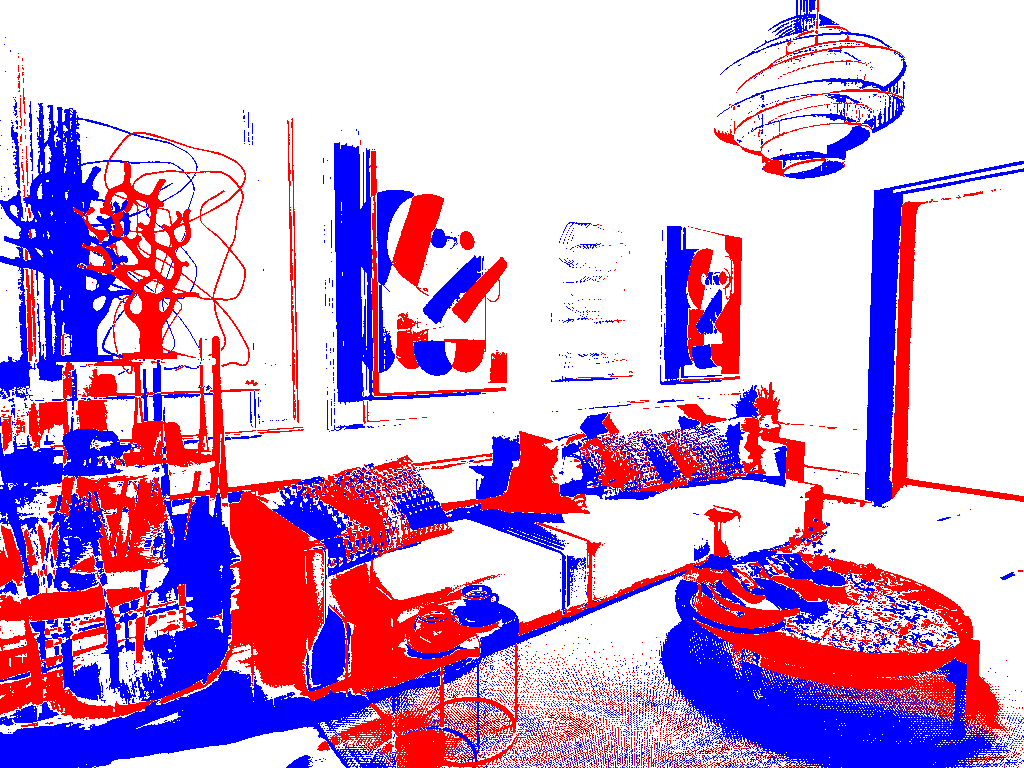}       & \includegraphics[valign=c,height=\sz\linewidth]{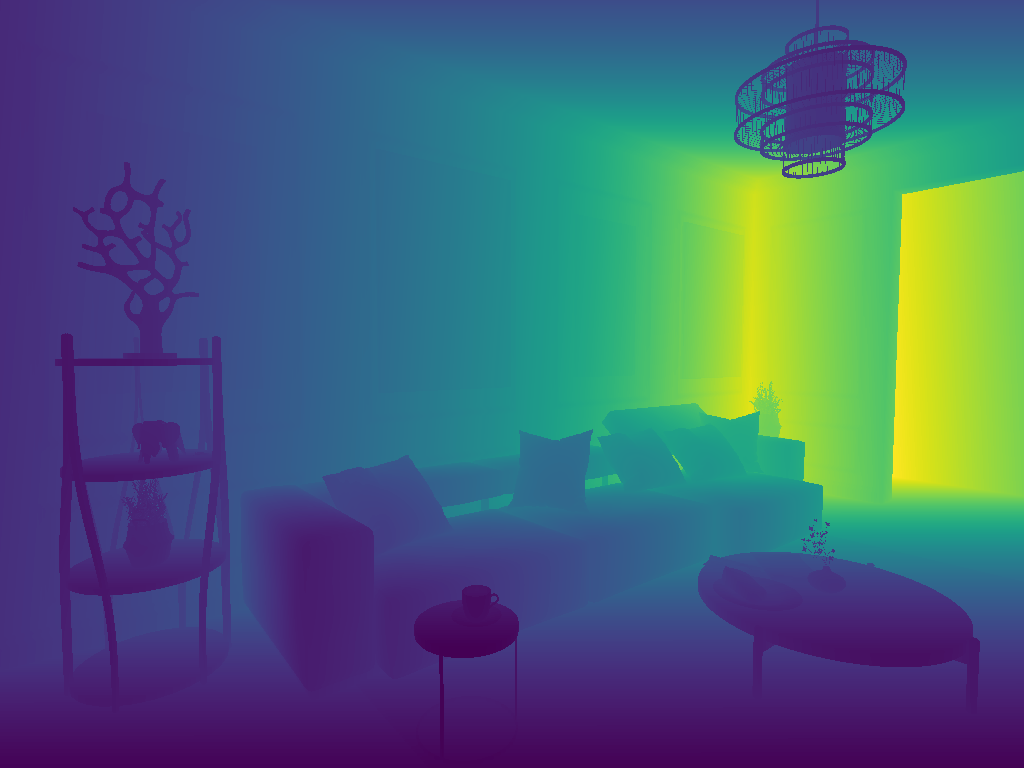}       & 1371                   & 55 s                      \\
            \hline
            \rotatebox[origin=c]{90}{\#Apartment Norm} & \includegraphics[valign=c,height=\sz\linewidth]{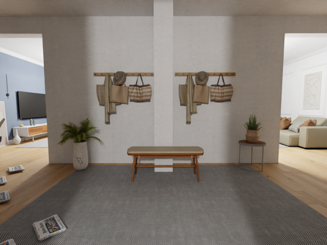} & \includegraphics[valign=c,height=\sz\linewidth]{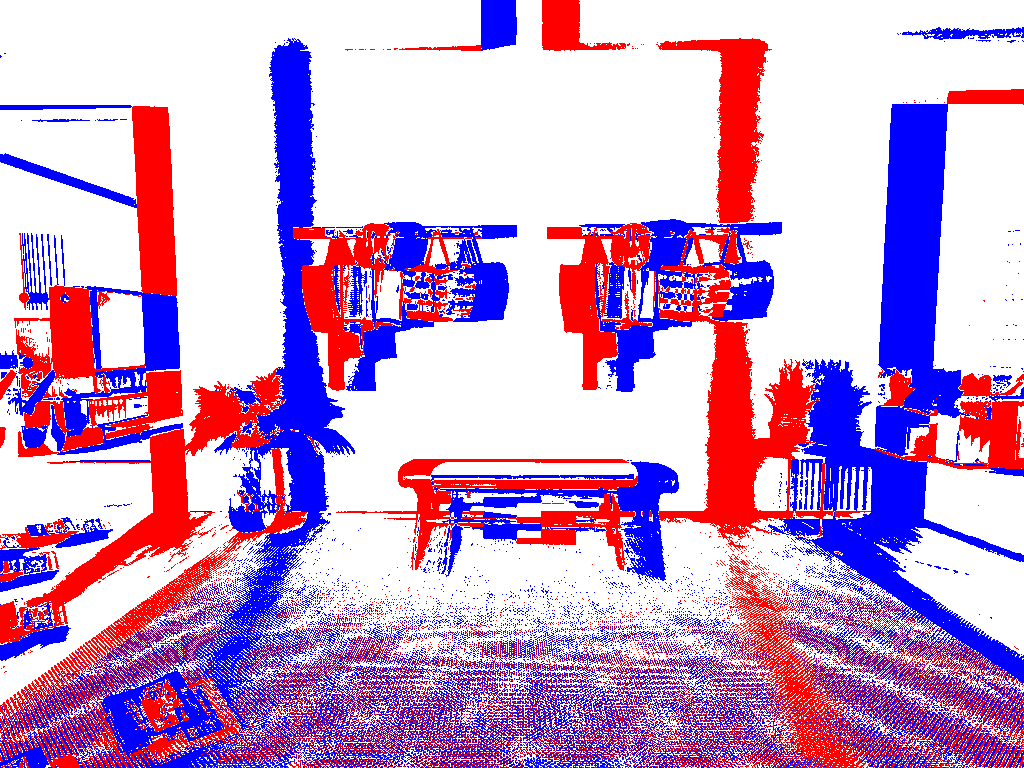} & \includegraphics[valign=c,height=\sz\linewidth]{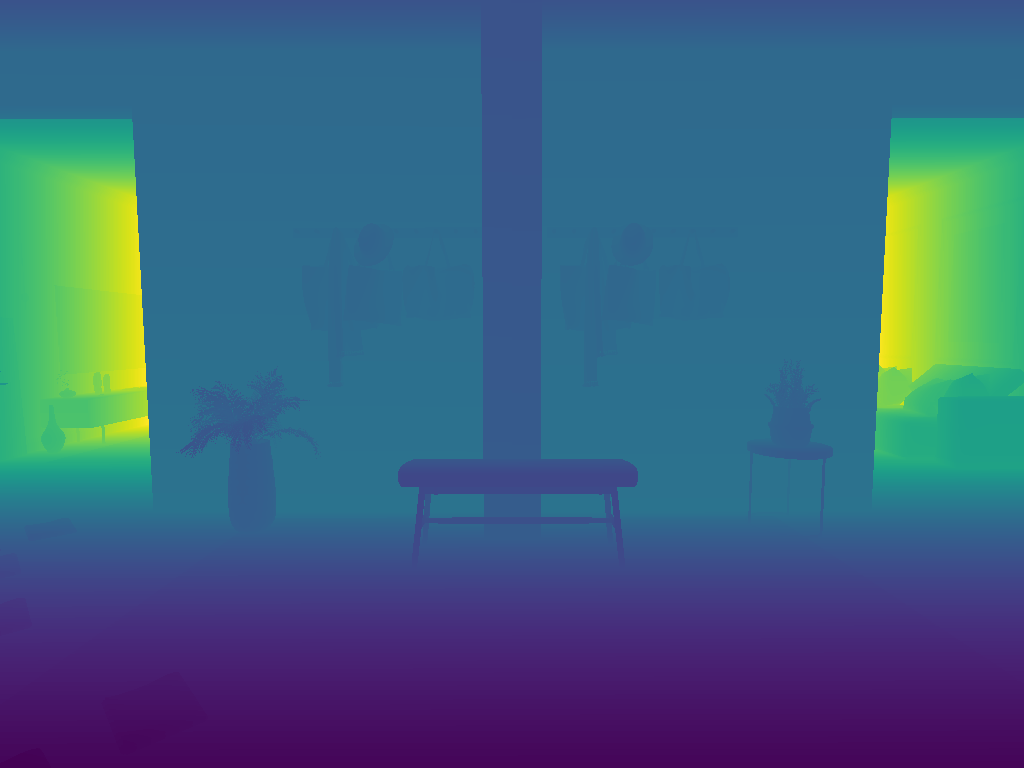} & 3000                   & 120 s                     \\
            \hline
            \rotatebox[origin=c]{90}{\#Apartment Blur} & \includegraphics[valign=c,height=\sz\linewidth]{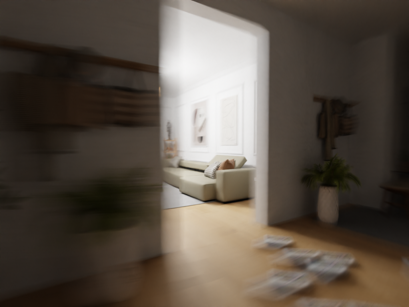} & \includegraphics[valign=c,height=\sz\linewidth]{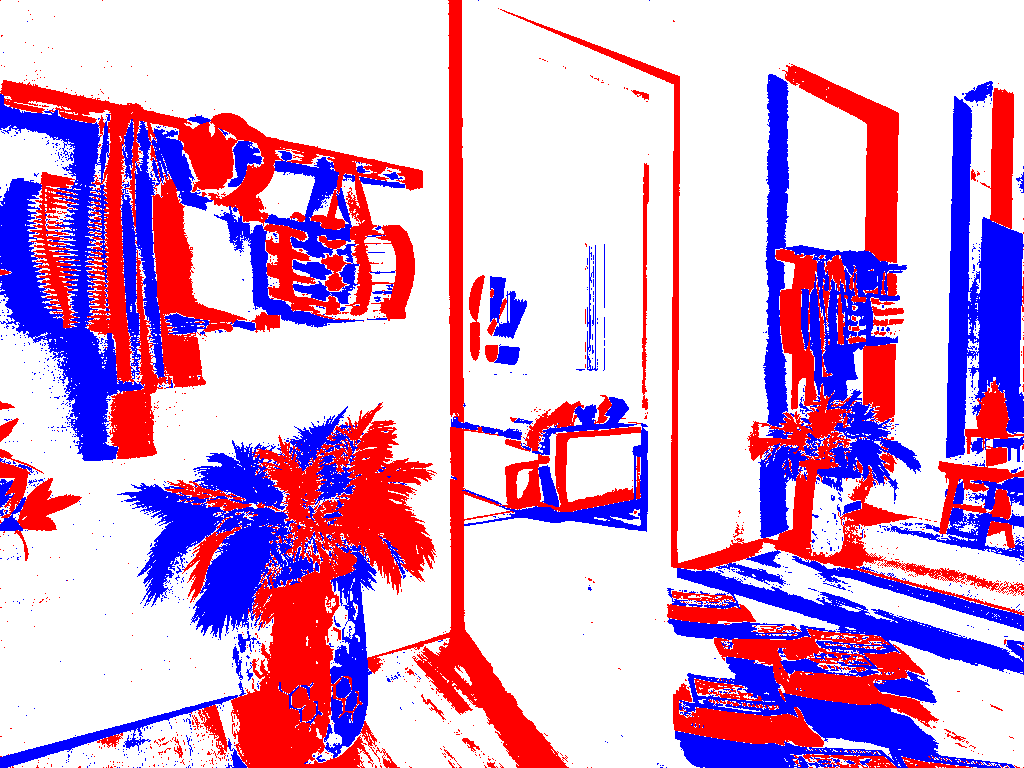} & \includegraphics[valign=c,height=\sz\linewidth]{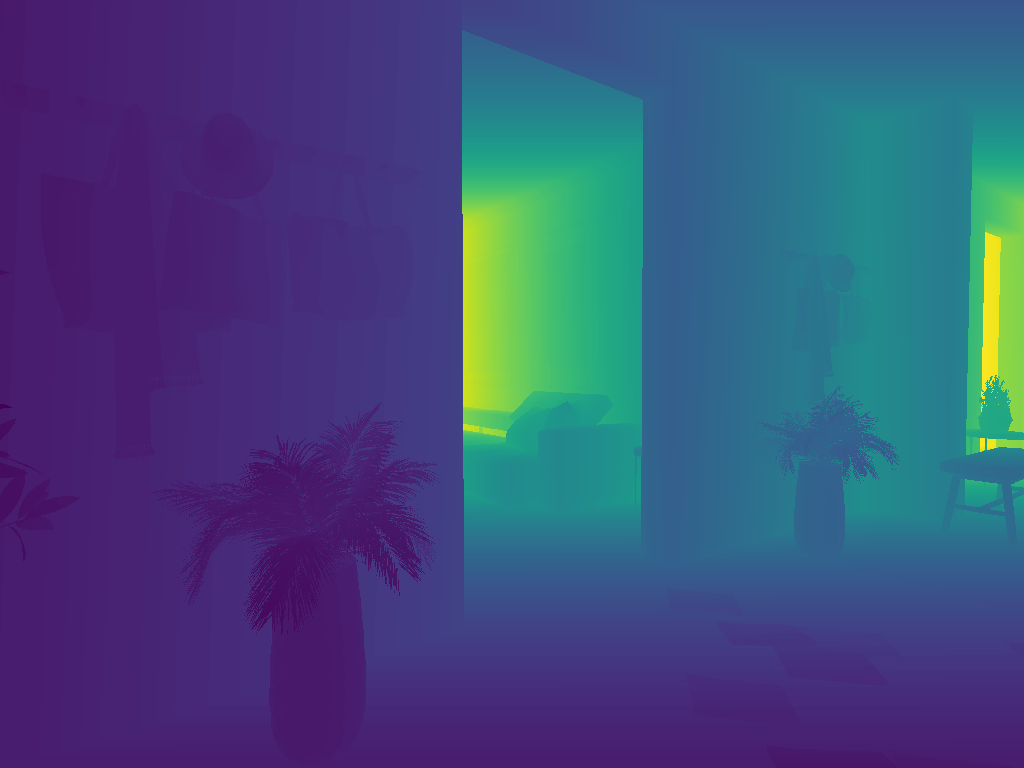} & 3000                   & 120 s                     \\
            \hline
            \rotatebox[origin=c]{90}{\#Apartment Dark} & \includegraphics[valign=c,height=\sz\linewidth]{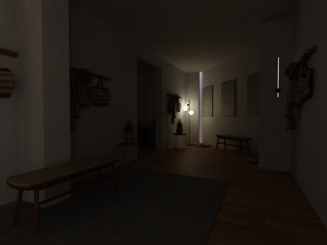} & \includegraphics[valign=c,height=\sz\linewidth]{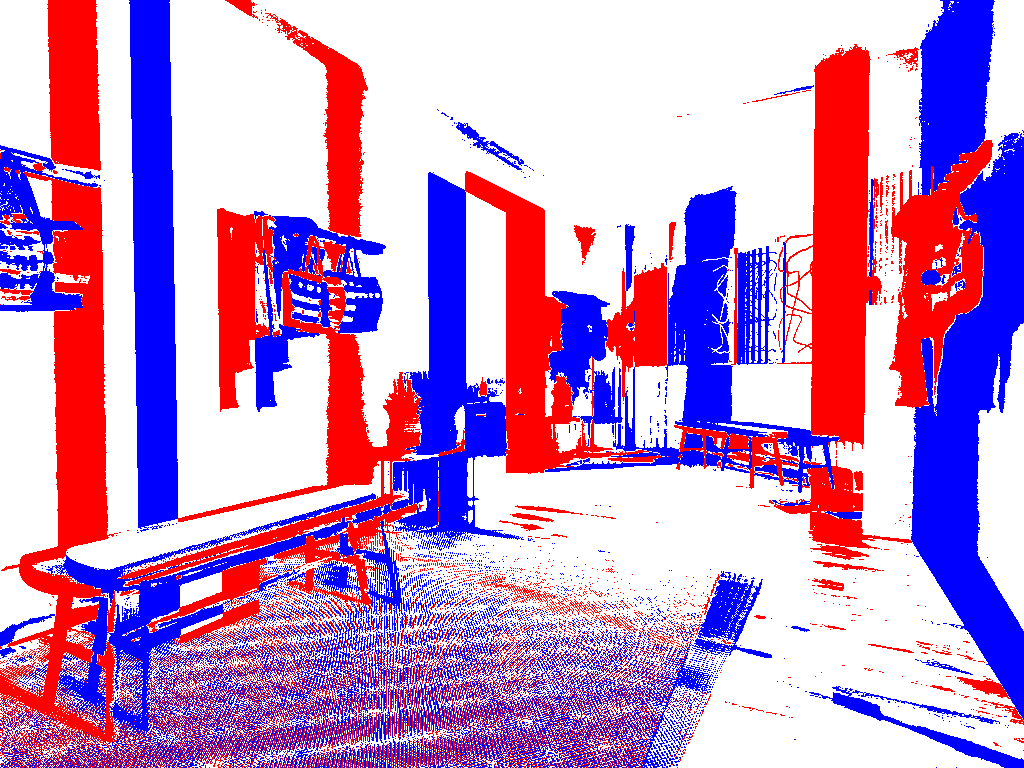} & \includegraphics[valign=c,height=\sz\linewidth]{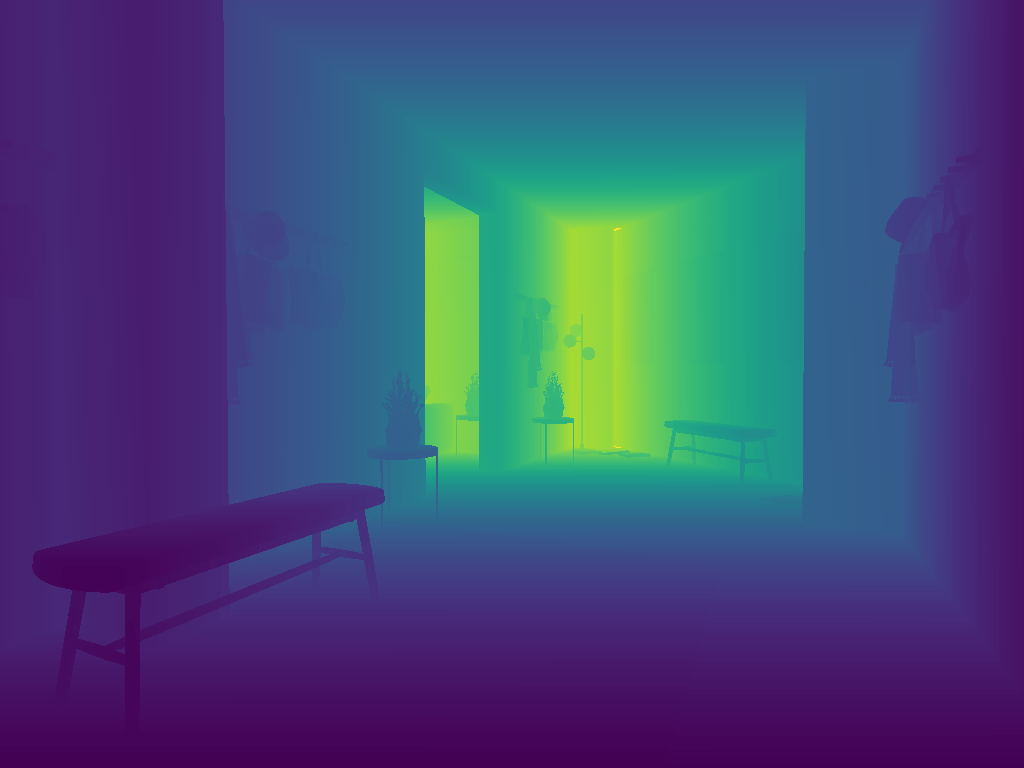} & 3000                   & 120 s                     \\
            \hline
            \rotatebox[origin=c]{90}{\#Workshop Norm}  & \includegraphics[valign=c,height=\sz\linewidth]{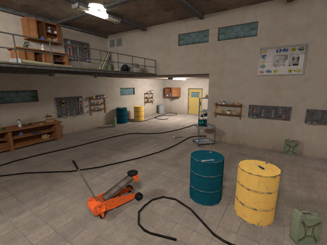}  & \includegraphics[valign=c,height=\sz\linewidth]{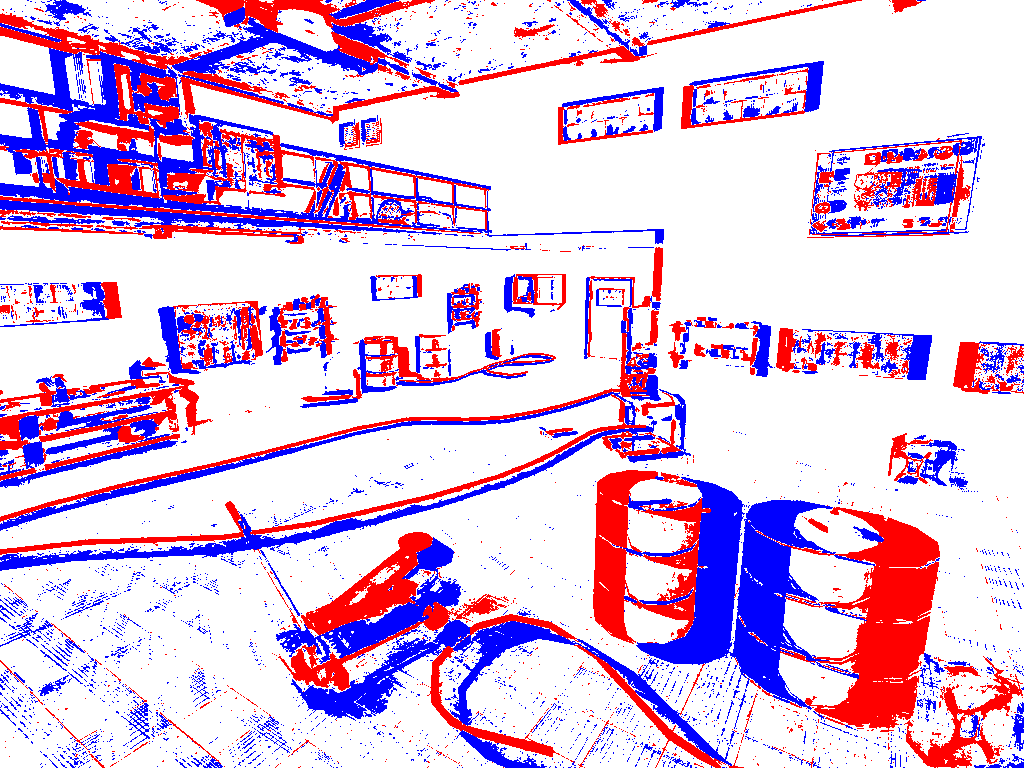}  & \includegraphics[valign=c,height=\sz\linewidth]{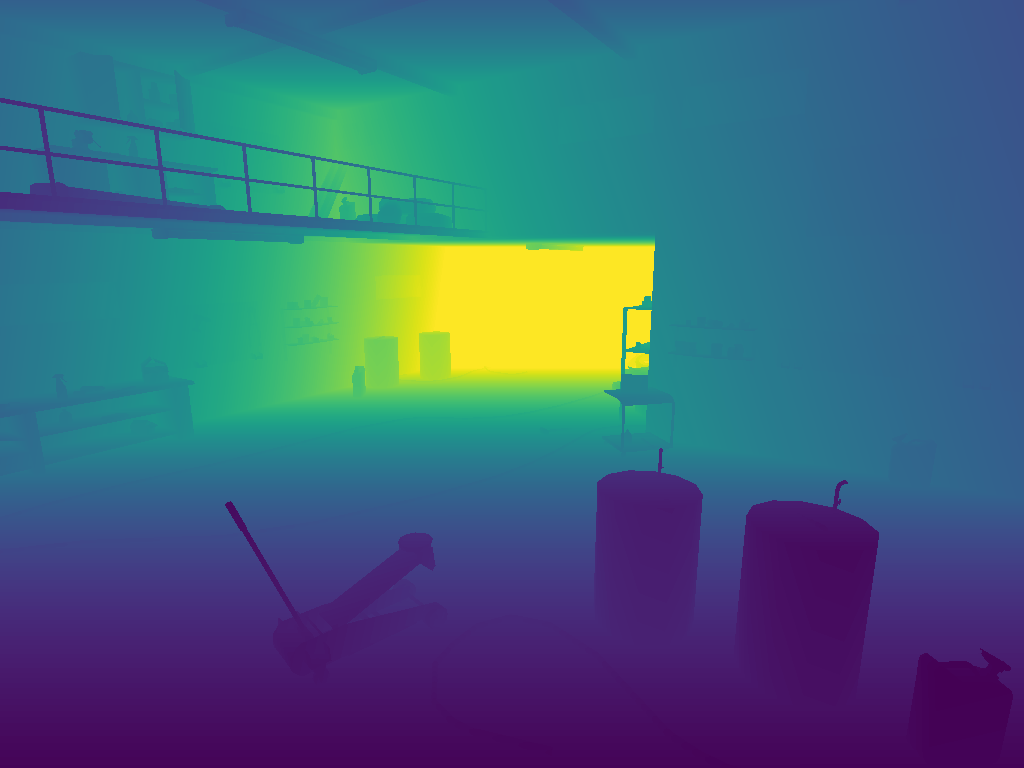}  & 1800                   & 72 s                      \\
            \hline
            \rotatebox[origin=c]{90}{\#Workshop Blur}  & \includegraphics[valign=c,height=\sz\linewidth]{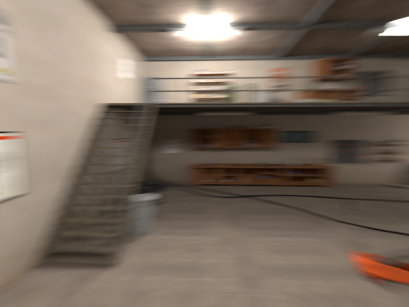}  & \includegraphics[valign=c,height=\sz\linewidth]{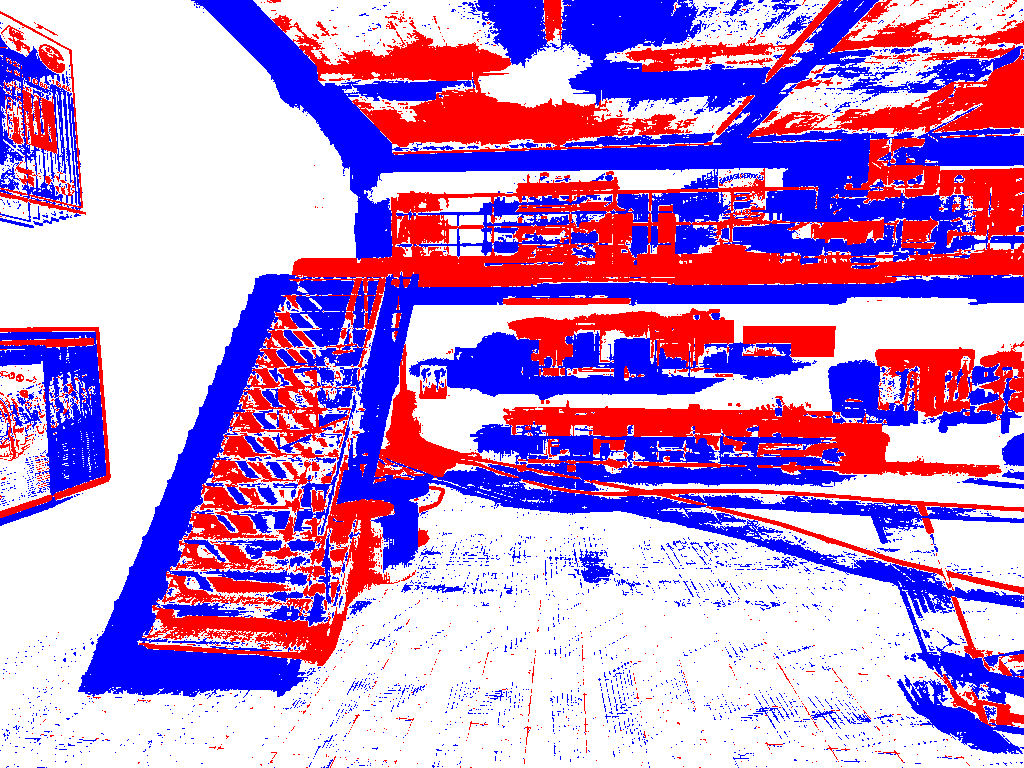}  & \includegraphics[valign=c,height=\sz\linewidth]{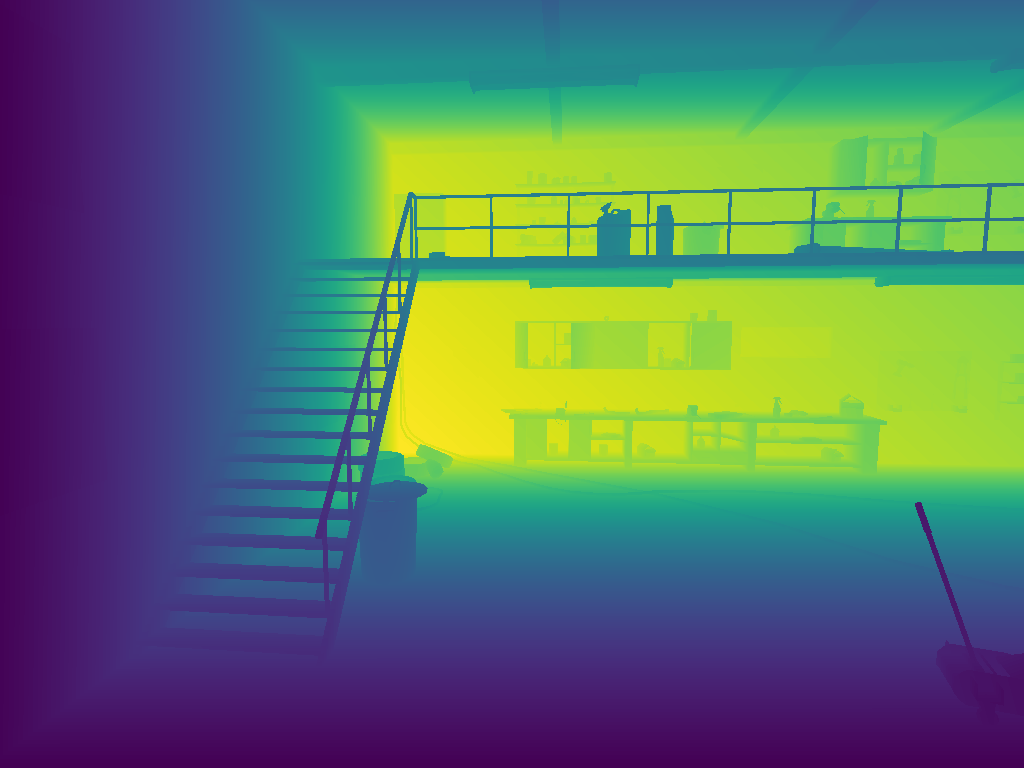}  & 1800                   & 72 s                      \\
            \hline
            \rotatebox[origin=c]{90}{\#Workshop Dark}  & \includegraphics[valign=c,height=\sz\linewidth]{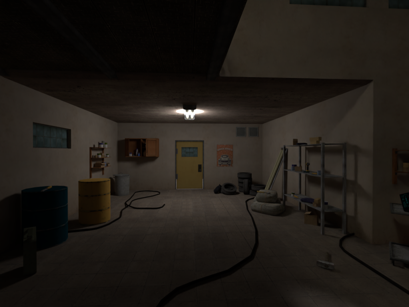}  & \includegraphics[valign=c,height=\sz\linewidth]{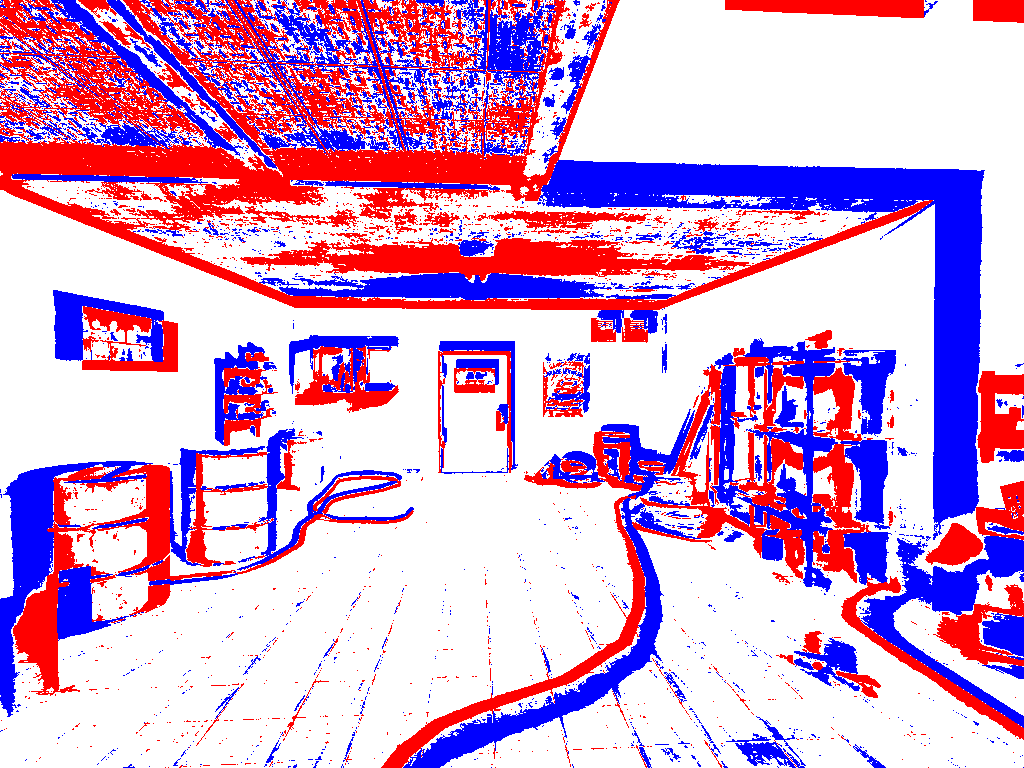}  & \includegraphics[valign=c,height=\sz\linewidth]{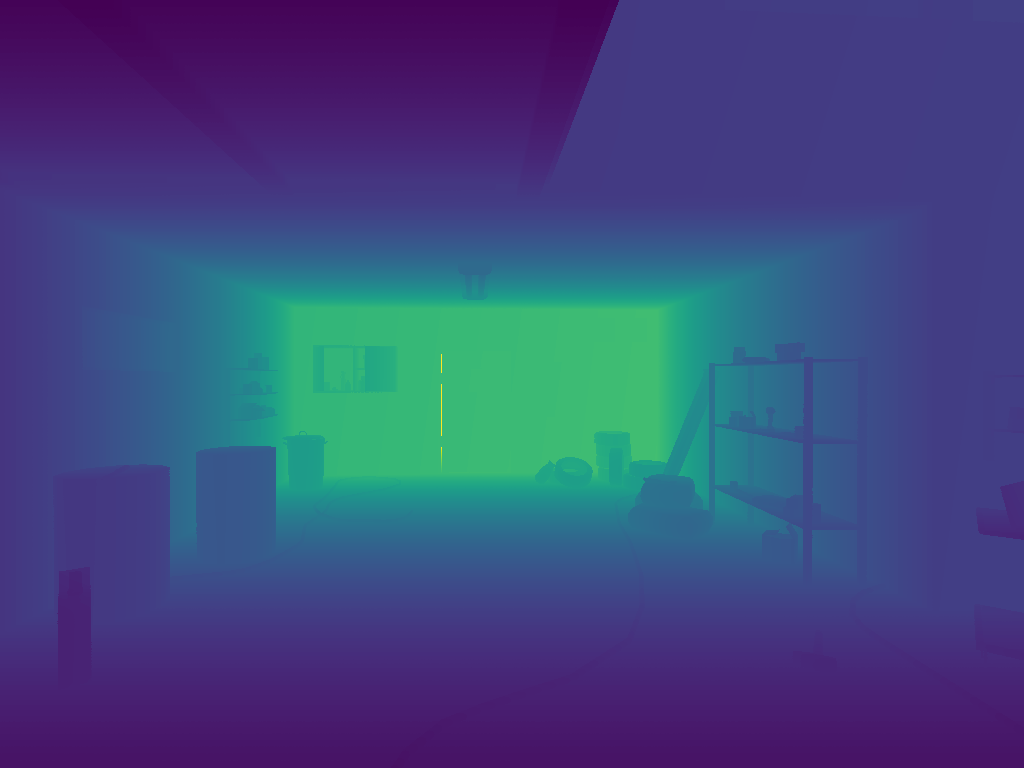}  & 1800                   & 72 s                      \\
            \hline
        \end{tabular}
    }
    \label{fig:devindoors_overview}
\end{figure*}

\begin{figure*}[t]
    \vspace{-4ex}
    \centering
    \caption{\textbf{Visualization of the DEV-Reals dataset.} DEV-Reals is captured from real scenes: \#Pioffice, \#Garage, and \#Dormitory, providing 8 challenging subsets containing color images, depth, and ground truth trajectories under motion blur and lighting variation.}
    {\footnotesize
        \setlength{\tabcolsep}{0.7pt}
        \newcommand{\sz}{0.08}
        \begin{tabular}{|ccccc|c|c|}
            \hline
                                                   & RGB                                                                                  & Event Data                                                                          & Depth                                                                                       & Trajectory                                                                           & \quad\#Length~ & \quad\#Duration~ \\\hline
            \rotatebox[origin=c]{90}{\#Pioffice1}  & \includegraphics[valign=c,height=\sz\linewidth]{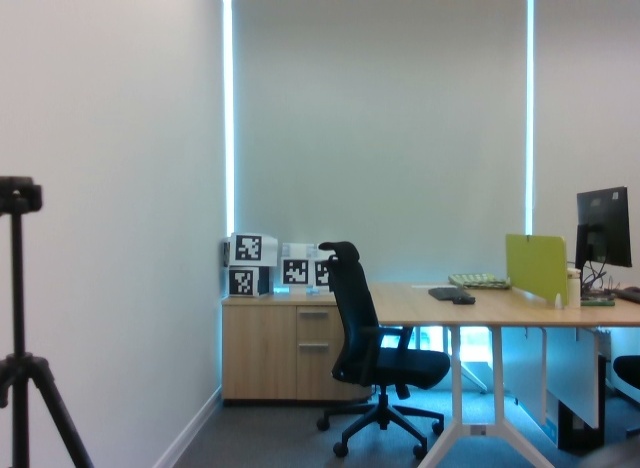} & \includegraphics[valign=c,height=\sz\linewidth]{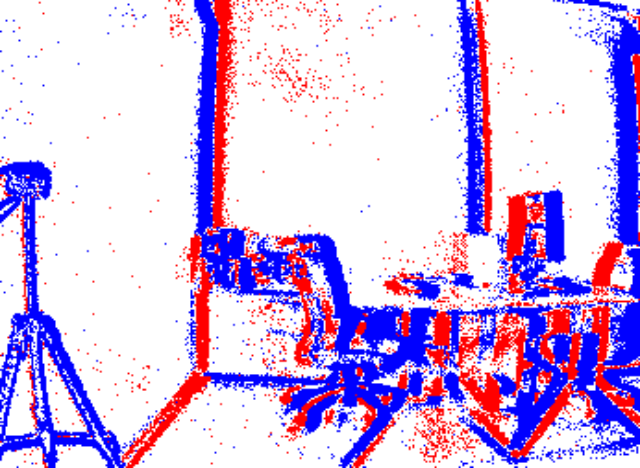} & \includegraphics[valign=c,height=\sz\linewidth]{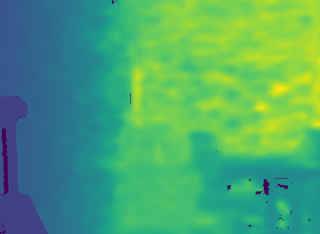} & \includegraphics[valign=c,height=\sz\linewidth]{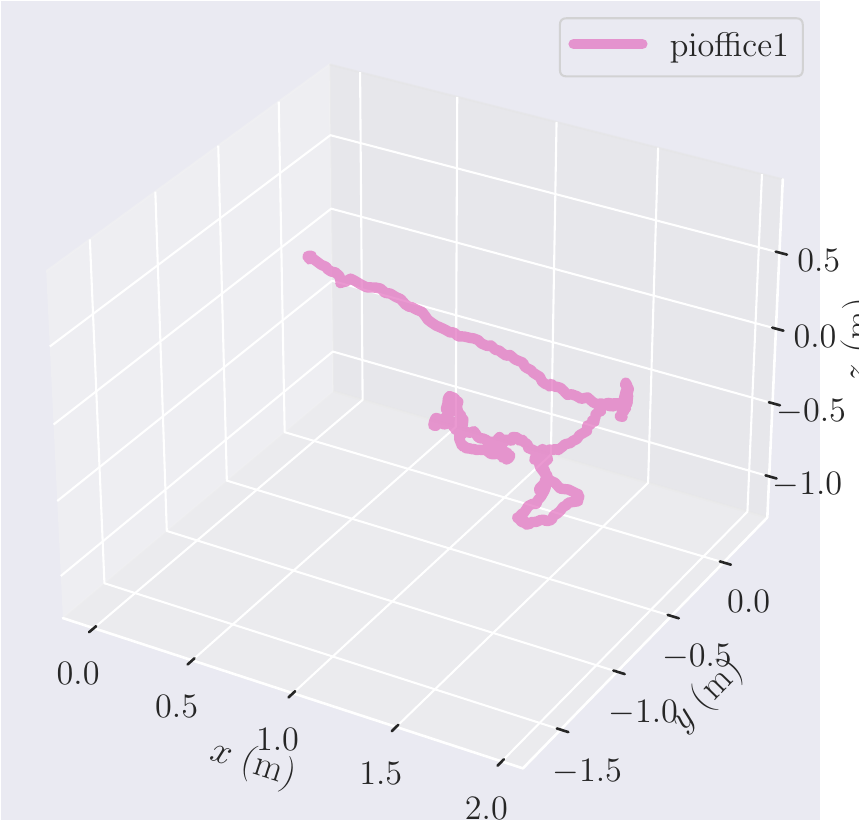}  & 1209           & 80.6 s           \\\hline
            \rotatebox[origin=c]{90}{\#Pioffice2}  & \includegraphics[valign=c,height=\sz\linewidth]{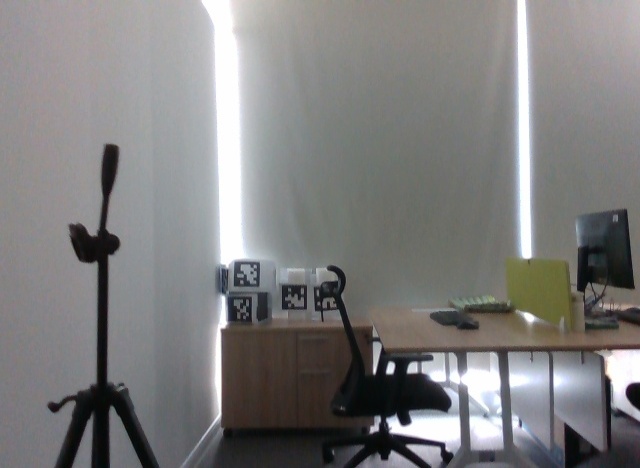} & \includegraphics[valign=c,height=\sz\linewidth]{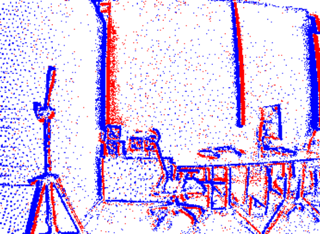} & \includegraphics[valign=c,height=\sz\linewidth]{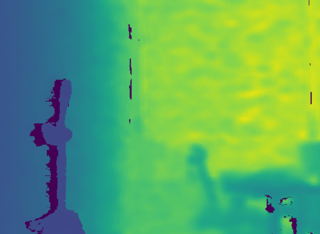} & \includegraphics[valign=c,height=\sz\linewidth]{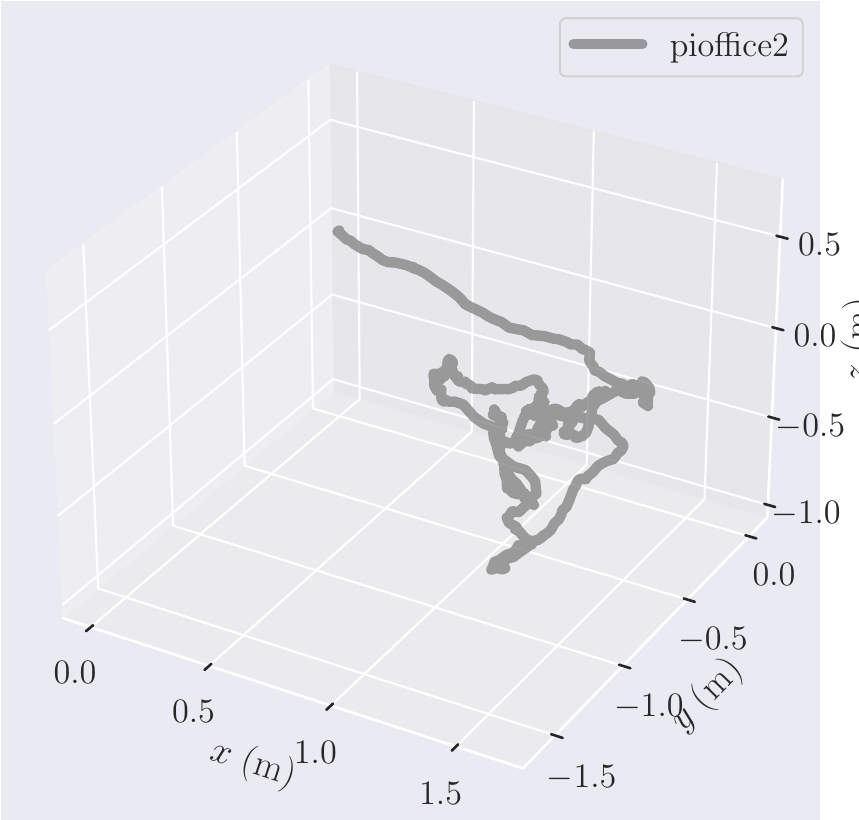}  & 1286           & 85.7 s           \\\hline
            \rotatebox[origin=c]{90}{\#Garage1}    & \includegraphics[valign=c,height=\sz\linewidth]{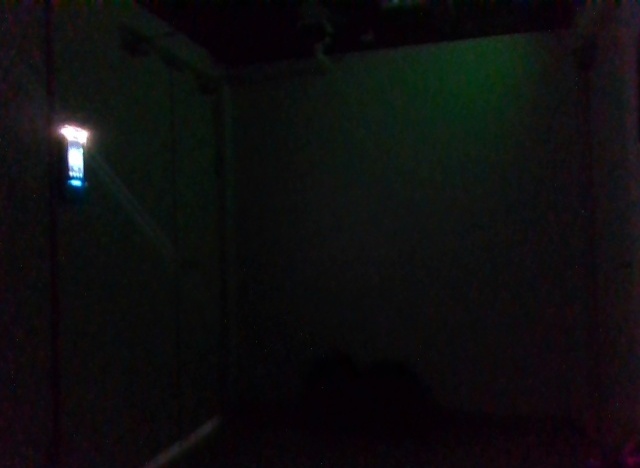} & \includegraphics[valign=c,height=\sz\linewidth]{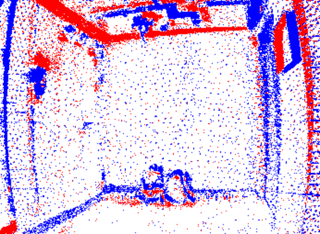} & \includegraphics[valign=c,height=\sz\linewidth]{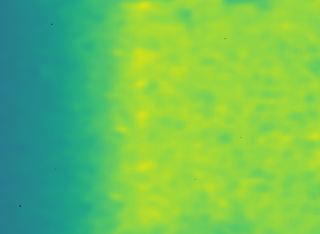} & \includegraphics[valign=c,height=\sz\linewidth]{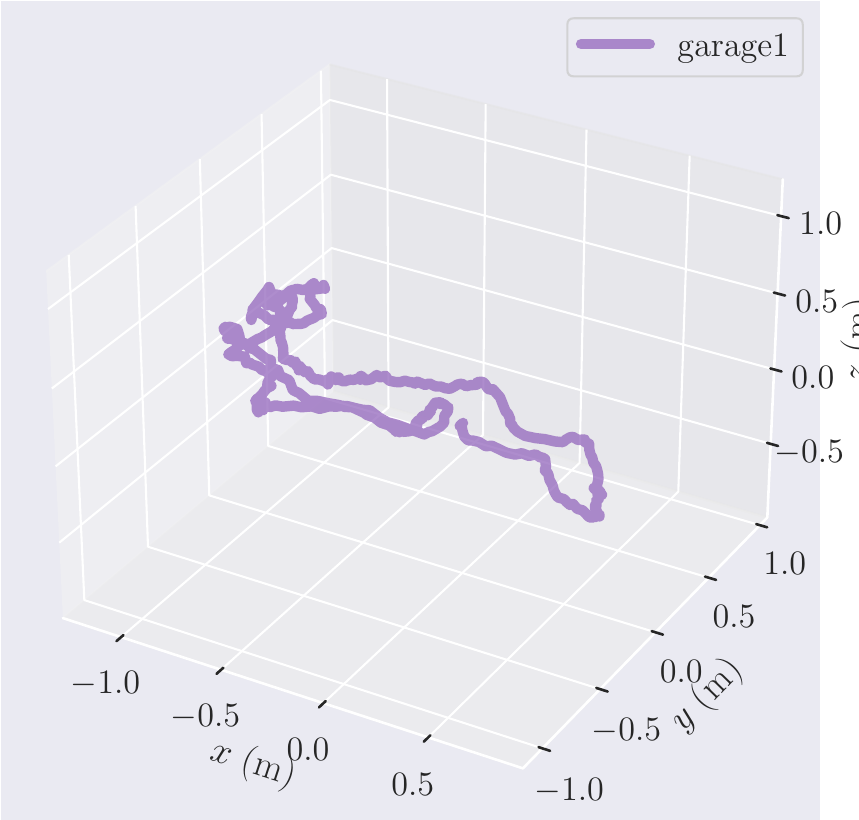}    & 1384           & 92.7 s           \\\hline
            \rotatebox[origin=c]{90}{\#Garage2}    & \includegraphics[valign=c,height=\sz\linewidth]{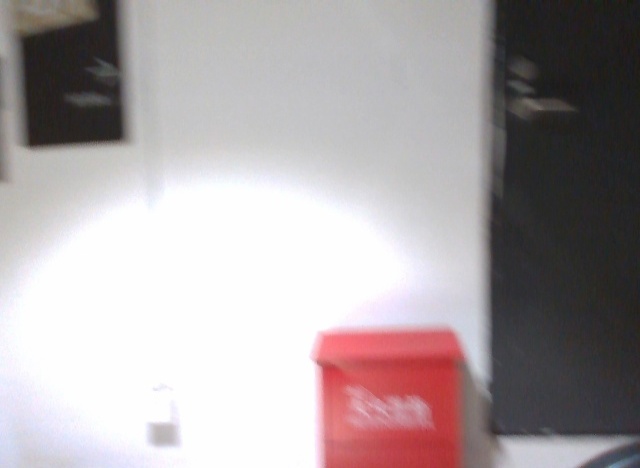} & \includegraphics[valign=c,height=\sz\linewidth]{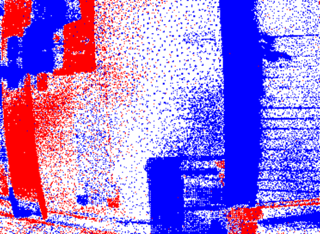} & \includegraphics[valign=c,height=\sz\linewidth]{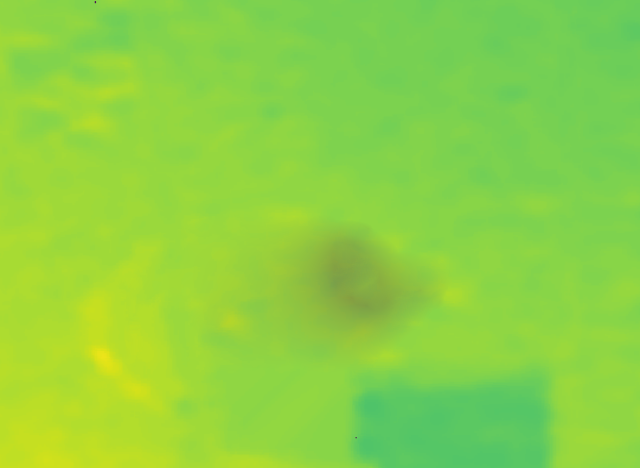} & \includegraphics[valign=c,height=\sz\linewidth]{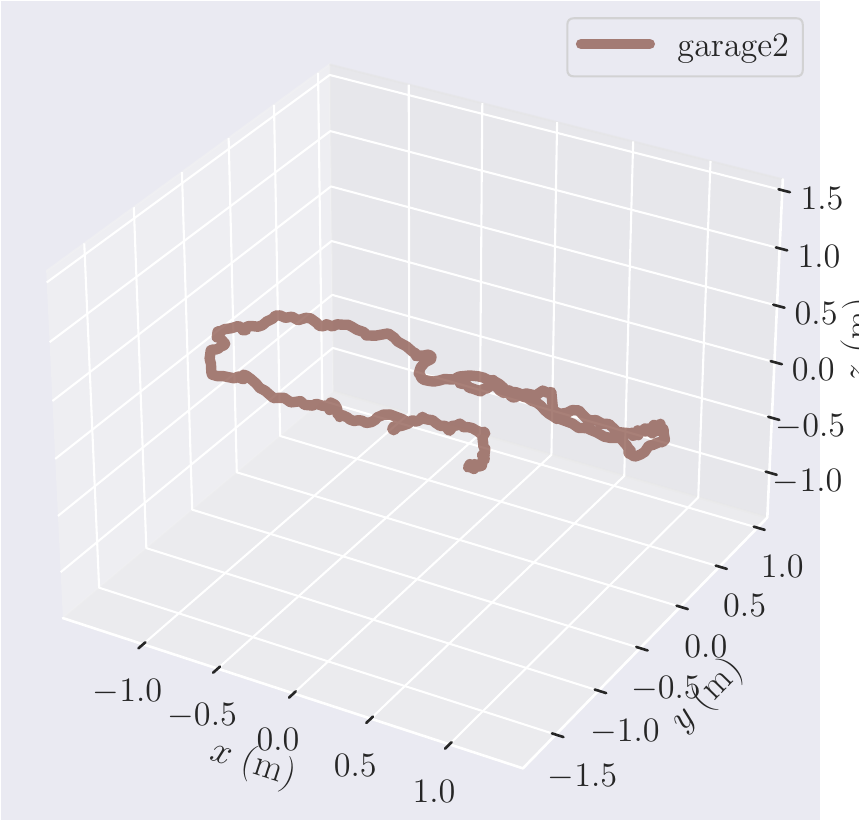}    & 989            & 65.9 s           \\\hline
            \rotatebox[origin=c]{90}{\#Dormitory1} & \includegraphics[valign=c,height=\sz\linewidth]{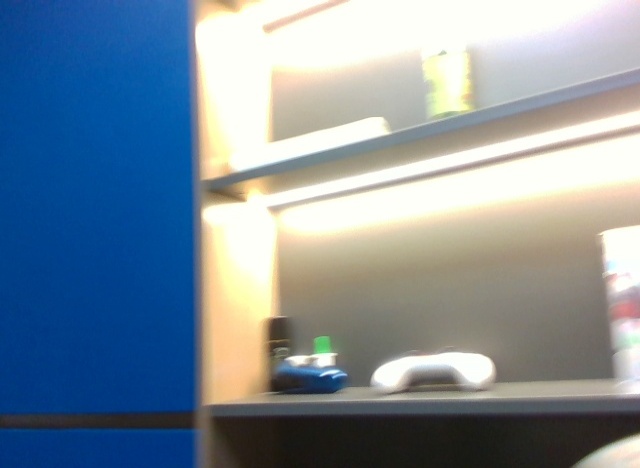} & \includegraphics[valign=c,height=\sz\linewidth]{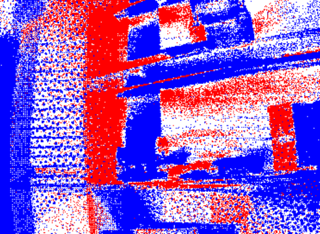} & \includegraphics[valign=c,height=\sz\linewidth]{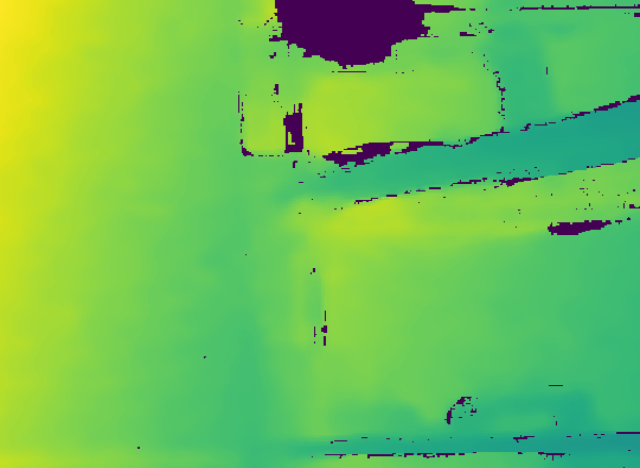} & \includegraphics[valign=c,height=\sz\linewidth]{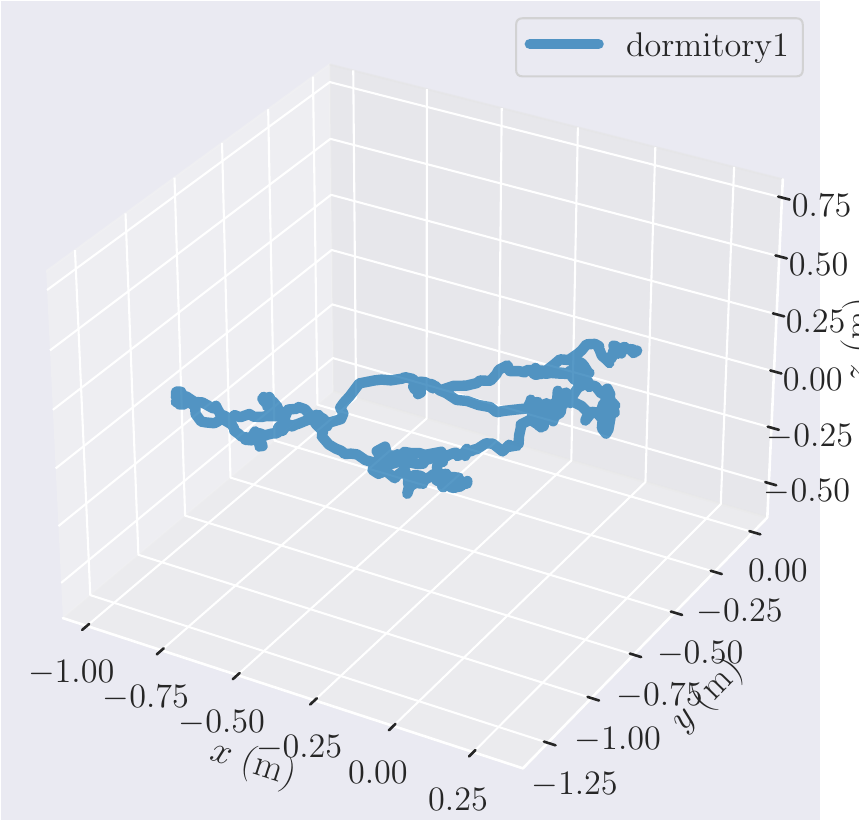} & 1799           & 119.9 s          \\\hline
            \rotatebox[origin=c]{90}{\#Dormitory2} & \includegraphics[valign=c,height=\sz\linewidth]{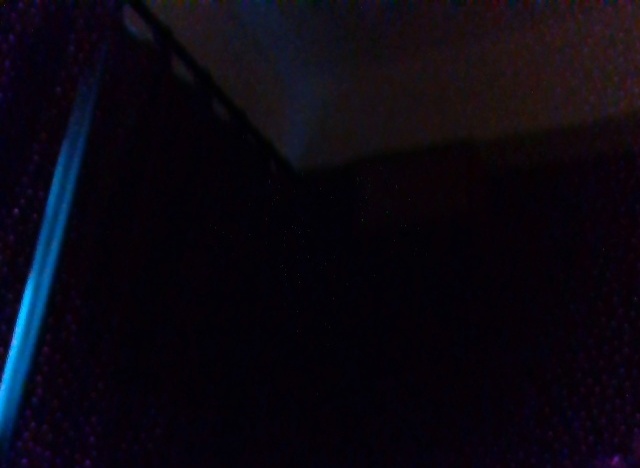} & \includegraphics[valign=c,height=\sz\linewidth]{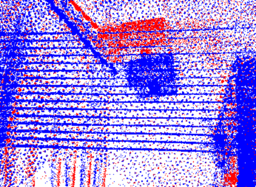} & \includegraphics[valign=c,height=\sz\linewidth]{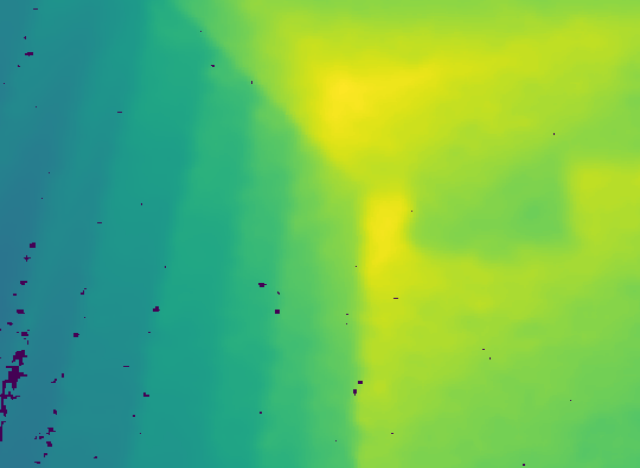} & \includegraphics[valign=c,height=\sz\linewidth]{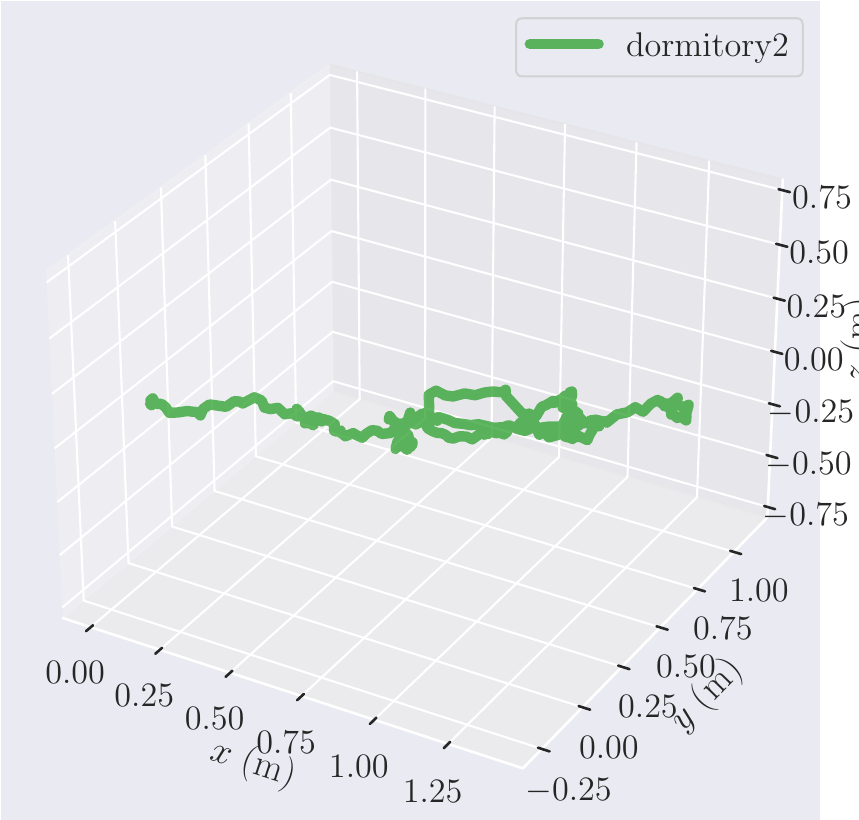} & 961            & 64.1 s           \\\hline
            \rotatebox[origin=c]{90}{\#Dormitory3} & \includegraphics[valign=c,height=\sz\linewidth]{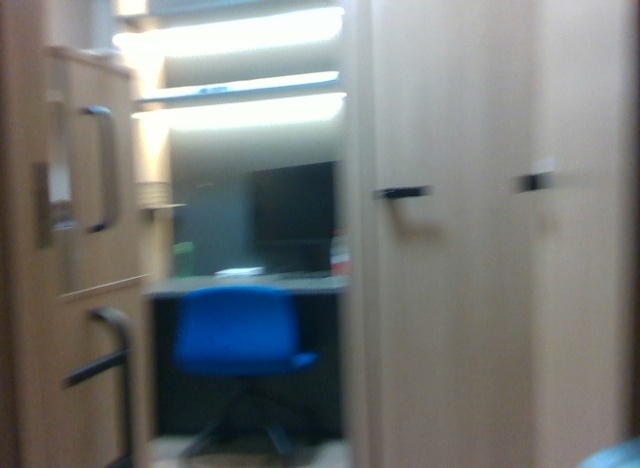} & \includegraphics[valign=c,height=\sz\linewidth]{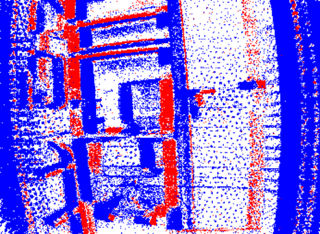} & \includegraphics[valign=c,height=\sz\linewidth]{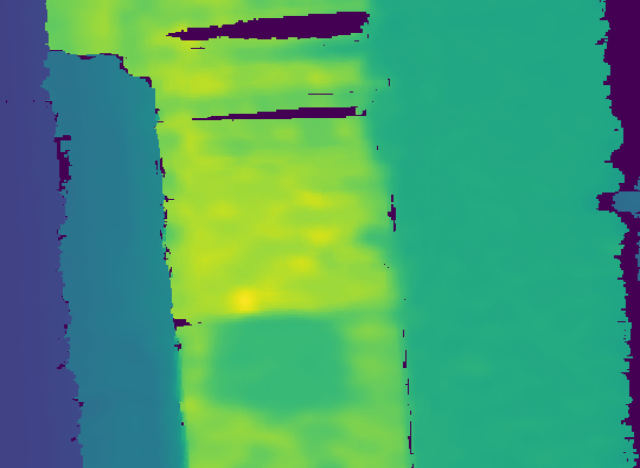} & \includegraphics[valign=c,height=\sz\linewidth]{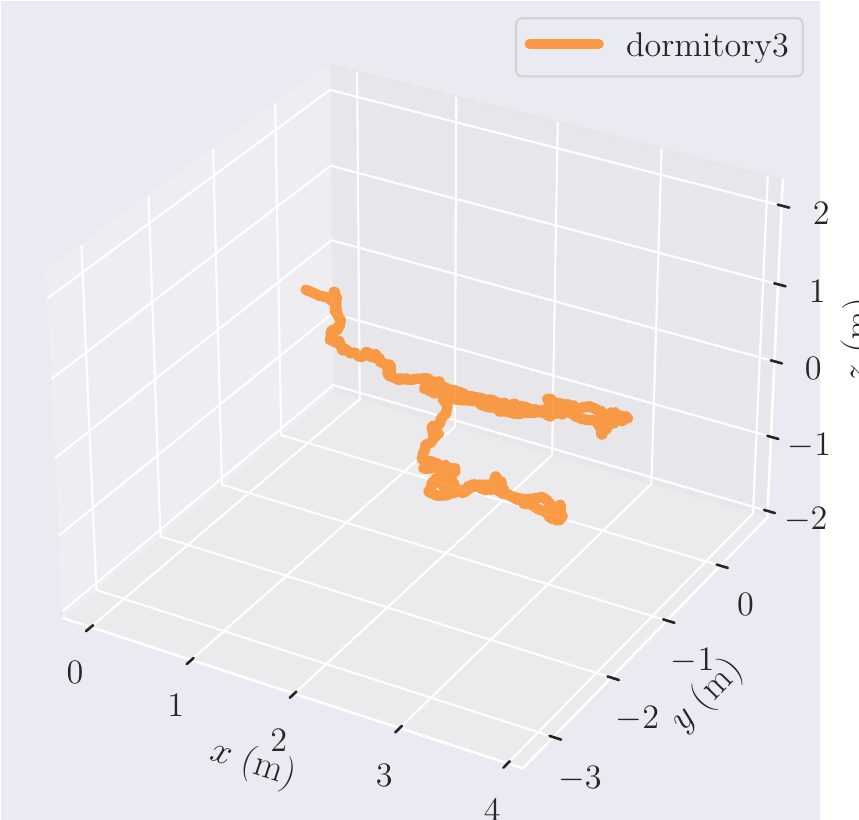} & 2726           & 181.7 s          \\\hline
            \rotatebox[origin=c]{90}{\#Dormitory4} & \includegraphics[valign=c,height=\sz\linewidth]{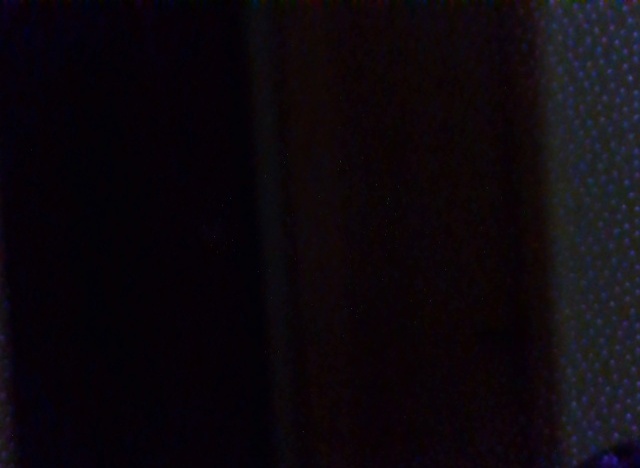} & \includegraphics[valign=c,height=\sz\linewidth]{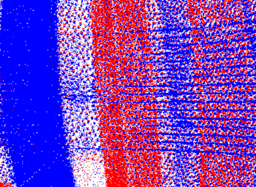} & \includegraphics[valign=c,height=\sz\linewidth]{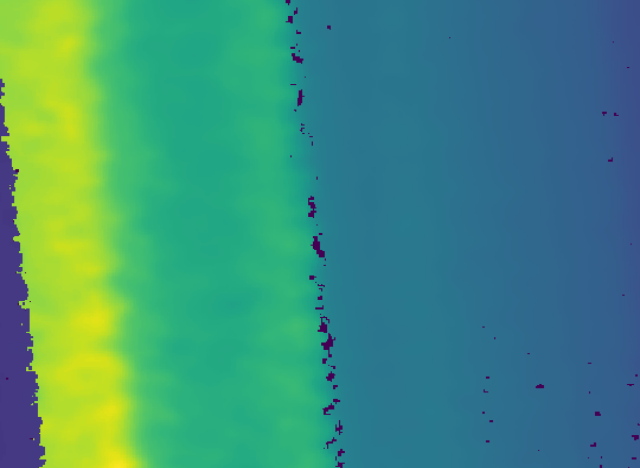} & \includegraphics[valign=c,height=\sz\linewidth]{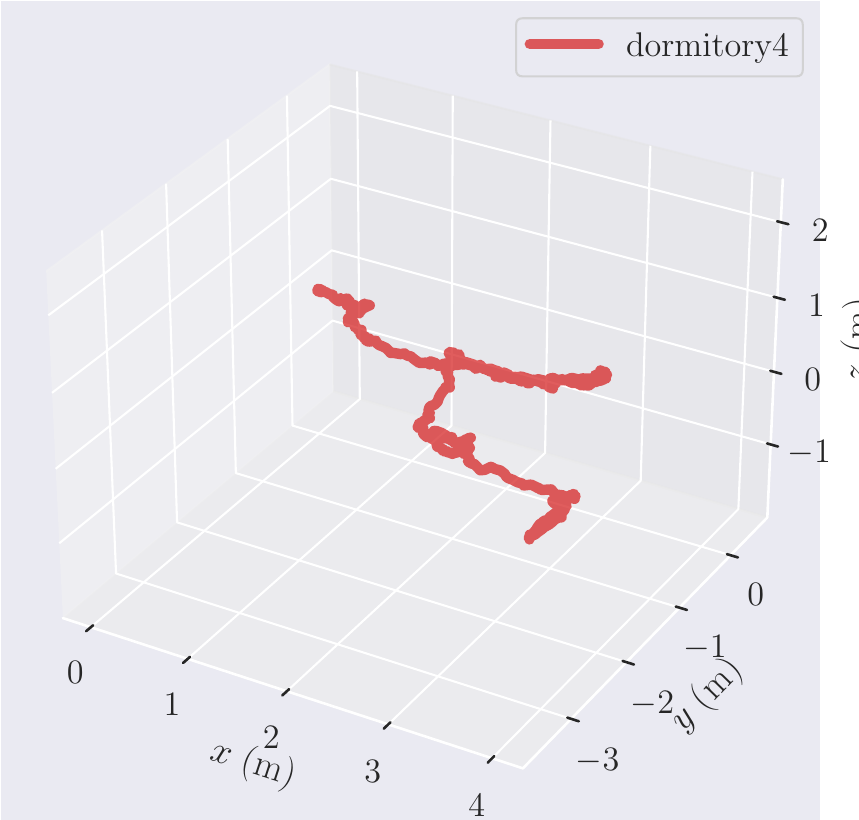} & 1928           & 128.5 s          \\\hline
        \end{tabular}
    }
    \label{fig:devreals_overview}
\end{figure*}
\clearpage \clearpage
{
    \small
    \bibliographystyle{ieeenat_fullname}
    \bibliography{main}
}

\end{document}